\documentclass{article}

\usepackage{arxiv}

\usepackage[utf8]{inputenc} %
\usepackage[T1]{fontenc}    %
\usepackage{hyperref}       %
\usepackage{url}            %
\usepackage{booktabs}       %
\usepackage{amsfonts}       %
\usepackage{amsmath}
\usepackage{nicefrac}       %
\usepackage{microtype}      %
\usepackage{graphicx}
\usepackage{tabularx}
\usepackage{longtable}
\usepackage{natbib}
\usepackage{doi}
\usepackage{cleveref}       %
\usepackage{orcidlink}      %
\usepackage{eso-pic}        %

\graphicspath{{figures/}}

\usepackage{xcolor}
\usepackage[most]{tcolorbox}
\definecolor{resource}{RGB}{0,120,60}
\newcommand{\resource}[1]{\textcolor{resource}{#1}}
\newcommand{\resurl}[1]{\href{#1}{\textcolor{resource}{\nolinkurl{#1}}}}

\definecolor{cardback}{RGB}{247,248,250}
\tcbset{card/.style={enhanced, colback=cardback, colframe=black!15, boxrule=0.5pt, arc=6pt,
  left=16pt, right=16pt, top=12pt, bottom=12pt}}
\renewenvironment{abstract}{\begin{tcolorbox}[card]%
  \centerline{\large\bfseries\scshape Abstract}\par\smallskip}{\end{tcolorbox}}
\newcommand{\field}[2]{%
  \par\noindent\makebox[6.8em][l]{\small\color{black!55}\textsc{#1}}#2\par\addvspace{2pt}}

\newcommand{\stat}[2]{{\Large\bfseries #1}\par\vspace{1pt}%
  {\scriptsize\color{black!60}\renewcommand{\newline}{\\}\begin{tabular}{@{}c@{}}#2\end{tabular}}}

\newcommand{\fig}[4]{%
  \begin{figure}[tbp]
    \centering
    \includegraphics[width=#3\linewidth]{#1}
    \caption{#4}
    \label{fig:#2}
  \end{figure}}

\title{MarathonTS: A Benchmark of\\Long Synthetic Time Series from Window Assembly\\and Long-Series Generators}

\date{}

\author{%
  Ilyas Varshavskiy\,\orcidlink{0000-0003-4924-2097} \quad
  Sergey Shulgin \quad
  Shuhrat Khalilbekov \\
  \textbf{Bonu Boboeva \quad Azizjon Azimi} \\
  {\normalfont zypl.ai} \\
  {\normalfont Corresponding author: \texttt{ilyas.varshavskiy@zypl.ai}} \\
}

\renewcommand{\shorttitle}{MarathonTS: Long Synthetic Time Series}

\hypersetup{
pdftitle={MarathonTS: A Benchmark of Long Synthetic Time Series from Window Assembly and Long-Series Generators},
pdfauthor={Ilyas Varshavskiy, Sergey Shulgin, Shuhrat Khalilbekov, Bonu Boboeva, Azizjon Azimi},
pdfkeywords={synthetic time series, benchmark, generative models, TSTR, fidelity},
colorlinks=true, linkcolor=blue!50!black, citecolor=blue!50!black, urlcolor=blue!50!black,
}

\begin{document}
\AddToShipoutPictureFG*{\put(\LenToUnit{1in},\LenToUnit{\paperheight-0.75in}){%
  \includegraphics[height=0.3in]{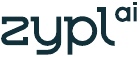}}}
\maketitle

\begin{abstract}
Many applications of synthetic time series require a series far longer than the windows on which
generative models are trained. Such a series can be assembled from a pool of short generated windows, by
choosing the windows, their order and the processing of their seams, or produced directly by a long-series
generator. We present MarathonTS, to our knowledge the first open benchmark that compares both families under
one protocol. Thirty methods each produce a series of the same length on 9 open hourly datasets with 10
seeds. The whole series is evaluated for fidelity (83 metrics in 12 groups, against train, validation and
out-of-time references), utility (models trained on synthetic and tested on real data in 14 classification
and regression tasks), memorisation and cost, with a statistical analysis frozen before the main run. The
methods differ in both confirmatory tasks, whereas the pre-specified contrasts of long-series generation,
window selection and seam processing are inconclusive under the joint rule of statistical and practical
significance. Descriptively, window selection often moves the series towards the real data, whereas seam
processing, tested on one assembly of the VAE pool, often reduces utility. No method trains the downstream
models as well as the real series (the transfer ratio is significantly below 1 for every method), although
the best method comes within a few per cent. A replication run of 13
ranked methods on a different hardware and software stack, which does not cover the proprietary zEDGE methods that lead the main leaderboard,
gives similar rankings with one change in top-tier membership. The code of the open methods, the data
hashes and the result tables are publicly available.


\vspace{0.6em}
\field{Release}{1.0.0}
\field{Install}{\resource{\texttt{pip install marathonts}}}
\field{Code}{\resurl{https://github.com/zypl-ai/MarathonTS}}
\field{Data}{\resurl{https://huggingface.co/datasets/zyplai/MarathonTS}}
\field{Leaderboard}{\resurl{https://huggingface.co/spaces/zyplai/MarathonTS}}
\field{Docs}{\resurl{https://zypl-ai.github.io/MarathonTS}}
\field{Licence}{Code: Apache-2.0; each dataset keeps the licence of its source (\cref{app:datasets})}
\end{abstract}

\vspace{0.35cm}
{\small\color{black!55}\textsc{MarathonTS at a glance}}\par\vspace{2pt}
{\setlength{\tabcolsep}{2pt}%
\noindent\begin{tabularx}{\linewidth}{@{}*{8}{>{\centering\arraybackslash}X}@{}}
\stat{9}{public\newline datasets} &
\stat{30}{methods\newline 2\,$\times$\,12\,+\,6} &
\stat{10}{seeds per\newline method} &
\stat{83}{fidelity\newline metrics} &
\stat{3}{fidelity refs\newline train\,/\,val\,/\,OOT} &
\stat{8\,+\,6}{TSTR tasks:\newline classif.\,+\,regr.} &
\stat{2}{utility tests\newline val\,/\,OOT} &
\stat{10}{research\newline questions} \\
\end{tabularx}}\par

\section{Introduction}
\label{sec:intro}

Synthetic time series are used to train and test forecasting and classification models when real data
are scarce, sensitive or cannot be shared. Many of these applications require a \emph{long} series: a
forecaster learns weekly and monthly patterns only from a series that contains them, a model of rare
events needs a sufficient number of such events, and a stress test needs regimes that last longer than a
few days. At the same time, generative models for time series are trained on windows of a fixed and
short length~\citep{yoon2019timegan,yuan2024diffusionts}. A window of one week at an hourly step contains
168 points, while the training series of a dataset covers months or years. This raises the question of how to obtain a synthetic series that is many windows long and remains realistic.

Two approaches to this problem can be distinguished. In the first approach, the series is
\emph{assembled} from a pool of short generated windows. Assembly involves three decisions: which windows
are taken, in what order, and how the seams between neighbouring windows are processed. Each of these
decisions has a long history outside time series, in concatenative speech synthesis~\citep{hunt1996unit},
video textures~\citep{schodl2000video} and image quilting~\citep{efros2001quilting}, and the window pool
can be produced by any generator. In the second approach, a \emph{long-series generator} produces the
whole series directly: a diffusion model extends the series window by window~\citep{yuan2024diffusionts},
or a pretrained forecasting foundation model~\citep{ansari2024chronos,das2024timesfm,petralia2026timestral}
continues the training series for as long as needed. The two families make different trade-offs.
Assembly keeps the quality of individual windows, but it can create seams and lose structure longer than
a window. A long-series generator does not select from a fixed window pool, but chunked generation can
introduce boundaries; it may also drift, smooth the series, or reproduce data seen in pretraining.
Both families therefore require evaluation of local transitions and longer-scale structure.

These families have not been compared under one protocol. A recent study varies how overlapping windows
are blended into a long series~\citep{long2026assembly}, but it does not vary which windows are taken and
in what order, and it does not compare assembly with long-series generators. Evaluations of synthetic
time series score sets of independent windows of the training length~\citep{ang2024tsgbench,
nikitin2024tsgm}; therefore, they cannot capture the seams, the structure longer than a window, or the
distance of the series to data from a later period. The real reference in such evaluations is usually the
training data or a random split of the same windows. Without a common protocol, it remains unclear when
assembly from windows is sufficient, which of its decisions matter, and whether a pretrained forecaster
is a good source of long synthetic series.

This study presents MarathonTS, a benchmark of long synthetic time series. To our knowledge, it is the
first open benchmark that compares assembly from a pool of generated windows with long-series
generators, including pretrained forecasting foundation models, under one protocol, and the first to
evaluate whole long series against a real reference from a later period (\cref{sec:related}). The contributions of the study are as follows:
\begin{itemize}
  \item \textbf{A common task and protocol} (\cref{sec:design,sec:methods}). Every method receives only the
  training part of a series and returns one series of length $K$ times the training window, the same
  for every method and approximately as long as the training part, in the same channels. The main run compares 30 methods on nine open hourly datasets with ten seeds:
  nine assemblers and three seam joiners on each of two window pools, and six long-series generators
  (Diffusion-TS, Chronos, TimesFM~3.0, Timestral and two sampling modes of zEDGE\_TS). Within one pool, all assemblers read the same
  windows, so they differ only in selection, order and seam processing.
  \item \textbf{Evaluation of the whole series} (\cref{sec:evaluation}). Fidelity is measured by 83 metrics
  in 12 groups, including groups for the seams, for structure longer than a window and for tail events,
  against three real references: train, a validation split and an out-of-time split. Utility is
  measured by training downstream models on the synthetic series only and testing them on real data
  (TSTR against TRTR) in 14 classification and regression tasks. Memorisation and cost are measured
  separately.
  \item \textbf{A pre-specified statistical analysis} (\cref{sec:statistics}). The analysis consists of
  rankings with uncertainty and robustness checks, research questions with an empirical noise level as
  the smallest effect of interest and tests of both difference and equivalence, and a small confirmatory
  family that carries the main conclusion. The analysis was frozen before the main run was carried out.
  \item \textbf{Open, reproducible and extensible software} (\cref{sec:software}). Every component is
  registered by name, so a new dataset, generator, assembler, metric or test enters the comparison through
  a plugin. Incremental recomputation and run profiles make it possible to run the benchmark on a single
  GPU.
\end{itemize}

The main run gives the following picture (\cref{sec:results}). The methods differ in both cells of the
confirmatory family, whereas the confirmatory comparisons of long-series generation with assembly, of
window selection with random selection and of seam processing with no processing are inconclusive under
the pre-specified joint rule of statistical and practical significance (\cref{sec:confirmatory}). Descriptively, window selection often moves the series towards the real data
compared with a random choice of windows, whereas the seam joiners, tested on the zseries assembly of the VAE pool, often reduce utility compared with
the same assembly without them. The top tier of the leaderboard, the methods not shown to be worse than the observed leader, holds 15 of
24 methods. Its first eleven places
are taken by the methods of the proprietary zEDGE pool and by zEDGE\_TS\footnote{zseries, zEDGE and zEDGE\_TS are developed by zypl.ai (\url{https://zypl.ai/}) and are evaluated as black boxes under the same protocol as the open methods; see Competing interests.}, and the best open method is the
kernel-herding assembler on the VAE pool. Among the long-series generators, again descriptively, Diffusion-TS and TimesFM are
competitive with assembly from the open VAE pool, Chronos and Timestral fall behind it, and assembly from
the zEDGE pool is ahead of all four. No method reaches the utility of the real series: the
transfer ratio is significantly below 1 for every method, although for the best methods the gap is small. A replication run on a different hardware and software stack, covering the 13 ranked
methods of the VAE pool and the open long-series generators (the zEDGE methods were not rerun), gives
similar rankings of these methods with one change in top-tier membership; the order within the tier differs
between the runs.

The rest of the paper is organised as follows. \Cref{sec:related} is devoted to related work;
\cref{sec:design} defines the task and the data; \cref{sec:methods} describes the compared methods;
\cref{sec:evaluation} presents the evaluation; \cref{sec:statistics} describes the statistical analysis;
\cref{sec:software} describes the software; \cref{sec:results} shows the results;
\cref{sec:limitations} discusses the limitations; \cref{sec:conclusion} contains the conclusions.

\section{Related work}
\label{sec:related}

Four directions of existing work are relevant to this study: generative models
for time series, the evaluation of synthetic time series, the assembly of long content from pieces in
other fields, and the resampling of real series in statistics.

\subsection{Generative models for time series}
\paragraph{Window generators.} Most generative models for time series learn the distribution of windows
of a fixed length. TimeGAN~\citep{yoon2019timegan} trains a GAN in a learned latent space with an
additional supervised loss on one-step-ahead transitions. Variational autoencoders, including the
$\beta$-VAE~\citep{higgins2017betavae} that is used as the pool generator in this study, encode a whole
window into one latent vector. Diffusion models such as CSDI~\citep{tashiro2021csdi} and
Diffusion-TS~\citep{yuan2024diffusionts} denoise a window as a whole and can also fill in its missing part
given the observed rest. In all of these models, the window length is fixed at training time, so the
length of a sample is bounded by what fits into the network and into the training windows. Few models
target long series directly: PSA-GAN~\citep{jeha2022psagan} grows the generated length progressively
during training, and a model with an imputation mode can extend a series window by window, conditioning
each new window on the tail of the previous one.

\paragraph{Pretrained forecasting models.} A second source of long series is pretrained forecasting
foundation models used as samplers. Chronos~\citep{ansari2024chronos} tokenises scaled values and samples
trajectories autoregressively; TimesFM~\citep{das2024timesfm} is a decoder-only model that forecasts
patches of the series; Timestral~\citep{petralia2026timestral} is a recent family of probabilistic forecasting
foundation models. Conditioned on the end of the training series, such a
model produces a continuation of any length without being trained on the dataset. Its output is a forecast rather than a sample of the training distribution: the median forecast is
a single deterministic path, smoother than any sampled trajectory. Benchmarks of these models, such as
TSFM-Bench~\citep{li2025tsfmbench}, evaluate them as forecasters, by the error of the forecast, and not as
sources of synthetic data. Whether these models are good sources of long synthetic series is an
empirical question, which the benchmark answers on equal terms with the trained generators.

\subsection{Evaluation of synthetic time series}
\paragraph{Evaluation axes.} Synthetic time series are evaluated along three axes~\citep{ang2024tsgbench}.
\emph{Fidelity} compares the synthetic and real distributions: through two-sample statistics such as the
maximum mean discrepancy~\citep{gretton2012mmd}, through classifier two-sample
tests~\citep{lopezpaz2017c2st} and their recurrent variant, the discriminative score
of~\citet{yoon2019timegan}, through the Fréchet distance~\citep{heusel2017fid} between window embeddings
(Context-FID uses TS2Vec~\citep{yue2022ts2vec} embeddings~\citep{jeha2022psagan}), through precision,
recall, density and coverage in a feature space~\citep{naeem2020prdc}, and through hand-crafted
statistics of the series such as catch22~\citep{lubba2019catch22}. \emph{Utility} measures whether the
synthetic data can replace the real data in a downstream task: train on synthetic, test on real (TSTR)
against train on real, test on real (TRTR)~\citep{esteban2017rcgan}, and the closely related predictive
score of~\citet{yoon2019timegan}. \emph{Privacy} measures whether the generator reproduces its training data,
usually through the distance from a synthetic sample to its nearest training sample.

\paragraph{Benchmarks and libraries.}
TSGBench~\citep{ang2024tsgbench} compares generators on a common set of datasets, metrics and a ranking
procedure; TSGM~\citep{nikitin2024tsgm} provides generators, augmentations and evaluation metrics in one
framework. Recent benchmarks narrow the question. ConTSG-Bench~\citep{lan2026contsg} evaluates generation
under a condition, by fidelity and by adherence to the condition. CTBench~\citep{ang2025ctbench} evaluates
generators of cryptocurrency series in a walk-forward design, in which TSTR utility is measured on the real
days that follow each training period. \Citet{amorim2026privacy} train
forecasters on synthetic series that fully replace the real ones (TSTR), test them on the final
observations of each series and weigh their error against a distance-based privacy risk. Seq2Synth~\citep{kwon2026seq2synth} shows for sequential tabular data that
rankings by static fidelity and by temporal fidelity diverge. In all of these works, as in the papers
that introduce the generators, the evaluated object is a set of independent windows or sequences of the
training length, and the fidelity reference is the data the generator was trained on or a random split
of the same data; a later period, where it is used, enters only as the test set of the utility tasks. Therefore, the properties that appear only in a long series are left out: the seams
between consecutive windows, structure longer than a window (slow trends, regimes, the distribution of
long excursions), and the distance to data from a later period. MarathonTS evaluates whole series of one
common length, adds metric groups for seams and long-range structure, and scores every series against
three real references, including an out-of-time split (\cref{sec:evaluation}).

\paragraph{Assembly of generated windows.} The closest study to ours is that
of~\citet{long2026assembly}, which starts from the same problem: differentially private generators
produce windows of a fixed length, while forecasters need a long series. In that study, the windows of a
differentially private diffusion model are placed in their generated order, overlapping windows are
blended with different overlap rates and weighting schemes, and the result is evaluated by the continuity
at the boundaries, by fidelity and by TSTR utility against TRTR on four datasets. The authors find that
the best assembly depends on the forecaster and that better continuity or fidelity does not reliably give
a lower forecast error, which agrees with our premise that the series must be judged by several criteria
at once. The study varies how overlapping windows are blended; it does not vary which
windows are taken and in what order, and it does not compare assembly with long-series generators.

\paragraph{Position of this study.} To our knowledge, no other open
benchmark compares the selection, order and seam processing of generated windows with long-series
generators, including pretrained forecasting foundation models, under one protocol, and scores the
fidelity and utility of whole long synthetic series against an out-of-time real reference. Each part of
this protocol has precedents, cited above and in \cref{sec:evaluation}; the combination is new. The
statement rests on searches of arXiv and the web in October and November 2026 for benchmarks of time series
generation, for the assembly of generated windows into long series, and for forecasting foundation models
used as generators of synthetic data.

\subsection{Assembling long content from pieces}
\paragraph{Selection and order of pieces.} The assembly of a long output from a library of short pieces
has a long history outside time series. Concatenative speech synthesis selects and orders recorded units
by minimising the sum of a target cost and a concatenation cost~\citep{hunt1996unit}. Video
textures~\citep{schodl2000video} and motion graphs~\citep{kovar2002motion} build a graph of transitions
between similar frames or poses and produce an unbounded sequence by walking it.

\paragraph{Processing of seams.} Image quilting~\citep{efros2001quilting} places overlapping patches and
cuts each seam along a minimum-error boundary, and graphcut textures~\citep{kwatra2003graphcut} find the
seam by a minimum cut. WSOLA~\citep{verhelst1993wsola} joins overlapping audio segments at the shift that
maximises waveform similarity, which is a refinement of plain overlap-add~\citep{allen1977stft}. These four fields (speech, video, images and audio) separate the same three decisions that define assembly from a window pool: which pieces are
used, in what order, and how the seams are processed. Several assemblers and seam joiners of the
benchmark (\cref{sec:methods}) are one-dimensional versions of these methods, each with the seam
criterion of its own literature.

\paragraph{Reconciliation during generation.} Diffusion models offer a different route to long content:
overlapping pieces are reconciled during generation instead of after it. RePaint~\citep{lugmayr2022repaint}
fills in a missing region by resampling the known part at every denoising step;
MultiDiffusion~\citep{bartal2023multidiffusion} averages overlapping windows at every step;
DiffCollage~\citep{zhang2023diffcollage} generates the pieces of a factor graph in parallel. The analogue
of these methods for time series is the window-by-window extension of a diffusion generator described
above.

\subsection{Resampling real series}
\paragraph{Bootstraps and surrogates.} Statistics has its own long tradition of producing new series from
an observed one. Block bootstraps concatenate blocks of the real series: of fixed
length~\citep{kunsch1989jackknife}, wrapped into a circle~\citep{politis1992circular}, of random
length~\citep{politis1994stationary}, with an automatically chosen length~\citep{politis2004blocklength},
or aligned with the seasonal phase~\citep{dudek2014seasonal}. The sieve bootstrap~\citep{buhlmann1997sieve}
fits an autoregression and resamples its residuals, the Markov bootstrap~\citep{rajarshi1990markov}
resamples from an estimated transition density, and AAFT and IAAFT
surrogates~\citep{theiler1992surrogate,schreiber1996iaaft} randomise Fourier phases while keeping the
spectrum and the distribution of values.

\paragraph{Matching of the next piece.} Two schemes choose the next piece by how well it continues the
current one. The matched-block bootstrap~\citep{carlstein1998matched} draws the next block with a kernel
of how well its beginning matches the end of the current block, and the nearest-neighbour
bootstrap~\citep{lall1996knn} takes the successor of one of the $k$ nearest neighbours of the current
state.

These methods resample the real training series itself, so their blocks are
copies of real data by construction; they answer a different
question from the one studied here. In this study, their matching and resampling ideas are used only on
top of a pool of generated windows: the matched-block and $k$-nearest-neighbour assemblers of
\cref{sec:methods} apply these two selection rules to the pool instead of to the real series.

The task, the data and the splits of the benchmark are described in the next section.

\section{Benchmark design}
\label{sec:design}

\subsection{Task: a long synthetic series}
\label{sec:task}
Let $x_1, \dots, x_n$ be an hourly univariate series, split in time into a training part of length
$n_{\mathrm{train}}$, a validation part (val) and an out-of-time part (OOT). A generative model is
trained on windows of length $T$. The task is to produce one synthetic series of length $K \cdot T$,
where
\begin{equation}
  T = 168, \qquad K = \operatorname{round}\!\left(n_{\mathrm{train}} / T\right).
\end{equation}
That is, the synthetic series is approximately as long as the real training series and $K$ times as
long as a training window. The window length is one week of hours for every dataset, so a window contains
the full weekly cycle, but nothing longer than a week.

Every method receives the training part only and returns its series in the same three channels: the
value scaled with the bounds of train and the hour of the week encoded on a circle
(\cref{sec:preprocessing}). For an assembly method, the calendar of the series is decoded from the two
calendar channels of its windows: their angle is rounded to the nearest hour of the week. A window
generator produces these channels together with the value, so the decoded calendar can deviate from a
valid one inside a window as well as at a seam. The open long-series generators generate the value only
and receive the calendar: Diffusion-TS starts it at a random hour of the week, and the pretrained
forecasters continue the calendar of train; zEDGE\_TS generates it as part of its windows. The calendar of
a series enters the weekly and daily profiles of the fidelity metrics (F3.8, F4.1--F4.3, F8.4) and the
hour-of-week feature of every downstream task (\cref{sec:utility}). The series is then converted back to real units and evaluated
(\cref{sec:evaluation}). Each method is run on every dataset with 10 seeds, which gives one series per
(dataset, method, seed).

The compared methods are divided into two families:
\begin{itemize}
  \item an \emph{assembly} method takes a pool of $N \ge K$ windows sampled from a window generator,
  selects $K$ of them without repetition, orders them, and optionally processes the seams between
  neighbours;
  \item a \emph{long-series generator} produces the series of length $K \cdot T$ directly, without a
  pool.
\end{itemize}
All assembly methods on one pool read the same windows, so they differ only in selection, order and seam processing.
The methods of both families are described in \cref{sec:methods}.

\subsection{Datasets}
\label{sec:datasets}
\Cref{tab:datasets} lists the nine open datasets used in the benchmark. Each dataset yields a single
target series, brought to an hourly step, and every dataset is used in every downstream task, so the
whole benchmark runs on one set of datasets.

The datasets are chosen so that they differ in the properties that make long series hard to generate:
\begin{itemize}
  \item Electricity, PJME, Metro and Bike have strong weekly seasonality;
  \item Jena and ETTh1 have strong daily but weak weekly seasonality, which checks that the hour-of-week
  calendar does not impose spurious structure;
  \item PM2.5, Household and Bitcoin have heavy tails and spikes;
  \item Bitcoin has volatility clustering and almost no seasonality;
  \item Bike, ETTh1 and Bitcoin drift in level.
\end{itemize}
Household, PM2.5, Metro, Bike and PJME contain gaps or duplicate hours that need preprocessing. The datasets also differ in length: on the full series, $K$ ranges from about 60
(ETTh1, Bike) to 519 (PJME). Sources and licences are listed in \cref{app:datasets}.

\begin{table}[t]
  \caption{Datasets. \emph{Hours}: length of the target series after conversion to an hourly step,
  from the first to the last valid value. $K$: number of windows on the full series (60\% train,
  $T = 168$, profile \texttt{max}); profiles \texttt{min} and \texttt{mid} take the last 10\,000 hours
  of every dataset, which gives $K = 30$ everywhere.}
  \label{tab:datasets}
  \centering
  \small
  \begin{tabular}{llrrr}
    \toprule
    Dataset & Target & Source step & Hours & $K$ \\
    \midrule
    ETTh1                              & \texttt{OT}                  & 1 h    & 17\,420          & 62 \\
    Electricity Load Diagrams          & sum over 370 clients         & 15 min & 35\,065          & 125 \\
    Household Electric Power           & \texttt{Global\_active\_power} & 1 min & $\approx 34\,590$ & 124 \\
    PJM Hourly Energy (PJME)           & \texttt{PJME\_MW}            & 1 h    & 145\,392         & 519 \\
    Beijing PM2.5                      & \texttt{pm2.5}               & 1 h    & 43\,800          & 156 \\
    Jena Climate 2009--2016            & \texttt{T (degC)}            & 10 min & $\approx 70\,000$ & 250 \\
    Metro Interstate Traffic           & \texttt{traffic\_volume}     & 1 h    & $\approx 29\,000$ & 103 \\
    Bike Sharing (hourly)              & \texttt{cnt}                 & 1 h    & 17\,544          & 63 \\
    Bitcoin Historical Data            & \texttt{Close} (log price)   & 1 min  & $\approx 85\,000$ & 305 \\
    \bottomrule
  \end{tabular}
\end{table}

\paragraph{Leakage into pretrained models.} Pretrained foundation models may have seen a benchmark
dataset during pretraining or post-training, and in this case their continuation of the training series is not
out-of-sample. Therefore, every pair of a pretrained model and a dataset is audited against the primary
descriptions of the training corpora and, where the corpus is public, against the data themselves.
A model is left out on a dataset if its corpus contains the same data (an \emph{exact} pair) or the same
object at other stations, years or resolution (a \emph{related} pair); pairs that share only the
phenomenon are kept. By this rule, Chronos is left out on Electricity (exact) and Beijing PM2.5
(related), and TimesFM~3.0 on Household (exact), Beijing PM2.5 and Bitcoin (related). Timestral is
pretrained on synthetic data only, but the real data of its post-training are not
listed~\citep{petralia2026timestral}, so its pairs are marked unknown and kept
(\cref{sec:limitations}). The exclusion applies to that method on that dataset in every
ranking and test, while the other methods keep every dataset. The full audit is given in
\cref{app:datasets}.

\subsection{Preprocessing and temporal split}
\label{sec:preprocessing}
\Cref{fig:preprocessing} shows the path from the raw files to the model inputs. The raw files are
converted losslessly to Parquet. Each dataset is then reduced to an hourly series as follows:
\begin{itemize}
  \item for the sub-hourly sources (Electricity, Household, Jena), the hourly mean is taken;
  \item for Bitcoin, the log of the last price of the hour is taken;
  \item for the hourly sources, the values are used as they are.
\end{itemize}
Gaps of up to 6 hours are interpolated; longer gaps stay missing, and windows that contain them are
dropped. Metro is taken after its last gap longer than 30 days, and a missing hour of Bike means zero
rentals.

The series is split in time into train, val and OOT, with an embargo of 168 hours between neighbouring
parts (\cref{fig:split}). The embargo is as long as the longest label horizon and the longest feature lag
of the downstream tasks (\cref{sec:utility}), so no label or feature of a test row reaches into train.
Profiles \texttt{min} and \texttt{mid} take the last 10\,000 hours of each dataset and split them
50/25/25. After the embargoes, every dataset has 5\,000 training hours, 2\,332 val hours and 2\,332 OOT
hours, and $K = 30$. Profile \texttt{max} takes the full series and splits it 60/20/20, which gives the
values of $K$ in \cref{tab:datasets}. The results in \cref{sec:results} are obtained for profile
\texttt{min}.

\fig{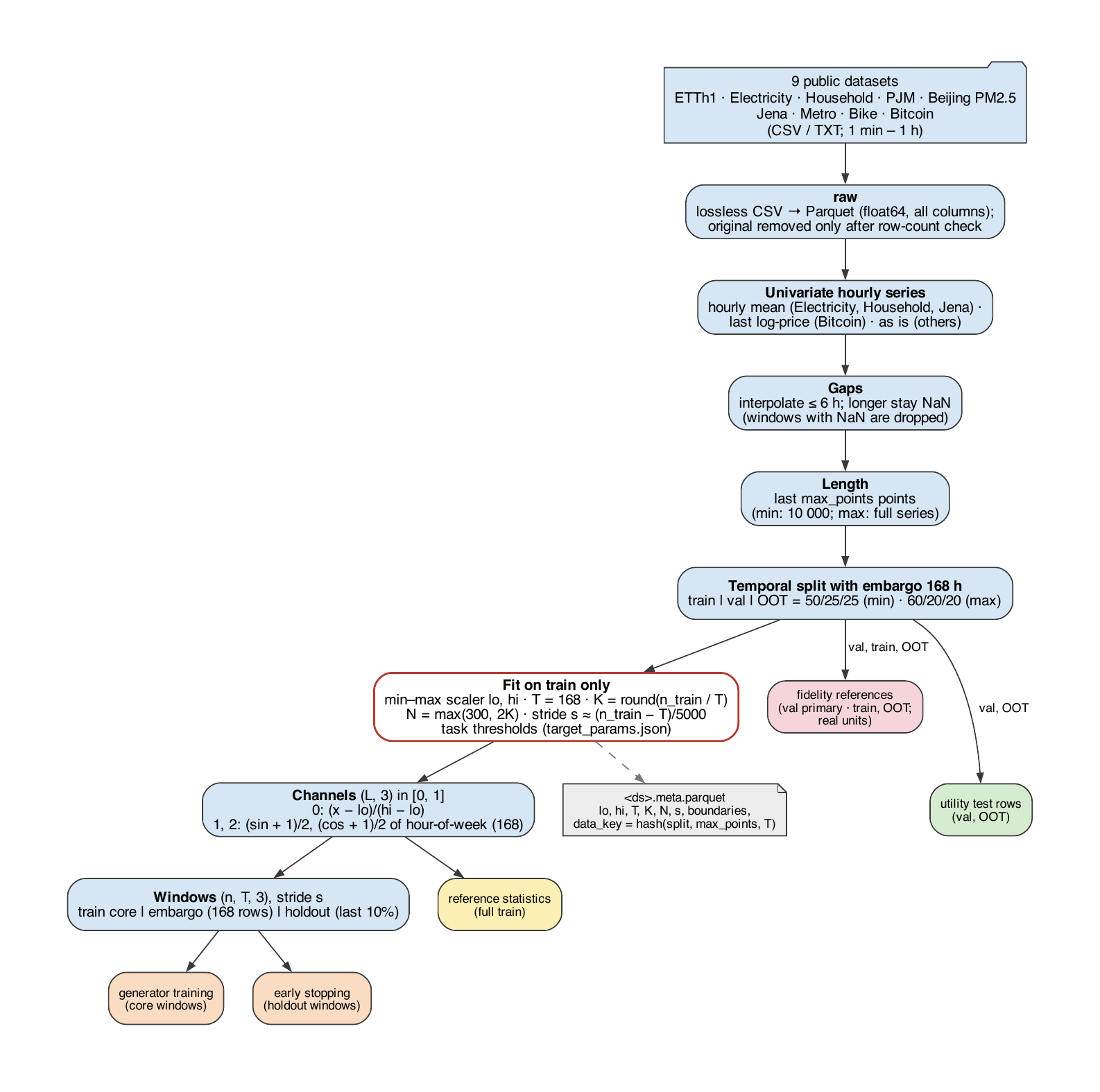}{preprocessing}{0.9}{From raw files to model inputs. Each dataset
becomes one hourly univariate series. The series is split in time with embargoes. All statistics
(scaler, window length, number of windows, task thresholds) are fitted on the training split only.
The series is then encoded as three channels in $[0, 1]$ and cut into windows (red box: the leakage
boundary).}

\fig{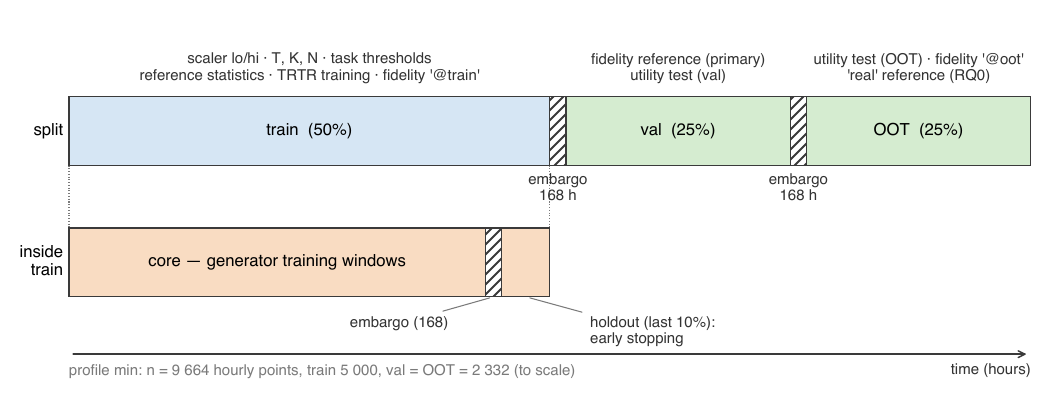}{split}{0.9}{Temporal split (profile \texttt{min}, drawn to scale).
Hatched bars are embargo gaps of 168\,h, wide enough to cover every label horizon $h \le 168$ and
the longest feature lag (168). Inside train, the last 10\% is held out to stop generator training
early. Every statistic the methods may use comes from train. Val is the primary fidelity reference;
train and OOT are also references. Val and OOT are the utility test sets.}

Everything a method may use is fitted on train only: the min--max bounds of the value, $T$, $K$, the
pool size $N = \max(300, 2K)$, the stride of the training windows, and the thresholds of the
downstream tasks. The task thresholds are computed once per split, reviewed and committed to the
repository. A run only reads them, and a file computed on another split is refused, so a threshold can
never come from the val or OOT part of a run. The series is encoded as three channels in $[0, 1]$:
\begin{equation}
  \left(\frac{x_t - \ell}{u - \ell},\;
        \frac{1 + \sin(2\pi\, w_t / 168)}{2},\;
        \frac{1 + \cos(2\pi\, w_t / 168)}{2}\right),
\end{equation}
where $\ell$ and $u$ are the minimum and maximum of train and $w_t \in \{0, \dots, 167\}$ is the hour of
the week. Training windows of shape $(T, 3)$ are cut with the smallest stride that gives at most about
5\,000 windows (stride 1 in profile \texttt{min}). The last 10\% of train is a holdout for early stopping
of the generators. It is separated from the rest of train (the \emph{core}) by another embargo of 168
rows, so that core and holdout windows do not overlap.

Each part of the split has a fixed role:
\begin{itemize}
  \item core windows train the generators, and holdout windows stop their training;
  \item val is the primary real reference for fidelity and one of the two utility test sets;
  \item train and OOT are the other two fidelity references: train shows how well a method reproduces
  the data it was fitted on, val measures generalisation, and OOT measures generalisation under a shift
  in time;
  \item OOT is also the second utility test set.
\end{itemize}

\subsection{Window-pool generator}
\label{sec:pool}
The benchmark uses two pool generators. The open one is a convolutional $\beta$-VAE
\citep{higgins2017betavae} from the TSGM library~\citep{nikitin2024tsgm}, shown in \cref{fig:vae}. The
other one, zEDGE, is a proprietary window generator of zypl.ai, evaluated as a black box under the
same protocol: it is trained on the core training windows only and returns a pool of windows in the same
three channels.

The VAE encodes a window of shape $(T, 3)$ into a 16-dimensional latent vector and is trained with the
loss ``reconstruction $+\, \beta \cdot \mathrm{KL}$'' with $\beta = 2$ (Adam, learning rate $10^{-3}$,
batch 64, up to 400 epochs). Training is stopped early on the holdout loss with a patience of 50 epochs,
and the best weights are restored; the first 150 epochs are not eligible for stopping. The pool
consists of $N = \max(300, 2K)$ windows decoded from latent vectors drawn from a shrunk prior
$\mathcal{N}(0, 0.5^2 I)$, that is, at a reduced sampling temperature, which is known to trade the
diversity of samples for their quality~\citep{vahdat2020nvae}. The learning rate and the batch size follow TSGM; $\beta$, the latent size, the
epoch limits and the prior scale were taken from earlier work of the authors on other data and
were not tuned on this benchmark (\cref{sec:limitations}). The decoder ends with a sigmoid, so every window stays within the range of
train. The pool is drawn
anew for every seed and written to disk, and every assembly method of that seed reads the same pool. A
zEDGE pool is used in the same way.

\fig{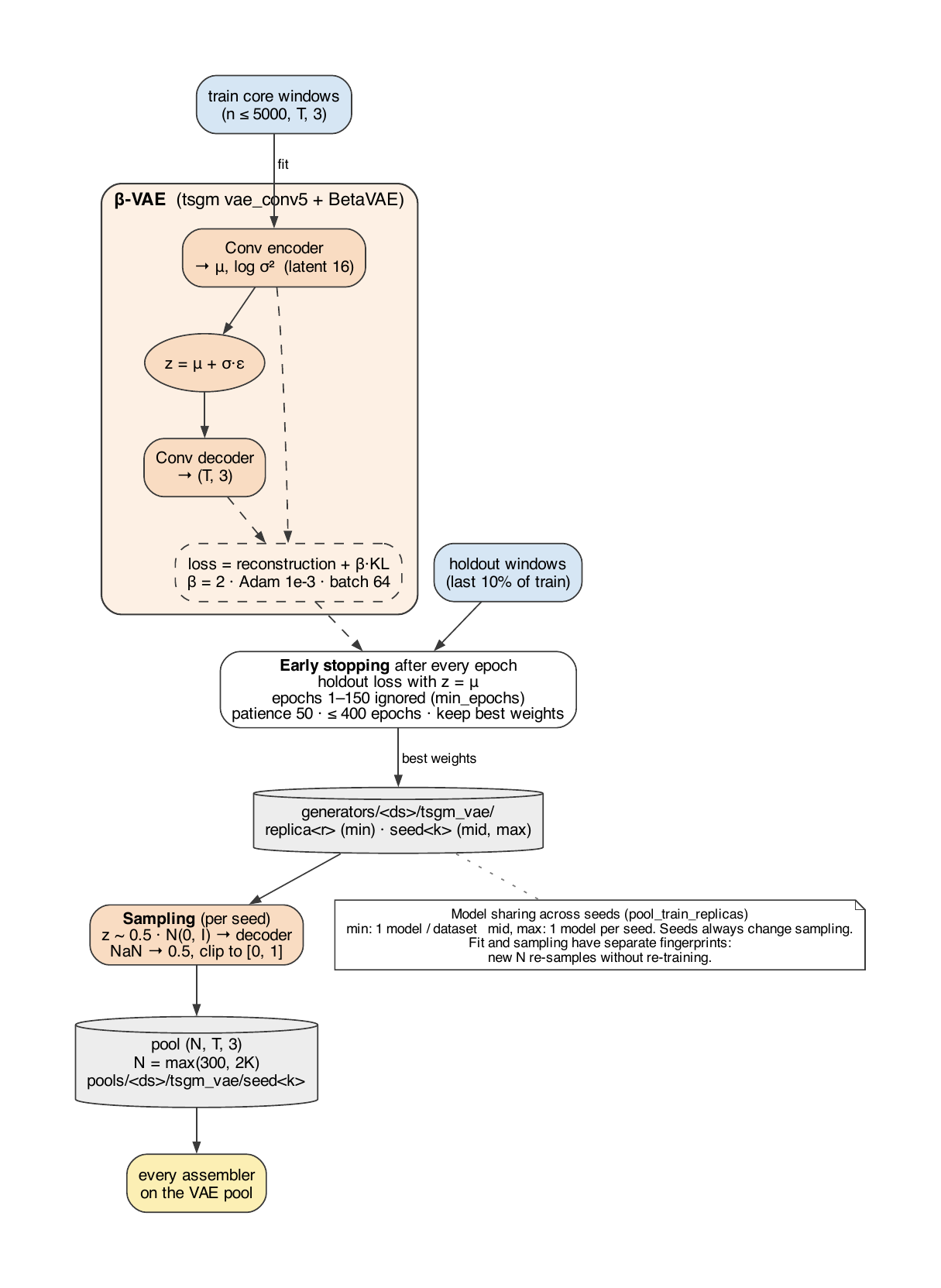}{vae}{0.85}{The window-pool generator. A convolutional $\beta$-VAE is
trained on core training windows and stopped early on the holdout loss. After training, $N$ windows
are sampled from a shrunk prior and written to disk as the pool that every same-pool method shares.}

In profile \texttt{min}, one VAE is trained per dataset, and the seeds change only the sampling of the
pool. In profiles \texttt{mid} and \texttt{max}, a VAE is trained on every seed, so the seeds also
capture the variance of training. When a model is shared by several seeds, its training time is still
charged to every seed in the cost comparison. The pool generator is a component like any
other: any generator that outputs windows in the same three channels can replace the VAE, and every
assembly method then runs on the new pool unchanged (\cref{sec:software}). The methods of both pools enter the same rankings and research questions as
every other method (\cref{sec:two-pools}).

The compared methods are described in the next section.

\section{Compared methods}
\label{sec:methods}

A compared method is either a pool generator combined with an assembler (and optionally a seam joiner),
or a long-series generator. \Cref{fig:methods} shows both families. The benchmark uses two pool
generators: the $\beta$-VAE of \cref{sec:pool} and zEDGE, a proprietary window generator of zypl.ai.
Every assembler is run on the pool of each generator, and the name of a method is the pool prefix
followed by the assembler. For example, \texttt{vae\_greedy} and \texttt{zedge\_greedy} apply the same
assembler to the two pools. Within one pool, all assemblers read the same windows, so they differ only
in which windows they select, in what order, and how they process the seams.

The main run (profile \texttt{min}) compares 30 methods:
\begin{itemize}
  \item on each pool, nine assemblers and the three seam joiners of \cref{sec:methods-b};
  \item six long-series generators: four open ones and two sampling modes of zEDGE\_TS.
\end{itemize}
The methods and their settings are listed in \cref{app:methods}.

The comparison is kept neutral by two principles. First, each alternative assembler uses the seam
criterion of its own literature. Second, the three solvers that have no criterion of their own (greedy,
ATSP and PCTSP) all minimise the same generic cost, the \emph{jump at the seam}: the Euclidean distance
between the last point of a window and the first point of the next, over all three channels. This is the
frame distance of video textures~\citep{schodl2000video}. Every generated series is stored with the
source of each point and a flag for the seams, so that the seam metrics (\cref{sec:fidelity}) apply to
every method. For a long-series generator, the boundaries between its generation chunks count as seams.

\fig{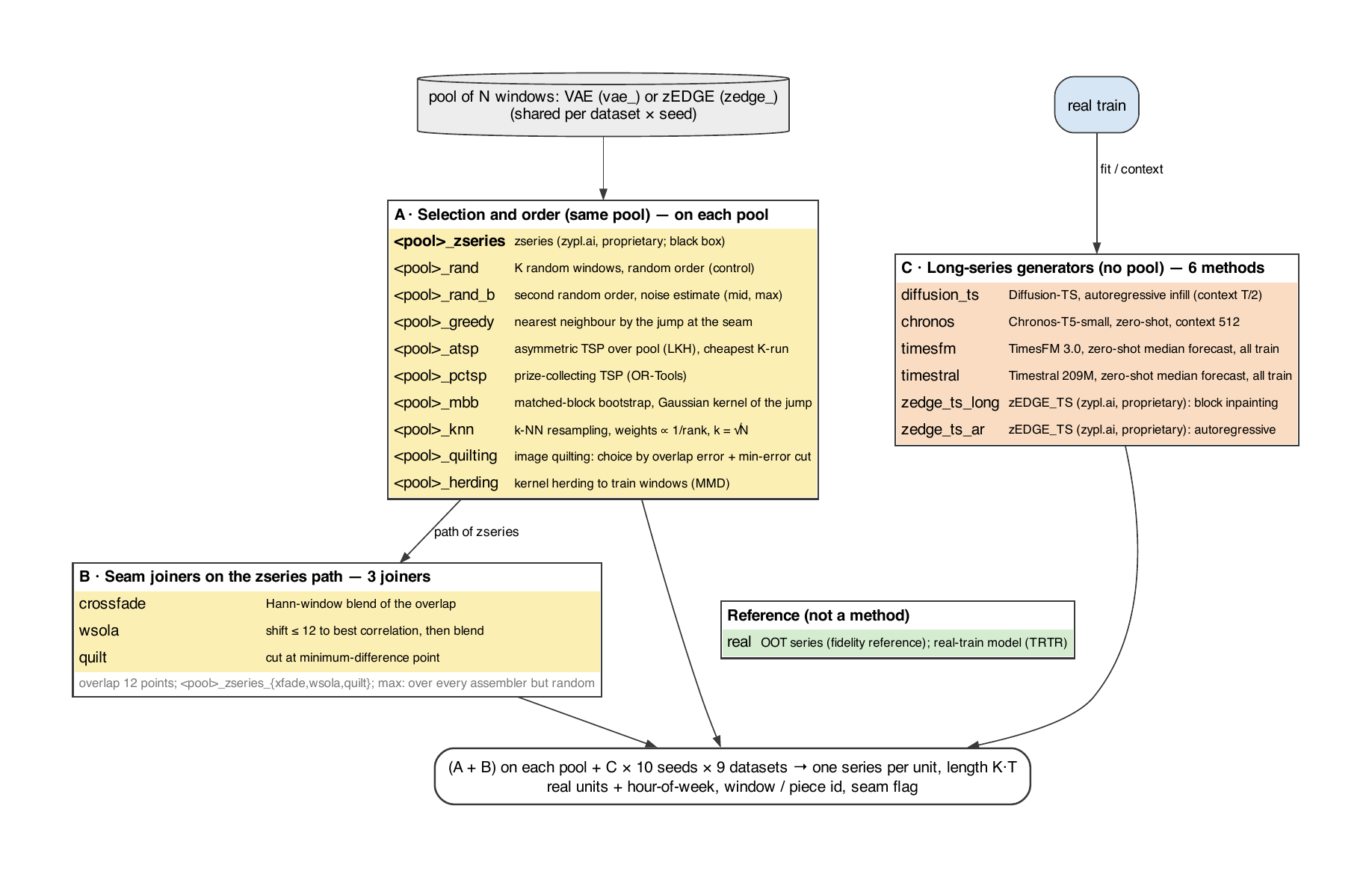}{methods}{1}{The compared methods. Groups A and B take windows
from the same pool, so they differ only in which windows they pick, in what order, and how seams
are handled. Group C generates long series directly. Every method returns one series of length
$K \cdot T$ per (dataset, seed).}

\subsection{A. Window selection and order}
\label{sec:methods-a}
Every assembler of group A draws $K$ windows from the pool of $N$ without repetition. The assemblers of
the group are described below.

\paragraph{Random path.} $K$ random windows are taken in random order (\texttt{rand}). This is the
control of the group: any effect of an assembler is measured against what ``just other windows of the
same pool'' give. In profiles \texttt{mid} and \texttt{max}, a second random path with an independent
seed (\texttt{rand\_b}) is used to estimate the noise level of the comparison.

\paragraph{Solvers by the jump at the seam.} \emph{Greedy nearest neighbour} starts from a random
window and always appends the window with the smallest jump. \emph{ATSP} solves the asymmetric
travelling salesman problem over the whole pool with the Lin--Kernighan--Helsgaun heuristic
\citep{helsgaun2000lkh} and takes the cheapest run of $K$ consecutive windows from the tour.
\emph{Prize-collecting TSP}~\citep{balas1989pctsp} may skip windows at a penalty, so it chooses at once
which $K$ of the $N$ windows to use and their order.

\paragraph{Resampling rules on the pool.} The \emph{matched-block bootstrap}
\citep{carlstein1998matched} is a Markov chain over the pool: the next window is drawn with probability
proportional to $\exp\!\left(-\tfrac{1}{2}(d_{ij}/h)^2\right)$, where $d_{ij}$ is the jump from window $i$
to window $j$ and the bandwidth $h$ is the median jump. \emph{$k$-nearest-neighbour resampling}
\citep{lall1996knn} draws the next window among the $k = \sqrt{N}$ windows whose first point is nearest to
the last point of the current one, with weights inversely proportional to the rank.

\paragraph{Image quilting.} Following~\citet{efros2001quilting}, the next window overlaps the current one
by 12 points and is drawn at random among the windows whose overlap error is within $1.1$ times the best.
The overlap is then cut at the point of minimum error.

\paragraph{Kernel herding.} $K$ windows are selected greedily to minimise the MMD to the training windows
\citep{chen2010herding,gretton2012mmd} and are put in random order. This assembler changes only the
selection, not the order, and therefore makes it possible to separate the two effects.

\paragraph{zseries.} zseries is a proprietary assembler of zypl.ai under development
(\texttt{vae\_zseries}, \texttt{zedge\_zseries}). It is evaluated as a black box: its results are
reported and ranked like those of any other method, but its code and description are not released. It uses the thresholds of task U2.1, which are fixed on
train (\cref{sec:preprocessing}); no open assembler uses them (\cref{sec:limitations}).

\subsection{B. Seam processing}
\label{sec:methods-b}
A seam joiner takes the ordered windows of an assembler and processes the seams between them.
Neighbouring windows overlap by 12 points. The following joiners are used:
\begin{itemize}
  \item \emph{crossfade} blends the overlap with a Hann window, as in overlap-add
  synthesis~\citep{allen1977stft};
  \item \emph{WSOLA}~\citep{verhelst1993wsola} first shifts the next window by up to 12 points to
  maximise the correlation in the overlap and then blends;
  \item the \emph{boundary cut} of image quilting~\citep{efros2001quilting} switches from one window to
  the next at the point of smallest difference, which in one dimension is the minimum-error cut.
\end{itemize}
Because of the overlap, the joined series is shorter than $K \cdot T$. Therefore, the assembler supplies
more than $K$ windows, and the result is truncated to $K \cdot T$. In profiles \texttt{min} and
\texttt{mid}, the joiners are applied to the path of zseries; in profile \texttt{max}, to the path of
every assembler except the random path and image quilting, which has its own cut. A seam-processing
variant is ranked only when every assembler has it (\cref{sec:statistics}).

\subsection{C. Long-series generators}
\label{sec:methods-c}
The methods of group C produce the series of length $K \cdot T$ without a pool, as shown in
\cref{fig:long}.

\paragraph{Diffusion-TS.} The official implementation of Diffusion-TS~\citep{yuan2024diffusionts} is trained
on the same core windows as the pool generators and stopped early on the holdout windows. The series is
grown by autoregressive infilling: the first window is generated unconditionally, and each next window
takes the last $T/2$ generated values as known context and fills in the remaining $T/2$ (50 DDIM steps).
The calendar channels are given at every position, and only the value is generated. One
model per dataset is trained in profiles \texttt{min} and \texttt{mid}, and one model per seed in profile
\texttt{max}.

\paragraph{Chronos.} Chronos~\citep{ansari2024chronos} (\texttt{chronos-t5-small}; \texttt{chronos-t5-base}
in profile \texttt{max}) is used zero-shot with the default sampling of its library. It samples one
trajectory that continues the last 512 training values, 64 points at a time. Each chunk is appended to
the context, which is then cut back to the last 512 points.

\paragraph{TimesFM and Timestral.} We evaluate median-output configurations of TimesFM~3.0~\citep{das2024timesfm}
and Timestral-209M~\citep{petralia2026timestral}, used zero-shot as documented, with pinned model revisions.
Both models forecast quantiles, and the benchmark takes their median, a deterministic configuration
(\cref{sec:limitations}). The context is the whole training series, up to the limit of the model (15\,360 and 32\,768
points), and the series is the median forecast of the $K \cdot T$ steps after train. TimesFM produces it
in one pass, without seams. Timestral produces it in blocks of 1\,024 steps and feeds the median of each
block back as context, so the block boundaries are seams. A median forecast contains no sampling noise.
Therefore, these two methods return the same series for every seed and have no between-seed variance;
this is taken into account in the tests (\cref{sec:statistics}).

\paragraph{zEDGE\_TS.} zEDGE\_TS is a proprietary long-series generator of zypl.ai under development. It
works on the latent vectors of the zEDGE pool generator: one sequence model is trained per dataset and
shared by the seeds, which change only the sampling. It is evaluated in two sampling modes, which use
the same trained model. \texttt{zedge\_ts\_long} builds the long latent sequence in blocks of 12 latent
vectors; each next block is generated by inpainting, conditioned on the last three latent vectors of the
previous block. \texttt{zedge\_ts\_ar} is autoregressive: it adds one latent vector at a time,
conditioned on the preceding 11; this is slower, but every step agrees with a longer context. In both
modes, every latent vector is decoded into a window, and neighbouring windows, which overlap by $T/2$
points, are joined by a cosine crossfade; the starts of the decoded windows count as seams.

\fig{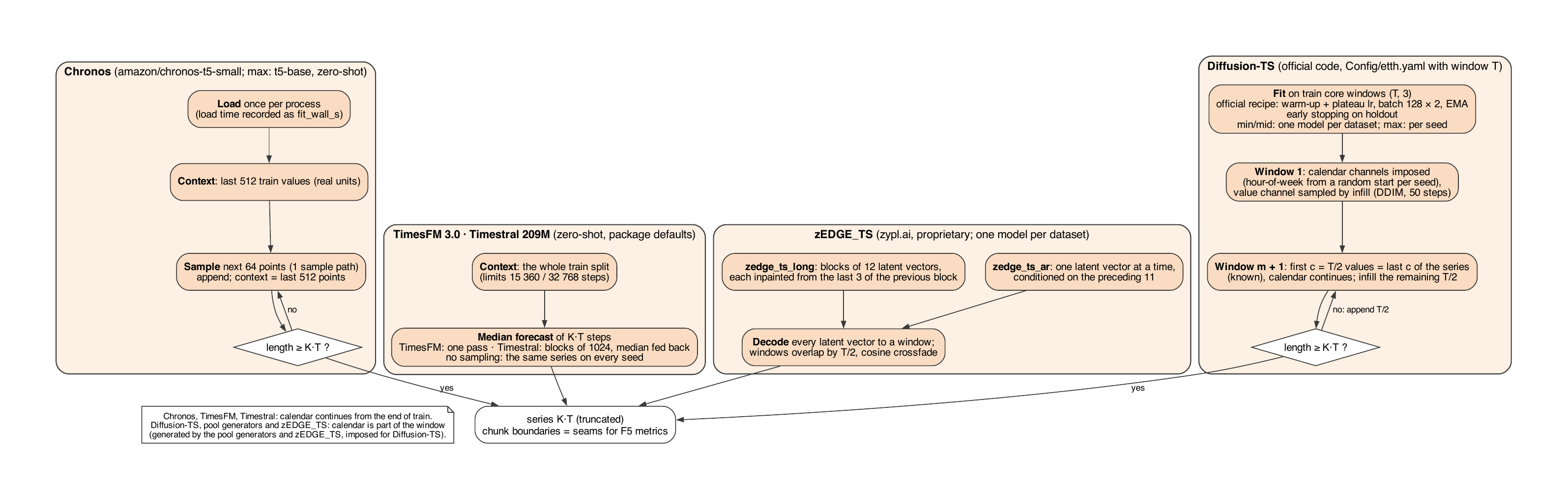}{long}{1}{The baselines that generate long series directly.
Diffusion-TS grows the series by autoregressive infilling: each new window takes the last $T/2$
generated values as known context and fills in the rest. Chronos samples continuations of the end
of the training series in chunks of 64 points. TimesFM and Timestral give their median forecast of the
$K \cdot T$ steps after train, with the whole train as context (up to the model's limit). zEDGE\_TS builds
a sequence of latent vectors in blocks (\texttt{zedge\_ts\_long}) or one vector at a time
(\texttt{zedge\_ts\_ar}) and decodes it into overlapping windows. Boundaries
between chunks or blocks count as seams in the seam metrics.}

The calendar is obtained in three ways: the pool generators
and zEDGE\_TS generate it as part of a window, Diffusion-TS has it imposed, and the pretrained forecasters continue the
real calendar of train. Therefore, the metrics of calendar continuity are left out of every ranking and
research question. A difference between methods can still come from the calendar as well as from the values;
\cref{sec:results-checks} gives a check. The pretrained forecasters are left out on the datasets that leaked
into their pretraining data (\cref{sec:datasets}).

\subsection{Methods not included}
\label{sec:methods-out}
Seam inpainting with CSDI~\citep{tashiro2021csdi}, which would fill in the zone around each seam
conditioned on both neighbours, needs a separately trained diffusion model and is not included. TimeGAN
\citep{yoon2019timegan} and other window generators can replace or extend the pool generators without
changes to the code (\cref{sec:software}), but they are not part of the main run. The block bootstraps
and surrogates of \cref{sec:related} resample the real training series and copy it by construction, so
they answer a different question. Their selection rules enter the benchmark only on top of a generated
pool, through the matched-block and nearest-neighbour assemblers.

The evaluation of the generated series is described in the next section.

\section{Evaluation}
\label{sec:evaluation}

In this study, every series is evaluated along three axes: its fidelity to the real data
(\cref{sec:fidelity}), its utility for models trained on it (\cref{sec:utility}), and the cost of
producing it (\cref{sec:cost}). Memorisation of the training data is part of fidelity, but it is ranked
separately. All series have the same length $K \cdot T$ and are evaluated in real units by the same code.

\subsection{Fidelity}
\label{sec:fidelity}
\paragraph{Three real references.} Every series is scored against three real references, as shown in
\cref{fig:fidelity}. Val is the primary reference: it follows train in time, and it has not been used by any
method. Train shows how well a method reproduces the data it was fitted on; every method has seen it, and
some methods take reference statistics from it. OOT is the latest part of the series and measures fidelity
under a shift in time. A method that copies train well but drifts away from the later data scores well
against train and worse against val and OOT. Scoring against all three references therefore separates the two cases.

\paragraph{Real-versus-real reference.} As a real-data reference, the OOT series is evaluated against the
other reference splits, val and train, using length-matched comparisons: it is scored as if it were a
synthetic series (the pseudo-method \texttt{real}). OOT is shorter than $K \cdot T$, and many metrics depend
on the length of the series; therefore, every method is also scored, on its first seed, on its first
$|\mathrm{OOT}|$ points, and this score is compared with that of the real series at equal length. Since OOT is the latest part of the series, a shift in time can place it farther from val or train than a
synthetic series that reproduces them; a method can therefore score better than the real series on a metric.

\fig{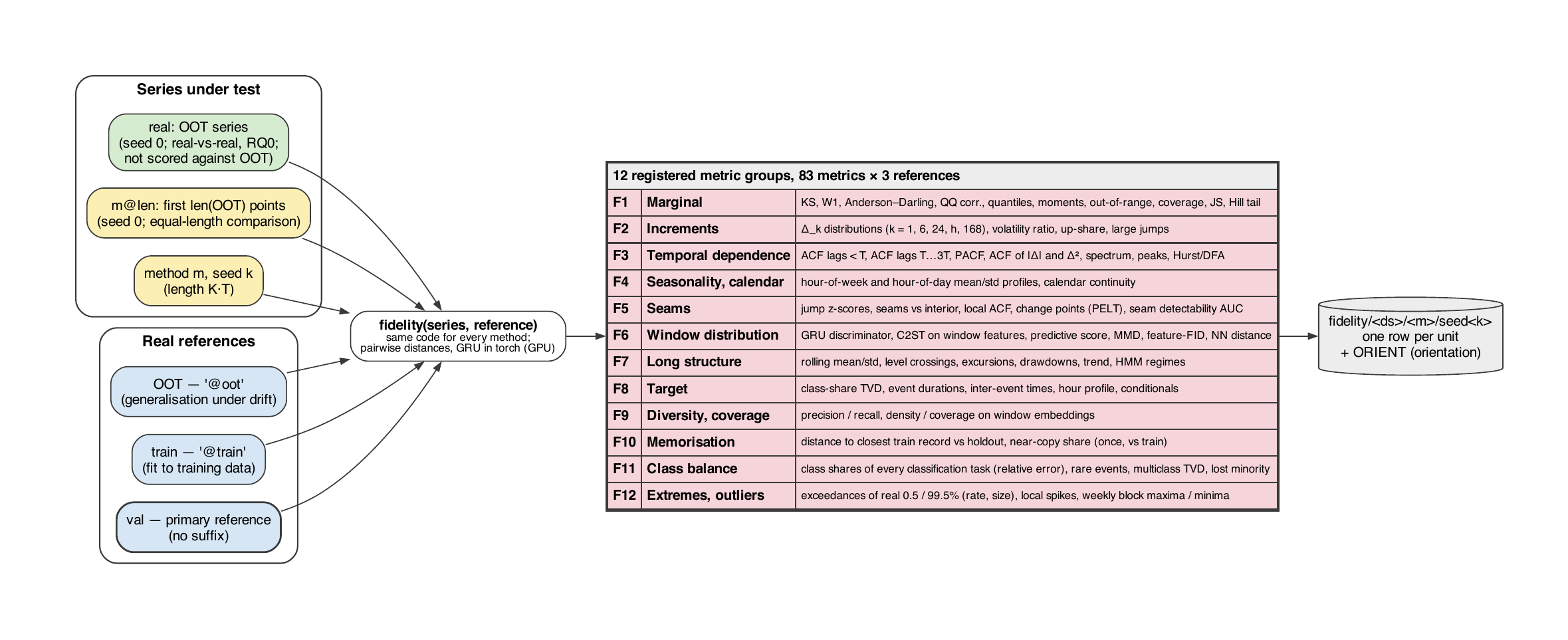}{fidelity}{1}{Fidelity evaluation. Every generated series,
the real OOT series (real-versus-real reference), and length-matched truncations of the generated series are scored by
the same code against three real references: val (primary), train and OOT (the real OOT series against val and train). The metrics are organised
in 12 groups (83 metrics); memorisation (F10) is computed once, against train.}

\paragraph{Metric groups.} The 83 metrics are organised in 12 groups. \Cref{tab:fidelity} lists the groups,
and the full list of metrics is given in \cref{app:metrics}. Groups F1--F4 cover the properties that most
evaluations of synthetic time series check:
\begin{itemize}
  \item the marginal distribution of values (F1);
  \item the distribution of increments at horizons from one hour to one week (F2);
  \item temporal dependence through the autocorrelation, the spectrum and detrended fluctuation analysis (F3);
  \item the daily and weekly profiles (F4).
\end{itemize}
Group F6 compares the distribution of windows with the standard measures of the literature: a post-hoc
recurrent discriminator~\citep{yoon2019timegan}, a classifier two-sample test~\citep{lopezpaz2017c2st}, a
predictive score, the MMD \citep{gretton2012mmd} for windows of length $T$ and $2T$, a Fréchet distance
between window features, and the nearest-neighbour distance from synthetic to real windows. Group F9
measures diversity and coverage through precision, recall, density and coverage~\citep{naeem2020prdc}.

Three groups examine properties of the assembled or generated series. Group F5 measures the
\emph{seams}: the distribution of the jumps at the seams against the real increments and against the
increments elsewhere in the same series, the local autocorrelation around seams, the share of detected
change points that fall on a seam, and how well a classifier distinguishes a snippet around a seam from a
snippet elsewhere. For a series without seams these metrics are undefined, except the rate of change points,
which compares the series as a whole. The share of change points at seams depends on how dense the seams
are: with seams every 64 to 1\,024 points, 0.6 to 11\,\% of the positions lie within three points of a seam, so
the share is compared with this chance level in a supplementary check (\cref{sec:results-checks,app:seams}). Group F7 measures \emph{structure longer than a window}: the
distributions of rolling means and standard deviations over one week and one month, the rates of crossing
the real quantile levels, the lengths of excursions above and below them, drawdowns from the rolling weekly
maximum, the trend, and the durations of high-volatility regimes of a two-state hidden Markov model. F3 also includes the autocorrelation at lags from $T$ to $3T$, beyond the length of a training window.

The remaining groups are related to what the downstream tasks depend on. Group F8 measures the structure of
the events of task U2.1, an increase over 24 hours above the 0.90 quantile of train: their share, durations,
inter-arrival times, weekly profile, and conditional rates given the recent past. Forecasting such tail
events is a common classification problem, and a synthetic series that reproduces the marginal distribution
can still get the timing and clustering of the events wrong. Group F11 measures the class balance over every
classification task, with relative errors, so that a rare class weighs as much as a frequent one. Group F12
measures outliers and extremes: exceedances of the real 0.5\% and 99.5\% quantiles, local spikes, and the
distributions of weekly maxima and minima.

\paragraph{Memorisation.} Group F10 measures how close the synthetic windows are to the training windows. It
consists of two metrics: the absolute logarithm of the ratio of the median distance to the closest training
window (DCR) of the synthetic windows to that of the real val windows, and the share of synthetic windows
that are closer to train than 95\% of the val windows (near copies). The first metric is unsigned; the signed ratio, which
separates windows closer to train than the val windows from windows farther from it, is reported in
\cref{app:memorisation}. This group is computed once, against train, and is ranked separately
from fidelity, since a series can be faithful by copying train, and a ranking that mixes the two would
reward such a series.

\paragraph{Conventions.} Each metric has a fixed orientation (lower or higher is better). Metrics that
compare levels, windows or durations are computed after z-scoring both series with the mean and standard
deviation of the reference. Deviations of a scale (standard deviations, rates, durations) are reported as
$|\log(\text{synthetic}/\text{real})|$, and accuracies of a classifier as $|\mathrm{AUC} - 0.5|$, so that 0
is ideal for both. Windows for F6, F9 and F10 are cut from the z-scored series with length $T$ and step
$T/4$. The window features of F6 and F9 are 16 simple statistics of a window, such as moments, quantiles,
lag-1 and lag-24 autocorrelation, and the positions of the maximum and the minimum.

\begin{table}[t]
  \caption{Fidelity metric groups. $n$: number of metrics in the group. Every group except F10 is computed
  against val, train and OOT.}
  \label{tab:fidelity}
  \centering
  \small
  \begin{tabularx}{\linewidth}{lrlX}
    \toprule
    Group & $n$ & What it measures & Examples \\
    \midrule
    F1  & 15 & Marginal distribution of values  & KS, Wasserstein-1, quantile errors, Hill tail index \\
    F2  &  9 & Increments                       & KS of $\Delta_k$, $k \in \{1, 6, 24, 168\}$; volatility ratio \\
    F3  & 10 & Temporal dependence              & ACF within and beyond a window, PACF, spectrum, DFA \\
    F4  &  5 & Seasonality and calendar         & weekly and daily profiles, calendar continuity \\
    F5  &  8 & Seams                            & jumps at seams, change points at seams, seam detector \\
    F6  &  7 & Distribution of windows          & discriminative score, C2ST, MMD, feature FID \\
    F7  &  8 & Long-range structure             & rolling statistics, level crossings, drawdowns, regimes \\
    F8  &  6 & Tail events of task U2.1         & event share, durations, inter-arrival times \\
    F9  &  4 & Diversity and coverage           & precision, recall, density, coverage \\
    F10 &  2 & Memorisation (ranked separately) & DCR ratio, share of near copies \\
    F11 &  4 & Class balance                    & class-share errors over all classification tasks \\
    F12 &  5 & Outliers and extremes            & exceedance rates, spikes, weekly block maxima \\
    \bottomrule
  \end{tabularx}
\end{table}

\subsection{Utility: TSTR versus TRTR}
\label{sec:utility}
Utility measures whether a model trained only on the synthetic series performs on real data as well as the
same model trained on real data. The protocol is shown in \cref{fig:utility}. Each model is trained in two
regimes: on the synthetic series of a method (TSTR, train on synthetic, test on real) and on the real
training series (TRTR). Both are tested on the same real rows of val and of OOT, and synthetic data never
enter a test set. The closer TSTR is to TRTR, the more of the information useful to the model the
synthetic series carries. The benchmark measures the quality of the synthetic
data, not its value as augmentation; therefore, regimes that mix real and synthetic training data are not
part of the benchmark.

\fig{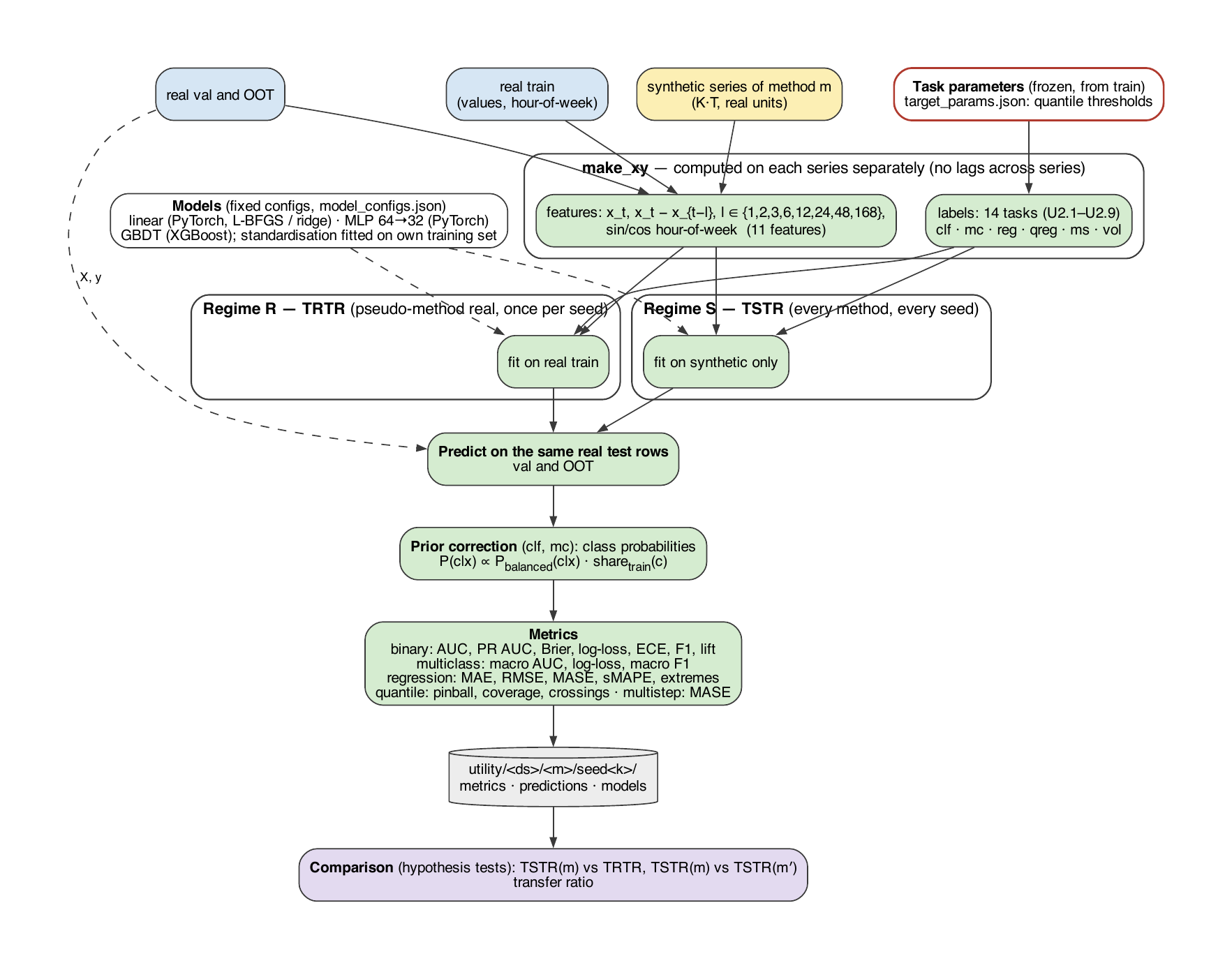}{utility}{1}{Utility protocol. Features and labels are computed
separately on each series: real train, the synthetic series, and the real test parts. Thresholds are
frozen on real train. Each model is trained either on real train (TRTR) or on the synthetic series
only (TSTR) and evaluated on the same real val and OOT rows. Synthetic data never enters a test set.}

\paragraph{Tasks.} The benchmark contains 14 tasks in nine groups U2.1--U2.9: eight classification tasks
and six regression tasks. \Cref{fig:tasks} shows the tasks, and their details are given in
\cref{app:utility}. The binary tasks ask whether the change over a horizon $h$ exceeds a quantile of the
changes in train:
\begin{itemize}
  \item the upper tail at $h \in \{1, 6, 24, 168\}$ with $q = 0.90$;
  \item rare events with $q \in \{0.95, 0.99\}$;
  \item the lower tail with $q = 0.10$.
\end{itemize}
A three-class task combines the two tails. The regression tasks forecast the value at
$h \in \{1, 24, 168\}$, the quantiles 0.1, 0.5 and 0.9 at $h = 24$, the whole next day as a vector, and the
volatility of the next day. All thresholds are fixed on real train before any run
(\cref{sec:preprocessing}).

\paragraph{Features and labels.} Every task uses the same 11 features: the value, its differences at lags
1, 2, 3, 6, 12, 24, 48 and 168, and the hour of the week on a circle. Features and labels are computed on
each series separately, so no lag crosses the boundary between a synthetic and a real series. The labels of
a synthetic series are computed from the series itself with the thresholds of real train. Thus, a method
that produces too few tail events also produces too few positive labels.

\fig{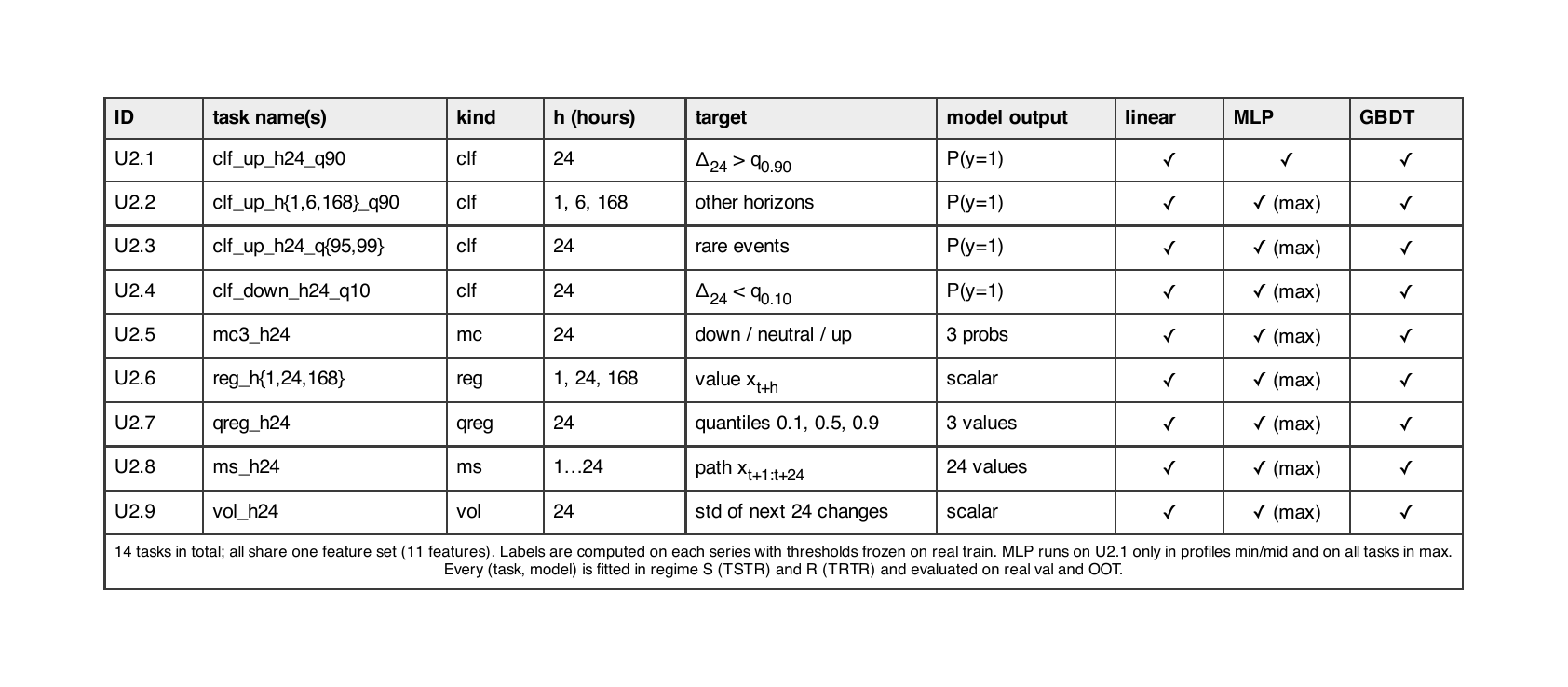}{tasks}{1}{The 14 downstream tasks (U2.1--U2.9) and the models
trained on each. All tasks use the same 11 features. Labels are functions of the series, with
thresholds frozen on real train.}

\paragraph{Models.} Three model families are trained:
\begin{itemize}
  \item linear models (logistic or softmax regression, ridge regression, linear quantile regression);
  \item gradient-boosted trees (XGBoost);
  \item a small fully connected network.
\end{itemize}
Their hyperparameters are fixed in advance in a configuration file and are not tuned on the data. Profile
\texttt{max} adds a grid search on the tail of each regime's own training data, so that neither val nor
OOT is used for tuning. To limit the cost, profiles \texttt{min} and \texttt{mid} run the network on task
U2.1 only. Classifiers are trained with balanced class weights, and their probabilities are mapped back to
the class shares of their own training data. This makes it possible for the Brier score, the log-loss and
the calibration error to measure calibration.

\paragraph{Metrics.} Classification is scored by ROC AUC, PR AUC, the Brier score, the log-loss, the
expected calibration error, F1 and lift; regression is scored by MAE, RMSE, MASE, sMAPE and the error on
extreme targets; quantile regression is scored by the pinball loss, the coverage of the 10--90\% interval
and the rate of crossing quantiles. MASE, the pinball loss and the error on extreme targets are divided by
the error of the seasonal naive forecast (lag 168) on real train. Each task type has one primary metric for
the rankings: AUC for binary classification, log-loss for the three-class task, MASE for regression,
volatility and the multi-step forecast, and the pinball loss for quantiles. The \emph{transfer ratio}
relates TSTR to TRTR on the same test rows; a value of 1 means that the synthetic series is as useful as the
real one.

\subsection{Cost}
\label{sec:cost}
The cost of a series is recorded per (dataset, method, seed). It includes the wall-clock and CPU time of
training the generator, sampling and assembly, the peak GPU memory, and the increase of resident memory.
When one generator model is shared by several seeds, its training time is charged to every seed, so that a
method whose model is trained once is not favoured. For the pretrained forecasters, model loading counts as
training time. Every record stores the hardware and library versions, and the comparison of costs is valid
only within one type of hardware; all costs in this paper come from the main run on one NVIDIA H200
(\cref{sec:runs}).

These measurements are the input of the statistical analysis described in \cref{sec:statistics}.

\section{Statistical analysis and research questions}
\label{sec:statistics}

The statistical analysis is carried out on the benchmark tables without generating anything; its
protocol and the date at which it was frozen are given in \cref{sec:protocol}. The pipeline of the analysis is shown in \cref{fig:stats}. It consists of three
parts:
\begin{itemize}
  \item rankings of every method (\cref{sec:rankings});
  \item research questions about families of methods and about the benchmark itself (\cref{sec:rq});
  \item a small pre-specified confirmatory family that carries the main conclusion
  (\cref{sec:confirmatory}).
\end{itemize}
All parts use the same rules for pairing, noise and verdicts (\cref{sec:rules}). Every method is treated
in the same way: the family of a method (assembly or long-series generator) is read from its
description, so a method added to the benchmark enters the rankings and the questions without changes to
the analysis.

\fig{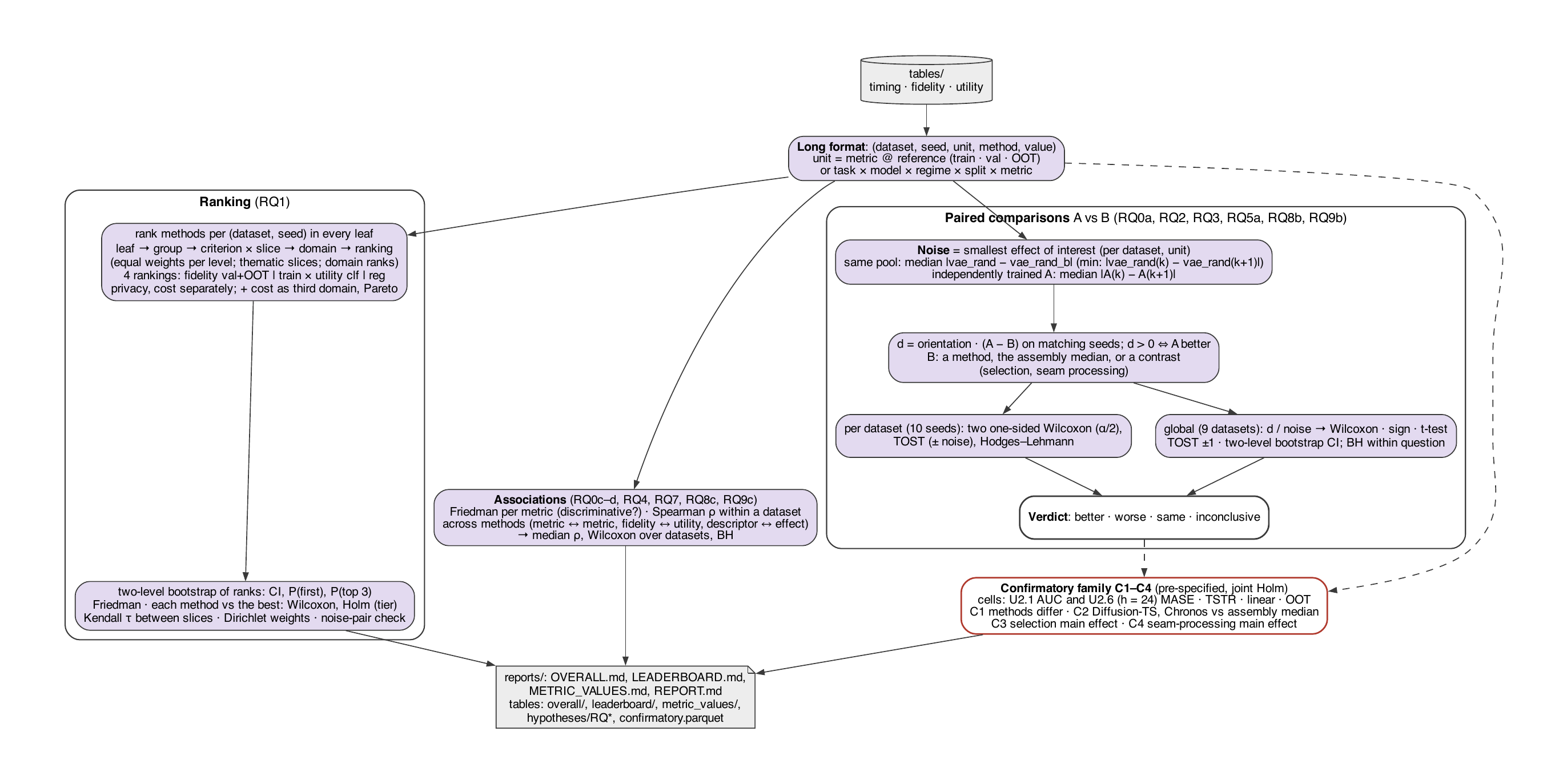}{stats}{1}{From benchmark tables to conclusions. The
long-format tables feed three analyses applied to every method: a ranking of every method in every
slice, paired comparisons A vs B measured in units of an empirical noise level, and associations
across methods. A pre-specified confirmatory family (C1--C4) is corrected jointly.}

\Cref{fig:rq} shows the research questions and their relation to the confirmatory family.

\fig{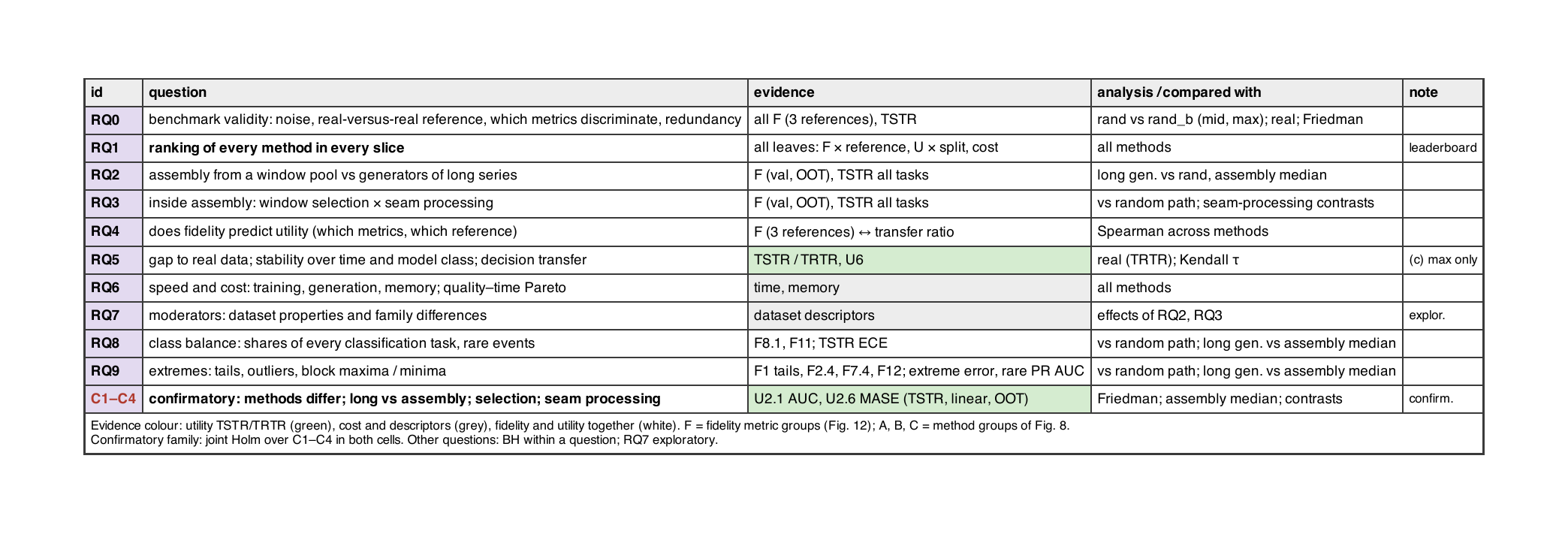}{rq}{1}{The research questions of the benchmark. RQ0 validates the
benchmark itself, RQ1 ranks every method in every slice, RQ2--RQ9 compare families of methods and
properties of the synthetic data. C1--C4 (red) form the pre-specified confirmatory family.}

\subsection{Pairing, noise and verdicts}
\label{sec:rules}
\paragraph{Units of replication.} Within a dataset, the replicates are the 10 seeds. A seed fixes the
pool, the random paths and the training of the generators and the models, so methods with the same seed
form a pair. Across datasets, the unit is the dataset: a global conclusion is drawn from the means over
seeds of the nine datasets, since the seeds of one dataset are not independent. In profile \texttt{min}, the VAE and Diffusion-TS are trained once per dataset; therefore, a per-dataset
conclusion is conditional on the trained model, while the global conclusion generalises over datasets.

\paragraph{Direction.} A comparison of $A$ with $B$ is carried out on one unit of analysis: a fidelity
metric against one reference, or a task $\times$ model $\times$ split of TSTR. The paired difference is
calculated as $d = o \cdot (A - B)$, where the orientation $o$ is $+1$ if higher is better and $-1$
otherwise. That is, $d > 0$ always means that $A$ is better.

\paragraph{Noise as the smallest effect of interest.} A difference is counted only if it exceeds the
\emph{noise}, that is, how much the score of a method changes for reasons that have nothing to do with
the method. For each dataset and unit, the noise of comparisons on one pool is calculated as the median
over seeds of $|\texttt{vae\_rand}(k) - \texttt{vae\_rand}(k{+}1)|$, the difference between random
assemblies of neighbouring seeds. This difference contains the sampling variance of the pool. The
noise is therefore large, which makes superiority harder to establish and the equivalence region wider;
equivalence is defined relative to it, that is, to the stochastic variability of the benchmark, not to a
tolerance of an application. For a comparison with an independently trained method, such as a
long-series generator, the noise is the same between-seed spread of whichever of the two methods has the
larger one. TimesFM and Timestral have no spread of their own (\cref{sec:methods-c}), so in their
comparisons the margin is set by the spread of the method they are compared with. The noise is fixed
before any comparison and serves two purposes~\citep{lakens2017equivalence}: it is the threshold of
practical significance and the margin of equivalence. Global effects are expressed in noise units,
$z = d / \text{noise}$, which makes metrics and datasets of different scales comparable.

\paragraph{Verdicts and tests.} The following tests are used.
\begin{itemize}
  \item Per dataset, two one-sided Wilcoxon signed-rank tests~\citep{wilcoxon1945} over seeds, each at
  level $\alpha/2$, test the hypotheses ``$A$ is better'' and ``$A$ is worse''; TOST
  \citep{schuirmann1987tost} with two one-sided Wilcoxon tests is used for equivalence within $\pm$noise; the
  effect is the Hodges--Lehmann estimate.
  \item Globally, the tests are applied to the dataset means: Wilcoxon, the sign test and the $t$-test,
  a two-level bootstrap (datasets, then seeds) for the confidence interval, and TOST within $\pm 1$
  noise unit. The conclusion is robust when the tests agree.
\end{itemize}
The verdict is \emph{better} or \emph{worse} if the one-sided test is significant after correction and
the effect exceeds the noise (per dataset, the Hodges--Lehmann estimate; globally, the median $z > 1$). The
verdict is \emph{same} if equivalence is shown by TOST, and \emph{inconclusive} otherwise. ``Same'' means demonstrated equivalence, not the absence of a significant difference. If the
noise of a unit on a dataset is zero, the margin is undefined: in this case the per-dataset verdict is
inconclusive and the dataset is left out of the global test.

\paragraph{Intervals.} The interval of a global effect is a two-level percentile bootstrap with 2\,000
replicates: datasets are drawn with replacement, and then seeds within each drawn dataset; the paired
difference $d$ of two methods is formed on the same dataset, seed and unit before resampling, so the pairing
of methods and seeds is kept. The interval of a mean rank uses the same scheme with 1\,000 replicates, the
ranks being formed within each dataset and seed. A method without sampling, such as TimesFM or Timestral,
contributes no variation of its own, and in profile \texttt{min} the seeds of the VAE and of Diffusion-TS
vary the sampling, not the training (\cref{sec:limitations}). The test rows are fixed, so the uncertainty
within a series is not resampled. The intervals are not adjusted for multiplicity.

\paragraph{Missing utility cells.} A synthetic series may be unable to train a task. For example, a
smooth median forecast may never move above the threshold, so its labels have one class. Leaving such a
cell out would rank the method only on the datasets where it succeeded. Instead, the cell enters as a
model that learned nothing: each metric takes the worse of two values, the chance level and the worst
value of any method on the same dataset, seed, task, model and split. The value of an imputed cell
therefore depends on the other methods of the block, so a poor method can lower it. The chance levels of
every metric are listed in \cref{tab:app-chance}; a metric without one takes the worst value of its block.
A method excluded for leakage (\cref{sec:datasets}) is not imputed; it
is absent on that dataset.

\paragraph{Multiplicity.} Corrections for multiple testing are applied at three levels:
\begin{itemize}
  \item the confirmatory family C1--C4 (\cref{sec:confirmatory}) is corrected jointly by
  Holm~\citep{holm1979}, and only this family gives the main conclusion;
  \item secondary questions (RQ0, RQ2--RQ6, RQ8, RQ9) are corrected by Benjamini--Hochberg
  \citep{benjamini1995} within each question, both for the global $p$-values and for the per-dataset
  verdicts of each question and dataset; they give a map of where a family of methods is better, worse or
  the same, with the false discovery rate controlled;
  \item exploratory analyses (RQ7) report $p$-values, and their results are stated as hypotheses.
\end{itemize}
With 10 seeds, the smallest one-sided Wilcoxon $p$-value is about 0.001, so a per-dataset verdict is
possible. With nine datasets it is about 0.002, so a global verdict after correction requires a large and
consistent effect. Thus, the main lever of power is the number of datasets.

\subsection{Rankings}
\label{sec:rankings}
RQ1 asks how the methods rank and with what uncertainty. The seam-processing variants are not ranked,
since in profile \texttt{min} they exist only on the path of zseries; ranking them would let one
assembler enter four times and every other assembler once. Image quilting selects windows and cuts the
seams in one method and is always ranked. Methods left out of the rankings take part in every other
question.

\paragraph{From leaves to rankings.} A \emph{leaf} is one fidelity metric against one reference, or one
task $\times$ model $\times$ metric $\times$ split of TSTR. Within a leaf, the methods are ranked
separately for each (dataset, seed), since methods of one seed share the pool and the test rows; ties
receive the average rank. The ranks are then averaged along a fixed hierarchy with equal weights at every
level, so that a group with many metrics does not outweigh a group with few:
\begin{itemize}
  \item leaves form groups;
  \item groups form criteria (\cref{tab:criteria}), and each criterion is averaged over its references or
  test splits;
  \item criteria form the fidelity and utility domains.
\end{itemize}
A ranking is the mean of the two domains. Memorisation (F10) and cost are ranked separately and do not
enter the rankings of fidelity and utility. \Cref{tab:criteria} lists the criteria and the metric groups that form them. A metric that
is undefined for a method (for example, the seam metrics of a series without seams) is not imputed: the
method is not ranked in that leaf, and a criterion without any defined leaf is left out of the method's
domain, whose remaining criteria are averaged. In the overall leaderboard this concerns the seam criterion
of TimesFM and Timestral; \cref{app:seams} gives the leaderboard without the seam criterion for every
method.

\begin{table}[t]
  \caption{Ranking criteria. Fidelity criteria are averaged over their references, utility criteria
  over the two test splits. The calendar-continuity metrics F4.4--F4.5 are left out, since the
  pretrained forecasters take the real calendar (\cref{sec:methods-c}).}
  \label{tab:criteria}
  \centering
  \small
  \begin{tabularx}{\linewidth}{llX}
    \toprule
    Domain & Criterion & Content \\
    \midrule
    Fidelity & distribution   & F1 values, F2 increments, F12 outliers and extremes \\
             & dependence     & F3 temporal dependence, F4 seasonality \\
             & windows        & F6 distribution of windows, F9 diversity and coverage \\
             & long structure & F7 \\
             & target         & F8 tail events of U2.1, F11 class balance \\
             & seams          & F5 \\
    Utility  & classification & U2.1--U2.5, TSTR on val and OOT \\
             & regression     & U2.6--U2.9, TSTR on val and OOT \\
    \midrule
    Separate & memorisation   & F10, against train \\
             & cost           & total time, training, generation, memory \\
    \bottomrule
  \end{tabularx}
\end{table}

\paragraph{The overall leaderboard and four rankings.} The headline ranking is the \emph{overall
leaderboard}. Its fidelity domain uses a core set of one or two canonical metrics per criterion, each
averaged over train, val and OOT: for example, the Wasserstein distance of values, the power spectrum, the
discriminative score and the MMD of windows, the class shares, and the detectability of seams. Its utility
domain averages classification and regression on both test splits. The core set was fixed by what the
metrics measure, not by how well they separated the methods: choosing the discriminative metrics on the
same data as the ranking would select noise that happens to favour some method. Within a leaf, the ranks
are noise-aware: the rank of a method is one plus the number of methods that beat it by more than the noise
of the leaf on that dataset, plus half the number of other methods within the noise, so near-identical
methods share a place instead of pushing the others apart. The margin is the same for every pair of methods
in a leaf, no groups are formed, and the ranks of a block sum to $k(k+1)/2$ for $k$ methods; when a method is
missing on a dataset because of leakage, the ranks of the others are rescaled to $1, \dots, k$. Closeness
within the noise is not transitive, so a method can tie with two others that do not tie with each other
(\cref{app:ties}). Four further rankings combine fidelity against val and OOT, or against train,
with utility in classification or in regression, over every fidelity metric rather than the core set.
These rankings show whether the order depends on the reference and on the task type.

\paragraph{Uncertainty and robustness.} A two-level bootstrap of the mean ranks (datasets, then seeds)
gives an interval for each rank and the probability of being first and in the top three. The Friedman test
\citep{friedman1937} with datasets as blocks tests whether the methods differ at all. The \emph{top tier}
consists of the best method and every method that is not significantly worse than it by a one-sided
Wilcoxon test over datasets with Holm correction~\citep{benavoli2016posthoc}. The top tier is thus a
non-rejection set, not an equivalence class: with limited power, a method can remain in it without being
close to the leader. The comparisons are made with the observed leader and are not adjusted for its
selection (\cref{app:tier}). The Nemenyi critical
difference~\citep{demsar2006} is reported for reference only, since it grows with the number of methods.
The robustness of the leaderboard is checked by variants fixed in advance, each of which is compared with
the leaderboard by Kendall's $\tau$:
\begin{itemize}
  \item ordinary average ranks;
  \item fidelity against train only, and against val and OOT only;
  \item every fidelity metric instead of the core set;
  \item only the datasets where every method is ranked;
  \item a field of one method per generator, so that many variants of one generator cannot move the
  others;
  \item leaving out one dataset at a time;
  \item without the seam detector and the discriminative score against val, the two metrics whose
  separation of real data from a random assembly could not be checked;
  \item 2\,000 random weights of the criteria (Dirichlet(1) within each domain).
\end{itemize}

\paragraph{Next to the rank.} Ranks show which method is better, but not by how much. Therefore, each core
metric is also reported on a normalised scale on which the random assembly is 0 (of the method's own pool;
of the VAE pool for a long-series generator) and the real data are 1; a value above 1 means closer to the
reference than the real data. Cost is reported next to each ranking as a Pareto front of rank against
median time and as the same ranking with cost as a third domain. Memorisation does not change the rank
but is flagged: a method is flagged when its share of windows very close to train is at least twice that
of the real OOT series, significantly over datasets.

\subsection{Research questions}
\label{sec:rq}
\paragraph{RQ0: validity of the benchmark.} This question asks whether the benchmark measures what it
should. It consists of four checks:
\begin{itemize}
  \item (a) two random assemblies of one pool should be equivalent on every metric; this check needs the
  second random path and is carried out in profiles \texttt{mid} and \texttt{max} only;
  \item (b) the real OOT series, scored as a method, is compared with the methods on each fidelity metric
  at equal length (\cref{sec:fidelity}); a metric on which the real OOT reference does not outperform
  most synthetic methods is flagged for inspection; the cause can be a shift between OOT and the
  reference, sampling variability or the sensitivity of the metric;
  \item (c) a metric is \emph{discriminative} if the Friedman test across methods is significant after
  correction and the spread of the methods exceeds the noise. The check covers 297 units: 79
  fidelity metrics against each of the three references (the 83 metrics less the two calendar-continuity
  metrics, which are left out of every question, and the two memorisation metrics), the two memorisation
  metrics once, against train, and 58 utility units (each of the 14 tasks with two downstream models on val and
  OOT, and task U2.1 with the third model as well);
  \item (d) pairs of metrics whose rank correlation across methods is at least 0.9 in absolute value
  duplicate each other, and the agreement of a metric against different references shows how much the
  conclusion depends on the reference.
\end{itemize}

\paragraph{RQ2: assembly versus long-series generation.} Each long-series generator is compared with the
random path of the VAE pool and with the \emph{assembly median}, the median over all assembly methods of
the pool for each dataset, seed and unit. The median is used rather than the best assembler, because the
best one would be selected on the same data on which it is compared.

\paragraph{RQ3: within assembly.} This question asks whether the selection and order of the windows and the
processing of the seams move the series towards the real data, each against its own control, when a series
is assembled from one pool. Each assembler is compared
with the random path, and each seam-processing variant with the same path without processing. The main
effect of each processing is averaged over the paths that have every processing.

\paragraph{RQ4: does fidelity predict utility?} For each fidelity metric and reference and each utility
criterion, the Spearman correlation across methods within a dataset between the metric and the transfer
ratio is calculated and tested over datasets. A metric is called \emph{predictive}, the term of the protocol, when
this correlation is positive over the datasets after correction. The correlation is an association, not a
causal statement, across the methods of the benchmark, several of which share a pool or a generator; whether
a metric selects a generator outside the benchmark is not tested.

\paragraph{RQ5: gap to real data and stability.} This question consists of three parts:
\begin{itemize}
  \item (a) the transfer ratio of each method and criterion is tested against 1, the value at which the
  synthetic series trains the models as well as the real one. For ROC AUC, transfer is measured as skill
  above chance: $(\mathrm{AUC}_{\mathrm{TSTR}} - 0.5)/(\mathrm{AUC}_{\mathrm{TRTR}} - 0.5)$, and is
  undefined when $\mathrm{AUC}_{\mathrm{TRTR}} \leq 0.52$. For lower-is-better losses, transfer is
  $\mathrm{loss}_{\mathrm{TRTR}}/\mathrm{loss}_{\mathrm{TSTR}}$; for other higher-is-better scores it is
  $\mathrm{score}_{\mathrm{TSTR}}/\mathrm{score}_{\mathrm{TRTR}}$. Absolute scores are reported in \cref{tab:app-absolute}.
  \item (b) Kendall's $\tau$ between the rankings on val and on OOT, and between model classes, shows
  whether the order of the methods survives a shift in time and a change of the model;
  \item (c) in profile \texttt{max} only, a grid of model configurations is ranked on real and on
  synthetic data, to test whether a choice made on synthetic data transfers to real data.
\end{itemize}

\paragraph{RQ6: cost.} The following characteristics are reported: the median time of each method over
datasets, the time relative to the fastest method, the separate costs of training and of generating a new
series, the peak memory, and the slope of log time on log $K$, which shows how the cost grows with the
length of the series. The slope needs different $K$ and is not available in profile \texttt{min}, where
$K = 30$ for every dataset. Generation is reported separately from training, because a generator is
trained once and sampled many times.

\paragraph{RQ7: moderators (exploratory).} The dataset effects of RQ2 and RQ3 are correlated with
descriptors of the datasets: autocorrelation at 24 and 168 hours, the kurtosis of the increments,
volatility clustering, trend and the share of missing values. With nine datasets, the results of this
question are stated as hypotheses only.

\paragraph{RQ8 and RQ9: class balance, outliers and extremes.} Two thematic slices collect metrics from
different groups:
\begin{itemize}
  \item class balance: the class shares of every task and the calibration of the TSTR classifiers;
  \item outliers and extremes: tail quantiles, the tail index, large jumps, the metrics of F12, the
  regression error on extreme targets and the PR AUC of rare events.
\end{itemize}
Inside the fidelity aggregates, these metrics are averaged with many others. Therefore, each slice is
ranked on its own, the families of methods are compared on it as in RQ2 and RQ3, and the fidelity metrics
of the slice are correlated with the corresponding utility.

\subsection{Confirmatory family}
\label{sec:confirmatory}
The main conclusion of the paper rests on ten tests in two cells fixed in advance: a classification cell,
task U2.1 with ROC AUC, and a regression cell, task U2.6 at $h = 24$ with MASE. Both cells are TSTR with
the linear model, tested on OOT, the split furthest from train. Each cell contains the following comparisons:
\begin{itemize}
  \item \textbf{C1} asks whether the methods differ at all: an omnibus Friedman test over the ranked methods with
  datasets as blocks. The design is complete: a method excluded for leakage on any dataset drops out of the
  test, and every dataset stays. In both runs these are Chronos and TimesFM, so C1 compares the other 22
  ranked methods of the main run, those of the zEDGE pool and zEDGE\_TS included, and the other 11 of the
  replication run, on all nine datasets (\cref{app:c1});
  \item \textbf{C2} compares Diffusion-TS and Chronos with the assembly median, as in RQ2; the two
  generators are fixed in advance, and other long-series generators are compared in RQ2 but do not enter
  the family;
  \item \textbf{C3} is the main effect of window selection: the mean of every ranked assembler against the
  random path;
  \item \textbf{C4} is the main effect of seam processing: every processing against no processing, on the
  paths that have every processing (in profile \texttt{min}, the path of zseries). In this profile the
  contrast is therefore the effect of seam processing on the zseries path of the VAE pool.
\end{itemize}
The ten two-sided $p$-values ($2 + 4 + 2 + 2$) are corrected together by Holm. A test that cannot be
computed enters with $p = 1$, so it cannot weaken the correction. C1 gives the verdict \emph{differ} or
\emph{inconclusive}. C2--C4 give \emph{better} or \emph{worse} if $p_{\mathrm{Holm}} < \alpha$ and the
median effect exceeds one noise unit, and \emph{same} if TOST is significant after Holm. This joint rule of
statistical and practical significance is referred to below as the pre-specified rule.

\subsection{Several pool generators}
\label{sec:two-pools}
The contrasts within assembly (RQ3, C3, C4) and the assembly median of RQ2 and C2 are computed on
one pool, the VAE pool, whose random path gives the noise level. The methods of the zEDGE pool
enter the rankings and all other questions in the same way as any other method. As an exploratory check
that was not pre-specified, the same analysis is repeated with the zEDGE pool in place of the VAE pool
(\cref{sec:results-rq}).

\subsection{Protocol and amendments}
\label{sec:protocol}
The analysis protocol (\texttt{docs/HYPOTHESES.md} and the code of the analysis) was frozen at commit
\texttt{7fea7530} on 21 October 2026, before the main run was carried out (24 to 26 October 2026); the replication run was completed
before the freeze, on the same day. The analysis is therefore pre-specified with respect to the main run. The protocol fixes the core metrics, the
contrasts of the research questions, the families of multiple comparisons, the exclusion rules and the
algorithms for noise, ties and tiers; \cref{tab:app-protocol} lists them with their place in the protocol.
Changes made after the freeze are listed in \texttt{docs/PROTOCOL\_AMENDMENTS.md} and are exploratory: the
reading of the leaderboard by tier rather than by place, which was adopted after the two runs were compared;
a correction to the rank bootstrap, in which tied methods now share a place; the wording of RQ3; and the supplementary analyses of
\cref{sec:results}, which are collected in \cref{sec:results-checks} or marked \emph{supplementary} where
they are reported.

\Cref{sec:software} describes the software that implements the benchmark and this analysis.

\section{Software and reproducibility}
\label{sec:software}

MarathonTS is released as a Python package with the code of every open method, the configuration of
the runs and the published tables. This section describes how a run is organised and how the benchmark is
extended; the architecture is described in more detail in \cref{app:software}.

\paragraph{Stages.} A run consists of a sequence of stages, shown in \cref{fig:pipeline}:
\begin{itemize}
  \item \texttt{raw} converts the source files to Parquet;
  \item \texttt{prepare} builds the hourly series and the splits;
  \item \texttt{pools} trains the pool generators and samples the pools;
  \item \texttt{generate} produces the series of every method;
  \item \texttt{fidelity} and \texttt{utility} evaluate them;
  \item \texttt{collect} merges the results into the published tables, one row per (dataset, method, seed)
  for fidelity and cost and one row per task, model, regime and split for utility.
\end{itemize}
The analysis of \cref{sec:statistics} is a separate package that only reads these tables. It runs on a CPU
and writes the rankings, the answers to the research questions and the confirmatory family as Parquet
tables and Markdown reports. Every stage writes one file per unit of work (\cref{fig:artifacts}). This allows the work to be split across machines by dataset and seed, and an error in one unit to
be logged without stopping the run.

\paragraph{Environments.} The pool generator and the long-series generators have conflicting
dependencies; therefore, each of them runs in its own environment with pinned requirements
(\cref{fig:environments}). The VAE runs in a TensorFlow environment without PyTorch; Diffusion-TS, Chronos,
TimesFM and Timestral each run in a PyTorch environment; assembly, evaluation and the analysis run in a
common core environment. The environments exchange data only through the Parquet files of the pools and
the series. A Docker image and Colab notebooks reproduce the environments.

\paragraph{Profiles.} Three run profiles fix everything that trades accuracy for compute; they are shown in
\cref{fig:profiles}.
\begin{itemize}
  \item Profile \texttt{min}, which is used for the main run, takes the last 10\,000 hours of each dataset,
  trains one VAE and one Diffusion-TS model per dataset, uses light configurations of the downstream
  models and can run on a single GPU, for example in Colab.
  \item Profile \texttt{mid} retrains the VAE on every seed and adds the second random path.
  \item Profile \texttt{max} uses the full series, retrains the generators on every seed, applies every
  seam joiner to every assembler except the random path and image quilting, and tunes the downstream
  models on the tail of each regime's own training data.
\end{itemize}
Each profile writes to its own folders, and its task thresholds are computed on its own training split.

\paragraph{Incremental recomputation.} Every unit of work stores a fingerprint: a hash of its data
settings, the settings of its components, the configuration files it depends on, and the fingerprints of
the units it is made from (\cref{fig:fingerprints}). A unit is skipped only if its outputs exist and its
fingerprint matches. Therefore, a changed setting recomputes exactly the units downstream of it. For
example, changing the settings of one assembler recomputes its series and their evaluation and keeps the
pools, the other methods, the labels and the TRTR models. Fidelity and utility are fingerprinted per
metric group and per task and model, so an added metric group, task or model computes only its own cells.
A long run can be stopped and resumed, and the cached results always correspond to the current
configuration.

\paragraph{Data integrity.} The sources of the datasets are downloaded from their original locations and
may change after a run; for example, Bitcoin is re-uploaded daily. The repository stores a hash of the
content of every prepared series (for ETT, of the series of the replication run; \cref{sec:reproduction}), and every stage that loads a prepared
series compares it against this hash. A series that differs gives a warning that the results are not comparable with
the published ones. The task thresholds are stored in the repository and are only read during a run
(\cref{sec:preprocessing}).

\paragraph{Extension.} Every component is registered by name: datasets, tasks, pool generators,
assemblers, seam joiners, long-series generators, downstream models, fidelity metric groups, utility
metrics, statistical tests and hypotheses (\cref{fig:extensibility}). A new component is a class or
function in a separate plugin file plus one line of configuration, and no benchmark code changes. A new
pool generator receives the whole set of assemblers and joiners, a new assembler runs on every pool, and a
new method enters the rankings and the research questions automatically, since its family is read from its
description. The results of methods whose code is not released, such as the
proprietary ones of this paper, are kept in the published tables and take part in the analysis, so that
any ranking can be recomputed with or without them.

\paragraph{Release and licences.} For the reproducibility of the study, the code is publicly available
under the Apache License 2.0 at \resurl{https://github.com/zypl-ai/MarathonTS} (release tag \texttt{v1.0.0}),
together with the published tables and analysis outputs of both runs, the configuration of the main run,
the task thresholds, the data hashes and the commands that regenerate the tables and figures. TimesFM~3.0
and Timestral-209M are loaded at pinned revisions; Chronos is loaded from \texttt{amazon/chronos-t5-small},
whose revision (\texttt{a971ba2}) dates from November 2025, before both runs. The unit files of both runs
(series, pools, the weights of the open generators, fidelity, utility metrics and predictions, labels) are on
Hugging Face under the same tag. Not released are the code of zseries, zEDGE and zEDGE\_TS and the weights of
zEDGE and zEDGE\_TS, the artifacts derived from ETTh1 and PJM (\cref{app:datasets}), and the weights of the
downstream models. The
rankings of both runs (\cref{sec:runs}) can be explored in an interactive
leaderboard at \resurl{https://huggingface.co/spaces/zyplai/MarathonTS}, which reads the same published
tables and opens on the main run. The datasets
keep their own licences (\cref{app:datasets}), and third-party components keep theirs: the weights of
TimesFM~3.0 are under a non-commercial licence, and the LKH solver used by the ATSP assembler is free for
academic and non-commercial use.

\fig{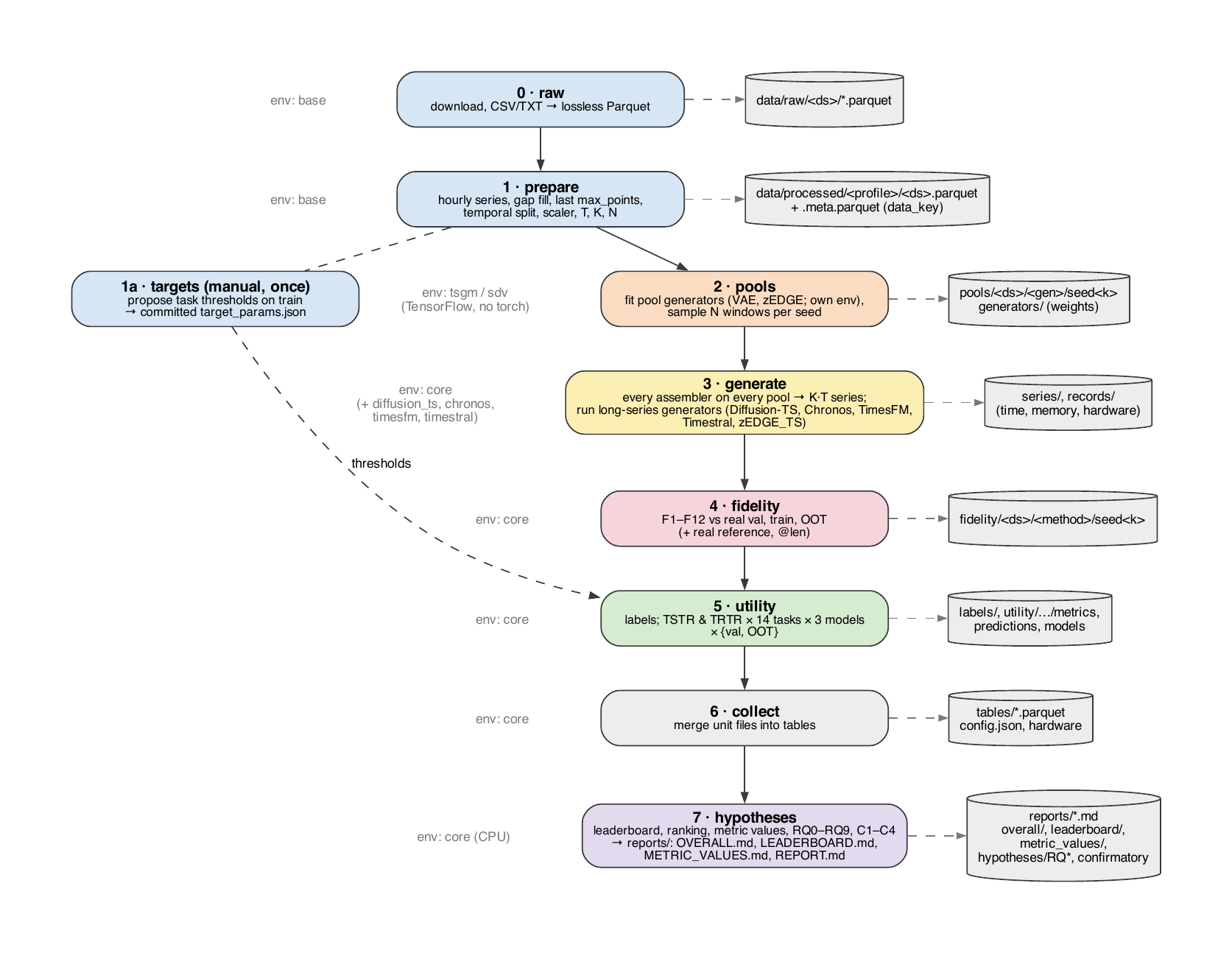}{pipeline}{1}{The benchmark pipeline. Stages run in order. Each
one writes its outputs (right) and can run in its own software environment (left). Task thresholds
are fixed once on the real training split, committed to the repository, and only read at run time
(dashed).}

The results of the main run are presented in the next section.

\section{Results}
\label{sec:results}

\subsection{Runs}
\label{sec:runs}
The benchmark was run twice in profile \texttt{min}, with the same configuration, datasets, task
thresholds and seeds, on different hardware and software stacks (\cref{tab:app-runs}). The settings of every generation unit are
the same in both runs: the fingerprints of all 1\,440 common series match. The code commits differ in naming
and in the handling of results published without code. Eight datasets were identical, whereas the prepared ETT series differed at the value level
(\cref{sec:reproduction}). The \emph{main run} was carried out on one
NVIDIA H200\footnote{The main run was carried out on the computing infrastructure of darya.ai (\url{https://www.greendarya.ai/}).} and covers all 30 methods of \cref{sec:methods}. Every result, ranking and cost reported in this
paper comes from the main run, unless the replication run is named. The \emph{replication run} was carried out on one NVIDIA L4 and covers the
methods of the VAE pool and the four open long-series generators; it serves only to check how well the
results reproduce on another hardware and software stack (\cref{sec:reproduction}). The two runs are never mixed: every ranking
is computed within one run, and times measured on different hardware are not compared.

\subsection{Overall leaderboard}
\label{sec:leaderboard}
\Cref{tab:leaderboard} gives the overall leaderboard of the main run in full: all 24 ranked methods. The six
seam-processing variants are not ranked, because not every assembler has them (\cref{sec:methods-b}).

The table is read by tier. The \emph{place} of a method is its position by mean rank; the \emph{top tier}
(\cref{sec:rankings}) consists of the best method and every method that a paired test over the datasets
does not find significantly worse than it. The tests do not separate the places within the tier, and the
order within it differs between the two runs (\cref{sec:reproduction}). Membership is not monotone in the
place: a method that loses to the best on almost every dataset leaves the tier even with a close mean rank,
whereas a method whose ranks vary between datasets stays in it. TimesFM and Chronos, left out on three and
two datasets for leakage, are compared with the leader on 6 and 7 datasets instead of 9, with less power.

{\small
\begin{longtable}{@{}rllrcccc@{}}
  \caption{Overall leaderboard of the main run (profile \texttt{min}, NVIDIA H200; 9 datasets, 10 seeds): all
  24 ranked methods. Mean rank over the datasets (lower is better), its 95\,\% two-level bootstrap interval (\cref{sec:rules}), the
  probability of being first, membership of the top tier and the number of datasets on which the method is
  ranked. $^{*}$~proprietary (zypl.ai), evaluated as a black box.}
  \label{tab:leaderboard}\\
  \toprule
  Place & Method & Family & Mean rank & 95\,\% interval & P(first) & Top tier & Datasets \\
  \midrule
  \endfirsthead
  \toprule
  Place & Method & Family & Mean rank & 95\,\% interval & P(first) & Top tier & Datasets \\
  \midrule
  \endhead
  \bottomrule
  \endfoot
  1 & \texttt{zedge\_quilting}$^{*}$ & assembly & 8.27 & [7.35, 9.20] & 0.85 & yes & 9 \\
  2 & \texttt{zedge\_zseries}$^{*}$ & assembly & 8.79 & [8.28, 9.47] & 0.08 & yes & 9 \\
  3 & \texttt{zedge\_greedy}$^{*}$ & assembly & 8.95 & [7.97, 10.00] & 0.01 & yes & 9 \\
  4 & \texttt{zedge\_pctsp}$^{*}$ & assembly & 9.18 & [8.13, 10.25] & 0.00 & yes & 9 \\
  5 & \texttt{zedge\_knn}$^{*}$ & assembly & 9.33 & [8.32, 10.48] & 0.00 & yes & 9 \\
  6 & \texttt{zedge\_atsp}$^{*}$ & assembly & 9.39 & [7.83, 11.10] & 0.03 & yes & 9 \\
  7 & \texttt{zedge\_ts\_ar}$^{*}$ & long-series & 9.40 & [7.99, 10.91] & 0.02 & yes & 9 \\
  8 & \texttt{zedge\_ts\_long}$^{*}$ & long-series & 9.50 & [8.25, 10.89] & 0.00 & yes & 9 \\
  9 & \texttt{zedge\_herding}$^{*}$ & assembly & 10.48 & [9.15, 11.88] & 0.00 & yes & 9 \\
  10 & \texttt{zedge\_mbb}$^{*}$ & assembly & 10.58 & [9.10, 12.25] & 0.00 & yes & 9 \\
  11 & \texttt{zedge\_rand}$^{*}$ & assembly (random) & 10.75 & [9.12, 12.49] & 0.00 & yes & 9 \\
  12 & \texttt{vae\_herding} & assembly & 13.53 & [12.79, 14.30] & 0.00 & no & 9 \\
  13 & \texttt{vae\_zseries}$^{*}$ & assembly & 13.68 & [12.15, 15.24] & 0.00 & yes & 9 \\
  14 & \texttt{diffusion\_ts} & long-series & 13.78 & [11.98, 15.39] & 0.00 & no & 9 \\
  15 & \texttt{timesfm} & long-series & 14.43 & [12.56, 16.23] & 0.00 & yes & 6 \\
  16 & \texttt{vae\_quilting} & assembly & 14.44 & [12.64, 16.20] & 0.00 & yes & 9 \\
  17 & \texttt{vae\_pctsp} & assembly & 15.27 & [13.80, 16.67] & 0.00 & no & 9 \\
  18 & \texttt{vae\_atsp} & assembly & 15.28 & [13.26, 17.10] & 0.00 & no & 9 \\
  19 & \texttt{vae\_greedy} & assembly & 15.57 & [13.90, 17.19] & 0.00 & no & 9 \\
  20 & \texttt{chronos} & long-series & 15.78 & [14.04, 17.45] & 0.00 & yes & 7 \\
  21 & \texttt{vae\_knn} & assembly & 16.01 & [14.16, 17.70] & 0.00 & no & 9 \\
  22 & \texttt{vae\_mbb} & assembly & 16.30 & [14.24, 18.20] & 0.00 & no & 9 \\
  23 & \texttt{vae\_rand} & assembly (random) & 16.47 & [14.43, 18.39] & 0.00 & no & 9 \\
  24 & \texttt{timestral} & long-series & 16.77 & [15.53, 18.08] & 0.00 & no & 9 \\
\end{longtable}
}

The top tier holds 15 of the 24 methods. The eleven ranked methods of the zEDGE pool and of zEDGE\_TS
take places 1 to 11 and are all in the tier; \texttt{zedge\_quilting} is first, with a probability of
being first of 0.85. The best open method is \texttt{vae\_herding} (place 12), which is outside the tier,
as are Diffusion-TS and Timestral; \texttt{vae\_zseries}, \texttt{vae\_quilting}, TimesFM and Chronos are
inside it. The two domains agree: the fidelity and utility ranks of the methods have Kendall's
$\tau = 0.88$.

The first place does not depend on the choices made in the leaderboard (\cref{sec:statistics}). It is
the same under every variant fixed in advance, with Kendall's $\tau$ between 0.92 and 0.98 against the
leaderboard, including fidelity against val and OOT only ($\tau = 0.94$), which removes the training
split as a reference. With ordinary average ranks instead of noise-aware ones, the first place is the same,
three places within the top tier change, and \texttt{zedge\_rand}, the random path of the zEDGE pool,
leaves it, so that the tier holds 14 instead of 15 methods (\cref{app:ties}). The
first place is also the same when any one of the nine datasets is left out, and under 2\,000
random weights of the criteria it is first in 99.7\,\% of the draws. 

\subsection{Rankings by slice}
\label{sec:slices}
\textbf{The four rankings (RQ1).} \texttt{zedge\_quilting} is first in all four rankings and in their
versions on the core metrics; the Friedman test separates the methods in each of them ($p < 10^{-7}$).
The task type changes the order more than the reference: rankings with the same task type and different
references agree with $\tau = 0.95$ (classification) and $0.93$ (regression), whereas rankings with the
same reference and different task types agree with $\tau = 0.72$ and $0.75$. The pretrained forecasters
gain most in regression: TimesFM has a mean rank of 11.4 in the regression ranking against val and OOT and
15.7 in the classification one. The leader of a single criterion is not always the overall leader: for
example, \texttt{zedge\_ts\_long} leads the marginal distribution against OOT, the random path of the zEDGE
pool the long structure against OOT, and \texttt{zedge\_ts\_ar} the utility against val.

\textbf{Memorisation and cost.} No synthetic window of any method is an exact copy of a training window,
whereas every window of a control copy of train is one (an exact copy has a DCR below 0.05 in z units; the
training windows are cut with step $T/4$). The median signed
log ratio of the DCR lies between $-0.31$ and $0.28$ for every method (the real OOT series: $-0.13$),
against $-2.7$ to $-1.1$ for control copies of train with noise of 0.05 to 0.25 standard deviations
(\cref{app:memorisation}); TimesFM and Timestral are slightly closer to train than the val windows
(intervals below 0). Copies with noise of 0.5 standard deviations and copies of up to half of the series are
not distinguished from the methods and the real series. The near-copy flag of the protocol is raised for
no method (ratio to the real OOT series at most 1.0, median over datasets; Friedman $p = 0.92$). The share
of near copies of the real OOT series is itself 1 on three datasets and 0 on three others, so the flag is
raised for none of the controls either, and the reading here uses the signed ratio and the exact copies.
These are distance-based diagnostics, not a privacy guarantee; membership and attribute inference attacks
are outside the scope (\cref{sec:limitations}). The cost and the Pareto fronts are described with RQ6 below.

\subsection{Confirmatory family}
\label{sec:results-confirmatory}
\Cref{tab:confirmatory} gives the ten tests of \cref{sec:confirmatory}. C1 covers every ranked method
except Chronos and TimesFM; C2--C4 are computed on the VAE pool (\cref{sec:two-pools}). C1 finds that the methods differ in
both cells. The verdict holds with a permutation calibration and with the incomplete datasets kept. In the
classification cell the difference comes from the methods of the zEDGE pool and zEDGE\_TS: among the 11
methods of the replication run the test is not significant in that cell in either run (\cref{app:c1}). The
other comparisons are inconclusive; the rule requires both a significant test after Holm and a median effect
of more than one noise unit:
\begin{itemize}
  \item C2: Diffusion-TS is better than the assembly median of the VAE pool on 5 (classification) and 6
  (regression) of nine datasets, with median effects of 1.9 and 3.0 noise units, but the tests are not
  significant; Chronos has no clear direction in classification and is better on 5 of 7 datasets in
  regression.
  \item C3: the mean of the assemblers is not separated from the random path in either cell.
  \item C4: seam processing is worse than no processing on every dataset with a verdict (4 in
  classification, 2 in regression). In regression the test is significant after Holm ($p = 0.03$), but
  the median effect, $-0.77$ noise units, is below one noise unit, so the verdict is inconclusive.
\end{itemize}
The intervals of \cref{tab:confirmatory} are not adjusted for multiplicity. The interval of the median effect
includes zero for Diffusion-TS and for window selection in the regression cell and excludes it for Chronos
in the regression cell and for seam processing in both cells. C3 averages over every assembler, zseries included, and
C4 is computed on the path of zseries. In a supplementary analysis that was not pre-specified, the mean of
the open assemblers gives the same verdicts of C3 in both runs (in the main run the median effect changes
from 0.06 to $-0.02$ noise units in the classification cell and from 0.61 to 0.48 in the regression cell),
and no other verdict of the family changes.

\begin{table}[t]
  \centering\small
  \caption{Confirmatory family of the main run: TSTR with the linear model on OOT, task U2.1 (ROC AUC) and
  task U2.6 at $h = 24$ (MASE). Effect: the median over datasets of the effect in noise units (positive: the
  first method is better); intervals: the 95\,\% bootstrap intervals of the mean effect over datasets
  (pre-specified) and of the median effect (supplementary), not adjusted for multiplicity; the number of datasets on which it is better or worse, and the Holm-corrected
  $p$-value over the ten tests.}
  \label{tab:confirmatory}
  \setlength{\tabcolsep}{3.5pt}
  \begin{tabular}{@{}llrccrrrl@{}}
    \toprule
    & & & \multicolumn{2}{c}{95\,\% interval} & & & & \\
    \cmidrule{4-5}
    Test & Comparison & Effect & mean & median & Better & Worse & $p_{\mathrm{Holm}}$ & Verdict \\
    \midrule
    \multicolumn{9}{@{}l}{\emph{Classification, U2.1}} \\
    C1 & all methods (Friedman) & & & & & & 0.012 & differ \\
    C2 & Diffusion-TS vs assembly median & 1.89 & [$-0.63$, 4.85] & [$-0.62$, 4.68] & 5 & 2 & 0.49 & inconclusive \\
    C2 & Chronos vs assembly median & 0.08 & [$-4.96$, 0.95] & [$-2.34$, 1.18] & 2 & 2 & 1.00 & inconclusive \\
    C3 & assemblers vs random path & 0.06 & [$-0.31$, 1.01] & [$-0.60$, 1.41] & 2 & 0 & 1.00 & inconclusive \\
    C4 & seam processing vs none & $-0.92$ & [$-3.47$, $-0.47$] & [$-4.44$, $-0.17$] & 0 & 4 & 0.14 & inconclusive \\
    \midrule
    \multicolumn{9}{@{}l}{\emph{Regression, U2.6, $h = 24$}} \\
    C1 & all methods (Friedman) & & & & & & $1.4 \cdot 10^{-7}$ & differ \\
    C2 & Diffusion-TS vs assembly median & 3.03 & [0.73, 8.59] & [$-0.29$, 13.40] & 6 & 1 & 0.27 & inconclusive \\
    C2 & Chronos vs assembly median & 2.47 & [1.58, 5.53] & [1.28, 5.77] & 5 & 0 & 0.19 & inconclusive \\
    C3 & assemblers vs random path & 0.61 & [0.02, 1.41] & [$-0.27$, 1.55] & 3 & 1 & 0.39 & inconclusive \\
    C4 & seam processing vs none & $-0.77$ & [$-1.32$, $-0.27$] & [$-1.13$, $-0.11$] & 0 & 2 & 0.03 & inconclusive \\
    \bottomrule
  \end{tabular}
\end{table}

\subsection{Research questions}
\label{sec:results-rq}
With nine datasets, the smallest one-sided $p$-value of the Wilcoxon test over datasets is 0.002. After the
Benjamini--Hochberg correction over the hundreds or thousands of comparisons of one question, no global
comparison of RQ2, RQ8 and RQ9 reaches significance, and RQ3 has only verdicts of equivalence. These
questions are therefore read from the direction and size of the effects and from the per-dataset verdicts,
which the published tables give for every comparison. A run with more datasets would be needed for global
verdicts at this level of correction.

\textbf{RQ0: validity.} Of the 297 units of check (c) of \cref{sec:rq} (237 combinations of a fidelity metric and a
reference, the two memorisation metrics and 58 utility units), 231 are discriminative: the methods differ
beyond the noise. Among the fidelity metrics the share is highest against train (72 of 79), then OOT (59 of
79), and lowest against val (44 of 79); one of the two memorisation metrics and 55 of the 58 utility units
are discriminative. On twenty combinations the real OOT series does not score better than most methods, so
they are flagged for inspection; they include the discriminator score (F6.1) against val and train and the
share of near copies (F10.2), a diagnostic rather than a ranking metric. The reading of a flag is given in
\cref{sec:rq}. Thirteen pairs of metrics duplicate
each other ($|\rho| \geq 0.9$), for example the Wasserstein distance and the error of the mean. The order of
the methods by a metric depends on the reference: the median rank correlation of a metric between val and
OOT is 0.63, and between val and train 0.51.

\textbf{RQ2: assembly versus long-series generation.} Descriptively, the long-series generators split into
two groups.
zEDGE\_TS is better than the assembly median of the VAE pool in 98 to 100\,\% of the utility units and in 84
to 85\,\% of the fidelity units, with median effects of 1.6 to 3.3 noise units. Diffusion-TS is better in
71\,\% of the utility units (median 0.4 noise units). TimesFM is better in utility (median 2.4 noise units)
but not in fidelity, Chronos is worse in both, and so is Timestral (median $-3.1$ noise
units in utility). In the exploratory repetition against the assembly median of the zEDGE pool, whose
methods lead the leaderboard, the open long-series generators fall behind assembly: Diffusion-TS, TimesFM,
Chronos and Timestral are better than this median in only 17 to 33\,\% of the utility units and 17 to
36\,\% of the fidelity units, whereas zEDGE\_TS is better in 53 to 76\,\% of the utility units.
Diffusion-TS and TimesFM are therefore competitive with assembly from the open VAE pool, but not with
assembly from the zEDGE pool.

\textbf{RQ3: within assembly.} On the VAE pool, every assembler except the matched-block bootstrap moves the
series towards the real data compared with the random path: the median effect is positive in 59 to 83\,\%
of the fidelity units and 72 to 90\,\% of the utility units. The matched-block bootstrap is equivalent to the
random path on 112 of 158 fidelity units, within the noise of the benchmark (96 to 131 under other
orderings of the seeds; \cref{sec:results-checks}). Seam processing goes the other way: on the path of zseries, the only path
with joiners in profile \texttt{min}, each of the three joiners makes the series worse than the same path
without processing in 83 to 93\,\% of the utility units.
The exploratory repetition on the zEDGE pool gives the same direction: the assemblers are better than the
random path in 59 to 90\,\% of the utility units, and the joiners are worse than no processing in 72 to
88\,\%. On this pool, the mean of the assemblers also passes the rule in the regression cell of C3 (Holm
$p = 0.03$, median effect 1.3 noise units); this test is exploratory and outside the confirmatory family.

\textbf{RQ4: does fidelity predict utility?} 388 of 956 pairs of a fidelity metric and a utility criterion
(the memorisation metrics included) are predictive in the sense of the protocol: the metric's rank correlation with the transfer ratio across the
methods of this run is positive over the datasets after correction. Fidelity against train predicts best (173 of 316 pairs, against 116 for OOT and
99 for val), and classification on OOT is the easiest criterion to predict. The strongest predictors are the
target metrics (F8), which measure the events that the tasks are built on, and the realism of window
features (F6.6), with median correlations above 0.8. In the replication run, with 16 methods from five
generator families, no pair is predictive after correction: with nine datasets the smallest attainable $p$
is 0.004, which 68 pairs reach where the correction over 956 pairs requires 75. The main run restricted to
the same methods has 85 predictive pairs and a median correlation of 0.21 against 0.20, so the runs differ at the
threshold of the correction rather than in the strength of the association. The number of predictive pairs
depends on the generator families compared: in a supplementary leave-one-family-out analysis, the main run
has 85 predictive pairs without the zEDGE family and 235 without the VAE family, and 44 of the 388 pairs
stay predictive whichever family is left out. A metric chosen on the other families orders the methods of a held-out family
against the rest in 75 to 88\,\% of the pairs (median over datasets), with the 95\,\% bootstrap interval
above chance in 19 of 24 combinations of a criterion and a held-out family; in the replication run it does
so for Chronos, TimesFM and Timestral but not for Diffusion-TS.

\textbf{RQ5: gap to real data and stability.} No method reaches the real series. For each of the 30 methods and each of the four
criteria (classification and regression, on val and OOT), the median transfer ratio over datasets is below 1,
and the one-sided Wilcoxon test that it is below 1 is significant after Benjamini--Hochberg correction (120 of
120; $p_{\mathrm{BH}} \leq 0.031$; the upper end of every 95\,\% bootstrap interval is below 1, at most 0.996).
The test is against the value 1, without a margin; for the best methods the gap is small: the largest median
ratio of a method on a single criterion is 0.98.
For ROC AUC the ratio is undefined in 160 of the 2\,700 cells (5.9\,\%), in which the real series itself is
within 0.02 of chance, mostly on Bitcoin and Beijing PM2.5; these cells are left out, and each test uses 6 to
9 datasets. Absolute scores are given in \cref{tab:app-absolute}. Over the four criteria, the median ratio is 0.95 for the best method
(\texttt{zedge\_ts\_ar}), 0.74 for Diffusion-TS, 0.67 for the best method of the VAE pool and 0.39 for
Timestral. The order of the methods survives the shift from val to OOT ($\tau$ from 0.51 for the windows to
0.97 for the seams, 0.80 to 0.83 in utility) and the change of the downstream model ($\tau$ from 0.75 to 0.86
between the linear model and gradient boosting).

\textbf{RQ6: cost.} Training dominates the cost of every trained method. The median time of a series is
1.2\,s for Timestral, 2.4\,s for TimesFM and 33\,s for Chronos. The VAE pool costs about 130\,s per dataset;
the solvers add 31\,s (ATSP) and 46\,s (PCTSP), and the other assemblers and joiners at most a few seconds.
The zEDGE pool costs about 800\,s, Diffusion-TS 2\,350\,s and zEDGE\_TS 2\,930 to 2\,950\,s. The Pareto front of
rank against time therefore runs from the pretrained forecasters, which are fastest but low in the ranking,
to \texttt{zedge\_quilting}, which is best and costs about as much as its pool; every method of the zEDGE
pool costs the same order of time. The slope of the cost in $K$ is not available in profile \texttt{min}.

\textbf{RQ7: moderators.} None of the 176 correlations between the dataset descriptors and the dataset
effects of RQ2 and RQ3 is significant after correction. With nine datasets the analysis has little power,
and the question remains open.

\textbf{RQ8 and RQ9: class balance, outliers and extremes.} In both slices, the order of the families is the
same as in RQ2 and RQ3. zEDGE\_TS is better than the assembly median in 95 to 97\,\% of the class-balance
units and 83\,\% of the extreme-value units; most assemblers of the VAE pool are better than the random path in
most units, and Timestral and Chronos are worse than the assembly median. The fidelity metrics of class
balance predict the utility of the classification tasks (for example, the error of the class shares against
OOT, $\rho = 0.80$), whereas no fidelity metric of the tails predicts the error on extreme targets after
correction.

\subsection{Supplementary checks}
\label{sec:results-checks}
The checks of this subsection were added after the protocol was frozen and are exploratory
(\cref{sec:protocol}).

\textbf{Calendar.} The calendar decoded from the VAE pool advances by exactly one hour at 68
to 78\,\% of the steps (median over datasets and seeds), against 99\,\% for the zEDGE pool and 100\,\% for
the long-series generators; inside a window, its median error is one hour, and 3 to 20\,\% of the steps
are three or more hours off (\cref{tab:app-calendar}). Most assemblers have fewer such steps than the random path (3 to 17\,\% against
18\,\%; the matched-block bootstrap 20\,\%): the jump at the seam, which several assemblers minimise, is
measured over the calendar channels too, and kernel herding, which compares only the value with train,
also tends to pick windows with a more accurate calendar. Part of the effect of window selection in RQ3 may
therefore act through the calendar. Without the fidelity metrics that
use the calendar of the series, the overall order changes only by a few swaps of neighbouring methods
(Kendall's $\tau = 0.96$), one borderline method leaves the top tier, and the fidelity shares of RQ2 and
RQ3 change by at most four percentage points. The utility tasks cannot be compared in this way, because
each uses the hour of the week as a feature.

\textbf{Seam metrics.} The chance level of the share of change points at seams (F5.6) depends on the density of
the seams: 0.6\,\% of the positions lie within three points of a seam for Timestral, 4\,\% for the
assemblers, 8\,\% for Diffusion-TS and zEDGE\_TS and 11\,\% for Chronos. Against its chance level, obtained
by shifting the seams, the share is 2.5 to 3.9 times the chance level for the random paths, the
matched-block bootstrap and herding, and below it for Chronos (0.83), Diffusion-TS (0.45) and zEDGE\_TS (0.38
and 0.69); the raw share therefore penalises the methods with dense seams, although the order of the methods
by the two statistics agrees ($\tau = 0.81$). Ranked on the seam metrics defined for every method, the
overall leaderboard keeps its order ($\tau = 0.94$; 0.92 in the replication run), one borderline method
joins the top tier in each run (\texttt{vae\_atsp} in the main run, \texttt{vae\_greedy} in the replication
run), and no verdict of C1--C4
changes (\cref{app:seams}).

\textbf{Noise margin.} The margins are estimated from ten seeds, that is, from nine
differences of neighbouring seeds. A zero margin, on which the dataset leaves the global test, occurs for
2.7\,\% of the dataset and unit pairs of RQ2 and 3.9\,\% of RQ3; about 0.1\,\% have a margin near zero,
and no verdict of equivalence rests on them. The verdicts of superiority do not depend on the order of the
seeds: with every pair of seeds and with ten random orderings, no verdict of RQ2 or of C1--C4 changes. The
verdicts of equivalence do: of the 435 global verdicts of equivalence in RQ3, 144 hold under every ordering,
and their number ranges from 385 to 521 (for the matched-block bootstrap, from 96 to 131 of 158 fidelity
units, against 112). They also depend on the size of the margin: with half the margin, no verdict of
equivalence remains and seam processing becomes worse than no processing in the regression cell of C4;
with twice the margin, window selection becomes equivalent to the random path in both cells of C3, and seam
processing in the regression cell of C4. Verdicts of equivalence are therefore stated relative to the
estimated noise margin (\cref{sec:rules}).

\textbf{Missing utility cells.} In the main run, 3\,462 of 156\,398 TSTR cells (2.2\,\%) could
not be trained, all of them classification cells whose synthetic labels have one class; no cell failed for a
technical reason. They concentrate in the pretrained forecasters, whose median forecast rarely crosses a
threshold: 31\,\% of the cells of Timestral, 22\,\% of TimesFM and 11\,\% of Chronos, against 2\,\% of
Diffusion-TS and at most six cells of an assembler. With a fixed chance-level penalty instead of the
pre-specified imputation (\cref{sec:rules}), the overall order changes by one swap of the last two methods; with the cells left out,
Timestral rises from 24th to 21st place. In neither case does the top tier, a verdict of RQ2 or RQ3, or a
test of C1--C4 change. In the replication run, with 4.4\,\% of the cells
missing, the top tier and the tests of C1--C4 are unchanged as well.

\subsection{Reproduction on another hardware and software stack}
\label{sec:reproduction}
Thirteen ranked methods are common to both runs: the nine assemblers on the VAE pool and the four open
long-series generators. Their overall leaderboards are compared in full in \cref{tab:app-replication}; for
this comparison the main run's own copies of these methods are ranked among themselves, as in the
replication run.
\begin{itemize}
  \item The orders agree with Kendall's $\tau = 0.64$. The first method is the same.
  \item Every mean rank of the main run lies inside the 95\,\% interval of the replication run, and 12 of 13
  mean ranks of the replication run lie inside the intervals of the main run.
  \item The top tier is the same except for one method at its boundary: \texttt{vae\_greedy}, outside the
  tier in the replication run ($p = 0.05$ after Holm) and inside it in the main run.
  \item The places within the tier do not reproduce: places 2 to 4 hold the same methods in a different
  order; the place shifts most for \texttt{vae\_rand} (7th in the replication run, 12th in the main run) and
  \texttt{vae\_greedy} (12th and 8th), and the mean rank for \texttt{vae\_zseries} (4th and 2nd place) and
  \texttt{vae\_rand}.
\end{itemize}

The differences have four sources, identified by comparing the per-unit tables of the two runs.
\begin{enumerate}
  \item \emph{Trained generators.} The VAE (TensorFlow) and Diffusion-TS were retrained on a different GPU
  and software stack. Early stopping chose different epochs (by up to 148 epochs for the VAE), so the
  weights, the pools and every method built on them differ: the median relative difference of the
  Wasserstein distance (F1.2) between the runs is 9 to 28\,\% for the methods of the VAE pool and 13\,\% for
  Diffusion-TS. This is the main source.
  \item \emph{Sampling.} Chronos is not trained, but with the same seeds its sampled trajectories differ
  between the GPUs (median relative difference of F1.2, 48\,\%).
  \item \emph{Downstream models.} On identical series (the real series, TimesFM and Timestral) the downstream
  scores coincide in the median, with a maximum difference of 0.03 in AUC and 0.04 in MASE.
  \item \emph{Data.} The prepared ETT series of the main run differs at the value level from that of the
  replication run, with the same length, split and dates. The cause is not established: the series that
  the source gives today is identical to that of the replication run and matches the published content
  hash. The effect on the metrics is small (at most $3 \cdot 10^{-4}$ for the real series); the other eight
  datasets are identical in both runs.
\end{enumerate}
The deterministic forecasters confirm this attribution. TimesFM and Timestral produce practically the same series on
both GPUs: the median difference of their fidelity values is 0, and fewer than 0.3\,\% of the values differ by more
than 1\,\% (at most 2.7\,\%). The two runs thus give similar rankings of the 13 shared methods: membership of the
top tier reproduces up to one borderline method, whereas the places within the tier do not. The agreement is
a reproducibility check; the variability across training, software, hardware and data is not estimated from
two runs.

\section{Limitations}
\label{sec:limitations}

This section describes the scope of the benchmark and the conditions that limit its conclusions.

\paragraph{Scope.} The benchmark covers univariate hourly series with a window of one week. Multivariate
series, other sampling frequencies, irregular sampling and exogenous covariates are not covered, and each
dataset contributes a single target series. The nine datasets are taken from energy, environment,
traffic, mobility and finance. Domains with other dynamics are not covered.

\paragraph{Length of the main run.} The main run is carried out in profile \texttt{min}: the last
10\,000 hours of each dataset are used, so the synthetic series consists of $K = 30$ windows, which is
about seven months. This length is sufficient for the seams and for structure of several weeks, but not
for yearly seasonality and slow regimes. The scaling of the methods to the long series of profile
\texttt{max} ($K$ up to 519) is not tested in this run. In profile \texttt{min} the VAE and Diffusion-TS
are trained once per dataset; therefore, the seeds do not capture the variance of training, and a
per-dataset conclusion is conditional on the trained model. The second random path is not run, so the
noise level is taken from the spread between seeds of the random path, and check (a) of RQ0
(\cref{sec:rq}) is not available. This noise level, which is also the equivalence margin (\cref{sec:rules}),
includes the sampling variance of the pool: it makes the verdicts of superiority
conservative and the equivalence region wide, and the verdicts of equivalence vary with the order of the
seeds and the size of the margin (\cref{sec:results-checks}). The independent estimate from the second random path
belongs to profiles \texttt{mid} and \texttt{max}.

\paragraph{Long-range structure.} Group F7 describes structure longer than a window through distributions of
rolling means and standard deviations, weekly-profile summaries and the mean duration of the high-volatility
regime. It does not isolate how variation changes with the aggregation scale, the chronological dependence
between weeks, or the full distribution of regime durations, transition probabilities and recovery from
drawdowns.

\paragraph{Statistical power.} The unit of a global conclusion is the dataset, and there are nine
datasets. After correction, a global verdict requires a large and consistent effect. An effect present on
two or three datasets appears only in the per-dataset verdicts, and an inconclusive verdict leaves the
question open. The top tier is defined by comparisons with the observed leader, without adjustment for its
selection. With nine datasets the smallest one-sided $p$ is $1/512$, against a corrected level near
$10^{-4}$ for all pairs, so a Holm procedure over every pair of methods rejects no pair (\cref{app:tier}). The moderator analysis (RQ7) has nine points and
is exploratory.

\paragraph{Stability of places.} The places within the top tier are not separated by the tests and differ
between the two runs, mainly because the trained generators are retrained on another stack
(\cref{sec:reproduction}); conclusions are therefore stated as tier membership. The replication run does not
cover the zEDGE pool or zEDGE\_TS, and the comparison of the two runs does not isolate the effect of the
hardware. 

\paragraph{Pool generators.} Every assembly method draws on the pool of a generator, so the conclusions
about assembly are conditional on the pool. The main run has two pool generators, the $\beta$-VAE and
zEDGE, with one pool size ($N = 300$) and fixed settings; other window generators may favour other
assemblers. The pre-specified contrasts within assembly and the comparison of long-series generators with
assembly are computed on the VAE pool only (\cref{sec:two-pools}). In their exploratory repetition on the
zEDGE pool, window selection and seam processing keep their direction, whereas the open long-series
generators fall behind assembly; the open long-series generators are therefore competitive only with
assembly from the VAE pool. The settings of the VAE that are not defaults of TSGM ($\beta$, the
latent size, the epoch limits and the prior scale) come from earlier work on other data, and the
sensitivity of the results to them, in particular to the prior scale, is not tested. The decoder of the
VAE ends with a sigmoid, so its windows never leave the range of train; in the main run, they do not reach
its upper bound either (their maximum is 0.80 to 0.98 of the range on the nine datasets). The VAE pool
therefore under-represents the extremes of train, which bears on the tail metrics of its assemblies.

\paragraph{Calendar.} The method families obtain the calendar differently (\cref{sec:methods-c}), and the
calendar decoded from the VAE pool is less accurate than the real one (\cref{sec:results-checks}). The
calendar-continuity metrics are left out of every ranking, and leaving out the other fidelity metrics that
use the calendar changes the order little. The hour of the week is, however, a feature of every utility
task, so the utility of the methods with a less accurate calendar, in particular those of the VAE pool,
may be lowered by their calendar and not only by their values; for the same reason, part of the effect of
window selection may act through the calendar. A comparison in which every method receives a common valid
calendar requires the utility stage to be run again.

\paragraph{Pretrained forecasters.} TimesFM and Timestral are evaluated in their median-output
configurations: their series is a median forecast, which is deterministic and smoother than a sample. The
conclusions concern these configurations, not the full probabilistic output of the models; a series built
from their quantiles or from sampled paths could give other scores. Chronos is run with its small model (\texttt{chronos-t5-small}) in profile
\texttt{min}. The leakage audit relies on the published descriptions of the data on which each model was
trained, pretraining and, where described, post-training, and, where possible, on the data themselves.
Timestral is pretrained on synthetic data only but post-trained on a mix of real and synthetic data whose
real part is not listed~\citep{petralia2026timestral}; its pairs are therefore marked unknown, and its
results may include datasets that the model has seen.

\paragraph{Proprietary methods.} zseries, zEDGE and zEDGE\_TS are evaluated as black boxes: their code is
not released, their design is not described apart from the sampling modes of zEDGE\_TS, and third parties cannot rerun them. zseries uses the
thresholds of task U2.1, which are fixed on train like every threshold of the benchmark; no open assembler
uses them. U2.1 is the classification cell of the confirmatory family, and C3 averages over every assembler,
zseries included; the verdicts of C3 are the same without it (\cref{sec:results-confirmatory}). In profile
\texttt{min} the joiners are applied to the path of zseries only (\cref{sec:methods-b}), so C4 and the seam
contrasts of RQ3 can be checked from the published results, while recomputing them requires the code of
zseries. The joiners themselves are released and can be applied to the path of any open assembler; this
extension is outside the pre-specified protocol. All authors are employed
by the company that develops these methods. The protocol was frozen before the main run (\cref{sec:protocol}), the
methods receive the same inputs and evaluation as every other method, and every ranking can be recomputed
without them (Competing interests).

\paragraph{Utility protocol.} The downstream models are linear models, gradient-boosted trees and a small
network with fixed hyperparameters and light configurations. Sequence models are not included, and in
profile \texttt{min} the network is trained on one task only. All tasks are derived from the series
itself, with the same 11 features. The benchmark measures the quality of the synthetic data (TSTR against
TRTR), not its value as augmentation of real data. Group F8 describes the events of one task (U2.1).

\paragraph{Metrics and aggregation.} Many fidelity metrics are correlated, and the equal weights of the
hierarchy are a choice of the protocol. The robustness checks of \cref{sec:leaderboard} show how much the order depends on this
choice. The seam metrics are undefined for a series without seams, and the
calendar-continuity metrics are left out of every ranking and question. Memorisation is measured by
distances to the training windows; formal privacy guarantees such as differential privacy and membership or
attribute inference attacks are outside the scope. The near-copy flag of the protocol is defined relative to
one real OOT series per dataset, whose own share of near copies is 0 or 1 on six of the nine datasets, so the
flag is raised for none of the controls, an exact copy of train included (\cref{app:memorisation}); the
signed ratio and the count of exact copies are reported instead. The validity check of the metrics against real data rests on one real block per
dataset, the OOT series; controlled perturbations of the real series and repeated length-matched real
blocks would separate the sensitivity of a metric from temporal drift. The fidelity--utility correlations (RQ4) are computed across methods, several
of which share a pool or a generator, so the number of predictive pairs depends on the families compared
(\cref{sec:results-rq}). The leave-one-family-out check covers six families in the main run and five in the
replication run; families outside the benchmark are not tested.

\paragraph{Cost and data.} Times and memory are comparable only on one type of hardware, and all costs
reported here are obtained in one run. Some sources change over time (Bitcoin is updated daily); the data
hashes detect such changes but cannot restore the original data. The ETT series of the main run differs
slightly from the one that the source gives today (\cref{sec:reproduction}), so a new run reproduces the
ETT results of the replication run rather than those of the main run. For ETTh1 and PJM the artifacts derived
from the data are not published for licensing reasons (\cref{app:datasets}) and have to be regenerated.

\paragraph{Planned extensions.} The next version adds an independent noise estimate from the second random
path, a replication with identical input data and fully pinned environments, a common valid calendar for
every method in the utility stage, the seam joiners on the paths of the open assemblers, a memorisation flag
calibrated on controls, and controlled perturbations and repeated length-matched real blocks for the
validity check of the metrics.

\section{Conclusion}
\label{sec:conclusion}

MarathonTS compares two ways of obtaining a long synthetic time series, the assembly of generated
windows and long-series generation, under one protocol, with fidelity, utility, memorisation and cost
measured on the whole series and a statistical analysis frozen before the main run was carried out. On nine
hourly datasets, the main run leads to four conclusions. First, the methods differ, and nine datasets
separate tiers rather than places: 15 of 24 methods are not shown to be worse than the observed leader, which is
not equivalence. In a replication run of the 13 ranked methods of the VAE pool and the open long-series
generators on a different hardware and software stack, in which the zEDGE methods were not rerun, the top
tier of these methods reproduced up to one method and the order within it changed;
results are therefore read by tier. Second, in assembly on the VAE pool, window
selection often moves the series towards the real data compared with a random choice of windows, whereas
the seam joiners, tested on the zseries assembly, often reduce its utility; neither confirmatory contrast is conclusive
under the pre-specified rule. Third, long-series generators do not form one group: Diffusion-TS and TimesFM are
competitive with assembly from the open VAE pool and Chronos and Timestral fall behind it, whereas assembly
from the zEDGE pool is, descriptively, ahead of all four open generators. Fourth, no method trains the downstream models as well as the real
series: the transfer ratio is significantly below 1 for every method, by a few per cent for the best ones, so
the gap to real data remains open.

Nine datasets set the resolution of this run: the global comparisons of
methods and of assembly choices (RQ2, RQ3, RQ8, RQ9) do not reach superiority after the correction for
multiple comparisons, and the moderators of the effects are not identified. A larger set of datasets, the second random path of profiles \texttt{mid}
and \texttt{max} and longer series are the natural next steps; the software is designed so that new
datasets, generators and assemblers enter the comparison through a plugin, and the published tables allow
every ranking to be recomputed with any set of methods.

\section*{Use of generative AI}
The authors used Claude (Anthropic), a generative AI assistant, to help write and review code, edit the
text and prepare figures. The authors checked all such output and take full responsibility for the content
of the paper and the software.

\section*{Competing interests}
All authors are employed by zypl.ai. Three of the evaluated components are proprietary methods of
zypl.ai that are under development: the assembler zseries (\texttt{vae\_zseries}, \texttt{zedge\_zseries}),
the pool generator zEDGE and the long-series generator zEDGE\_TS (\texttt{zedge\_ts\_long},
\texttt{zedge\_ts\_ar}). Their code is not released, and of their design only the two sampling modes of
zEDGE\_TS are described (\cref{sec:methods-c}). Their results are
reported so that they are measured under the same protocol and on the same leaderboard as the open
methods. The following steps are taken to keep this interest from shaping the comparison, and the
limits of these steps are stated.
\begin{itemize}
  \item[(i)] The protocol, the metrics, the rankings and the pre-specified confirmatory family were frozen
  before the main run was carried out (\cref{sec:protocol}); the changes made after the freeze are listed
  and are not part of the pre-specified analysis. They are identical for every method.
  \item[(ii)] No open method uses any component of the proprietary ones: each alternative assembler uses
  the seam criterion of its own literature, and the solvers without one use a generic jump at the seam.
  \item[(iii)] The proprietary methods use the training split only, like every other method, are evaluated
  by the same code, and their training time is included in their cost. One input differs: zseries uses
  the thresholds of task U2.1, fixed on train like every threshold of the benchmark, which no open
  assembler uses; U2.1 is also the classification cell of the confirmatory family
  (\cref{sec:confirmatory,sec:limitations}).
  \item[(iv)] The results of the proprietary methods cannot be reproduced by third parties. All other
  results can be reproduced from the released code, and the released analysis recomputes every ranking
  with any set of methods excluded.
\end{itemize}

\bibliographystyle{unsrtnat}
\bibliography{references}

\begin{thebibliography}{67}
\providecommand{\natexlab}[1]{#1}
\providecommand{\url}[1]{\texttt{#1}}
\expandafter\ifx\csname urlstyle\endcsname\relax
  \providecommand{\doi}[1]{doi: #1}\else
  \providecommand{\doi}{doi: \begingroup \urlstyle{rm}\Url}\fi

\bibitem[Yoon et~al.(2019)Yoon, Jarrett, and van~der Schaar]{yoon2019timegan}
Jinsung Yoon, Daniel Jarrett, and Mihaela van~der Schaar.
\newblock Time-series generative adversarial networks.
\newblock In \emph{Advances in Neural Information Processing Systems},
  volume~32, 2019.

\bibitem[Yuan and Qiao(2024)]{yuan2024diffusionts}
Xinyu Yuan and Yan Qiao.
\newblock Diffusion-{TS}: Interpretable diffusion for general time series
  generation.
\newblock In \emph{International Conference on Learning Representations}, 2024.
\newblock arXiv:2403.01742.

\bibitem[Hunt and Black(1996)]{hunt1996unit}
Andrew~J. Hunt and Alan~W. Black.
\newblock Unit selection in a concatenative speech synthesis system using a
  large speech database.
\newblock In \emph{IEEE International Conference on Acoustics, Speech, and
  Signal Processing (ICASSP)}, 1996.
\newblock \doi{10.1109/ICASSP.1996.541110}.

\bibitem[Sch{\"o}dl et~al.(2000)Sch{\"o}dl, Szeliski, Salesin, and
  Essa]{schodl2000video}
Arno Sch{\"o}dl, Richard Szeliski, David~H. Salesin, and Irfan Essa.
\newblock Video textures.
\newblock In \emph{Proceedings of SIGGRAPH}, 2000.
\newblock \doi{10.1145/344779.345012}.

\bibitem[Efros and Freeman(2001)]{efros2001quilting}
Alexei~A. Efros and William~T. Freeman.
\newblock Image quilting for texture synthesis and transfer.
\newblock In \emph{Proceedings of SIGGRAPH}, 2001.
\newblock \doi{10.1145/383259.383296}.

\bibitem[Ansari et~al.(2024)Ansari, Stella, Turkmen, Zhang, Mercado, Shen,
  Shchur, Rangapuram, Pineda~Arango, Kapoor, Zschiegner, Maddix, Wang, Mahoney,
  Torkkola, Wilson, Bohlke-Schneider, and Wang]{ansari2024chronos}
Abdul~Fatir Ansari, Lorenzo Stella, Caner Turkmen, Xiyuan Zhang, Pedro Mercado,
  Huibin Shen, Oleksandr Shchur, Syama~Sundar Rangapuram, Sebastian
  Pineda~Arango, Shubham Kapoor, Jasper Zschiegner, Danielle~C. Maddix, Hao
  Wang, Michael~W. Mahoney, Kari Torkkola, Andrew~Gordon Wilson, Michael
  Bohlke-Schneider, and Yuyang Wang.
\newblock Chronos: Learning the language of time series.
\newblock \emph{Transactions on Machine Learning Research}, 2024.
\newblock arXiv:2403.07815.

\bibitem[Das et~al.(2024)Das, Kong, Sen, and Zhou]{das2024timesfm}
Abhimanyu Das, Weihao Kong, Rajat Sen, and Yichen Zhou.
\newblock A decoder-only foundation model for time-series forecasting.
\newblock In \emph{International Conference on Machine Learning}, 2024.
\newblock arXiv:2310.10688.

\bibitem[Petralia et~al.(2026)]{petralia2026timestral}
Adrien Petralia et~al.
\newblock {Timestral}, 2026.
\newblock arXiv:2610.37585.

\bibitem[Long et~al.(2026)Long, Huang, Cai, Huang, and Hou]{long2026assembly}
Guoxiong Long, Huizhen Huang, Qikun Cai, Tao Huang, and Chen Hou.
\newblock Evaluating sequence assembly strategies for differentially private
  synthetic time-series forecasting, 2026.
\newblock arXiv:2610.10222.

\bibitem[Ang et~al.(2023)Ang, Huang, Bao, Tung, and Huang]{ang2024tsgbench}
Yihao Ang, Qiang Huang, Yifan Bao, Anthony K.~H. Tung, and Zhiyong Huang.
\newblock {TSGBench}: Time series generation benchmark.
\newblock \emph{Proceedings of the VLDB Endowment}, 17\penalty0 (3), 2023.
\newblock arXiv:2309.03755.

\bibitem[Nikitin et~al.(2024)Nikitin, Iannucci, and Kaski]{nikitin2024tsgm}
Alexander Nikitin, Letizia Iannucci, and Samuel Kaski.
\newblock {TSGM}: A flexible framework for generative modeling of synthetic
  time series.
\newblock In \emph{Advances in Neural Information Processing Systems, Datasets
  and Benchmarks Track}, 2024.
\newblock arXiv:2305.11567.

\bibitem[Higgins et~al.(2017)Higgins, Matthey, Pal, Burgess, Glorot, Botvinick,
  Mohamed, and Lerchner]{higgins2017betavae}
Irina Higgins, Loic Matthey, Arka Pal, Christopher Burgess, Xavier Glorot,
  Matthew Botvinick, Shakir Mohamed, and Alexander Lerchner.
\newblock beta-{VAE}: Learning basic visual concepts with a constrained
  variational framework.
\newblock In \emph{International Conference on Learning Representations}, 2017.

\bibitem[Tashiro et~al.(2021)Tashiro, Song, Song, and Ermon]{tashiro2021csdi}
Yusuke Tashiro, Jiaming Song, Yang Song, and Stefano Ermon.
\newblock {CSDI}: Conditional score-based diffusion models for probabilistic
  time series imputation.
\newblock In \emph{Advances in Neural Information Processing Systems},
  volume~34, 2021.
\newblock arXiv:2107.03502.

\bibitem[Jeha et~al.(2022)Jeha, Bohlke-Schneider, Mercado, Kapoor, Nirwan,
  Flunkert, Gasthaus, and Januschowski]{jeha2022psagan}
Paul Jeha, Michael Bohlke-Schneider, Pedro Mercado, Shubham Kapoor,
  Rajbir~Singh Nirwan, Valentin Flunkert, Jan Gasthaus, and Tim Januschowski.
\newblock {PSA-GAN}: Progressive self attention {GANs} for synthetic time
  series.
\newblock In \emph{International Conference on Learning Representations}, 2022.
\newblock arXiv:2108.00981.

\bibitem[Li et~al.(2025)Li, Qiu, Chen, Wang, Cheng, Shu, Hu, Guo, Zhou, Jensen,
  and Yang]{li2025tsfmbench}
Zhe Li, Xiangfei Qiu, Peng Chen, Yihang Wang, Hanyin Cheng, Yang Shu, Jilin Hu,
  Chenjuan Guo, Aoying Zhou, Christian~S. Jensen, and Bin Yang.
\newblock {TSFM-Bench}: A comprehensive and unified benchmark of foundation
  models for time series forecasting.
\newblock In \emph{Proceedings of the 31st ACM SIGKDD Conference on Knowledge
  Discovery and Data Mining}, 2025.
\newblock \doi{10.1145/3711896.3737442}.
\newblock arXiv:2410.11802.

\bibitem[Gretton et~al.(2012)Gretton, Borgwardt, Rasch, Sch{\"o}lkopf, and
  Smola]{gretton2012mmd}
Arthur Gretton, Karsten~M. Borgwardt, Malte~J. Rasch, Bernhard Sch{\"o}lkopf,
  and Alexander Smola.
\newblock A kernel two-sample test.
\newblock \emph{Journal of Machine Learning Research}, 13:\penalty0 723--773,
  2012.

\bibitem[Lopez-Paz and Oquab(2017)]{lopezpaz2017c2st}
David Lopez-Paz and Maxime Oquab.
\newblock Revisiting classifier two-sample tests.
\newblock In \emph{International Conference on Learning Representations}, 2017.
\newblock arXiv:1610.06545.

\bibitem[Heusel et~al.(2017)Heusel, Ramsauer, Unterthiner, Nessler, and
  Hochreiter]{heusel2017fid}
Martin Heusel, Hubert Ramsauer, Thomas Unterthiner, Bernhard Nessler, and Sepp
  Hochreiter.
\newblock {GANs} trained by a two time-scale update rule converge to a local
  {Nash} equilibrium.
\newblock In \emph{Advances in Neural Information Processing Systems},
  volume~30, 2017.
\newblock arXiv:1706.08500.

\bibitem[Yue et~al.(2022)Yue, Wang, Duan, Yang, Huang, Tong, and
  Xu]{yue2022ts2vec}
Zhihan Yue, Yujing Wang, Juanyong Duan, Tianmeng Yang, Congrui Huang, Yunhai
  Tong, and Bixiong Xu.
\newblock {TS2Vec}: Towards universal representation of time series.
\newblock In \emph{AAAI Conference on Artificial Intelligence}, 2022.
\newblock arXiv:2106.10466.

\bibitem[Naeem et~al.(2020)Naeem, Oh, Uh, Choi, and Yoo]{naeem2020prdc}
Muhammad~Ferjad Naeem, Seong~Joon Oh, Youngjung Uh, Yunjey Choi, and Jaejun
  Yoo.
\newblock Reliable fidelity and diversity metrics for generative models.
\newblock In \emph{International Conference on Machine Learning}, 2020.
\newblock arXiv:2002.09797.

\bibitem[Lubba et~al.(2019)Lubba, Sethi, Knaute, Schultz, Fulcher, and
  Jones]{lubba2019catch22}
Carl~H. Lubba, Sarab~S. Sethi, Philip Knaute, Simon~R. Schultz, Ben~D. Fulcher,
  and Nick~S. Jones.
\newblock catch22: {CAnonical} time-series {CHaracteristics}.
\newblock \emph{Data Mining and Knowledge Discovery}, 33:\penalty0 1821--1852,
  2019.
\newblock \doi{10.1007/s10618-019-00647-x}.

\bibitem[Esteban et~al.(2017)Esteban, Hyland, and R{\"a}tsch]{esteban2017rcgan}
Crist{\'o}bal Esteban, Stephanie~L. Hyland, and Gunnar R{\"a}tsch.
\newblock Real-valued (medical) time series generation with recurrent
  conditional {GANs}, 2017.
\newblock arXiv:1706.02633.

\bibitem[Lan et~al.(2026)Lan, Gu, Xiong, and Ren]{lan2026contsg}
Shaocheng Lan, Shuqi Gu, Zhangzhi Xiong, and Kan Ren.
\newblock {ConTSG-Bench}: A unified benchmark for conditional time series
  generation, 2026.
\newblock arXiv:2603.04767.

\bibitem[Ang et~al.(2025)Ang, Wang, Huang, Bao, Xi, Tung, Jin, and
  Huang]{ang2025ctbench}
Yihao Ang, Qiang Wang, Qiang Huang, Yifan Bao, Xinyu Xi, Anthony K.~H. Tung,
  Chen Jin, and Zhiyong Huang.
\newblock {CTBench}: Cryptocurrency time series generation benchmark, 2025.
\newblock arXiv:2508.02758.

\bibitem[Amorim et~al.(2026)Amorim, Cerqueira, Santos, Azevedo, and
  Soares]{amorim2026privacy}
Luis Amorim, Vitor Cerqueira, Moises Santos, Paulo~J. Azevedo, and Carlos
  Soares.
\newblock Benchmarking time series generation methods for privacy-preserving
  forecasting, 2026.
\newblock arXiv:2608.10891.

\bibitem[Kwon et~al.(2026)Kwon, Kim, Lee, Jung, Kong, Potluru, Park, and
  Lee]{kwon2026seq2synth}
Kiwan Kwon, Kangmin Kim, Hojin Lee, Yeseong Jung, Hyeongwoo Kong, Vamsi~K.
  Potluru, Saerom Park, and Yongjae Lee.
\newblock {Seq2Synth}: Benchmarking temporal fidelity in synthetic sequential
  tabular data.
\newblock In \emph{Proceedings of the ACM International Conference on
  Information and Knowledge Management}, 2026.
\newblock \doi{10.1145/3799682.3841108}.
\newblock arXiv:2607.15606.

\bibitem[Kovar et~al.(2002)Kovar, Gleicher, and Pighin]{kovar2002motion}
Lucas Kovar, Michael Gleicher, and Fr{\'e}d{\'e}ric Pighin.
\newblock Motion graphs.
\newblock In \emph{Proceedings of SIGGRAPH}, 2002.
\newblock \doi{10.1145/566654.566605}.

\bibitem[Kwatra et~al.(2003)Kwatra, Sch{\"o}dl, Essa, Turk, and
  Bobick]{kwatra2003graphcut}
Vivek Kwatra, Arno Sch{\"o}dl, Irfan Essa, Greg Turk, and Aaron Bobick.
\newblock Graphcut textures: Image and video synthesis using graph cuts.
\newblock In \emph{Proceedings of SIGGRAPH}, 2003.
\newblock \doi{10.1145/882262.882264}.

\bibitem[Verhelst and Roelands(1993)]{verhelst1993wsola}
Werner Verhelst and Marc Roelands.
\newblock An overlap-add technique based on waveform similarity ({WSOLA}) for
  high quality time-scale modification of speech.
\newblock In \emph{IEEE International Conference on Acoustics, Speech, and
  Signal Processing (ICASSP)}, 1993.

\bibitem[Allen and Rabiner(1977)]{allen1977stft}
Jont~B. Allen and Lawrence~R. Rabiner.
\newblock A unified approach to short-time {Fourier} analysis and synthesis.
\newblock \emph{Proceedings of the IEEE}, 65\penalty0 (11):\penalty0
  1558--1564, 1977.
\newblock \doi{10.1109/PROC.1977.10770}.

\bibitem[Lugmayr et~al.(2022)Lugmayr, Danelljan, Romero, Yu, Timofte, and
  Van~Gool]{lugmayr2022repaint}
Andreas Lugmayr, Martin Danelljan, Andres Romero, Fisher Yu, Radu Timofte, and
  Luc Van~Gool.
\newblock {RePaint}: Inpainting using denoising diffusion probabilistic models.
\newblock In \emph{IEEE/CVF Conference on Computer Vision and Pattern
  Recognition (CVPR)}, 2022.
\newblock arXiv:2201.09865.

\bibitem[Bar-Tal et~al.(2023)Bar-Tal, Yariv, Lipman, and
  Dekel]{bartal2023multidiffusion}
Omer Bar-Tal, Lior Yariv, Yaron Lipman, and Tali Dekel.
\newblock {MultiDiffusion}: Fusing diffusion paths for controlled image
  generation.
\newblock In \emph{International Conference on Machine Learning}, 2023.
\newblock arXiv:2302.08113.

\bibitem[Zhang et~al.(2023)Zhang, Song, Huang, Chen, and
  Liu]{zhang2023diffcollage}
Qinsheng Zhang, Jiaming Song, Xun Huang, Yongxin Chen, and Ming-Yu Liu.
\newblock {DiffCollage}: Parallel generation of large content with diffusion
  models.
\newblock In \emph{IEEE/CVF Conference on Computer Vision and Pattern
  Recognition (CVPR)}, 2023.
\newblock arXiv:2303.17076.

\bibitem[K{\"u}nsch(1989)]{kunsch1989jackknife}
Hans~R. K{\"u}nsch.
\newblock The jackknife and the bootstrap for general stationary observations.
\newblock \emph{The Annals of Statistics}, 17\penalty0 (3), 1989.
\newblock \doi{10.1214/aos/1176347265}.

\bibitem[Politis and Romano(1992)]{politis1992circular}
Dimitris~N. Politis and Joseph~P. Romano.
\newblock A circular block-resampling procedure for stationary data.
\newblock In Raoul LePage and Lynne Billard, editors, \emph{Exploring the
  Limits of Bootstrap}. Wiley, 1992.

\bibitem[Politis and Romano(1994)]{politis1994stationary}
Dimitris~N. Politis and Joseph~P. Romano.
\newblock The stationary bootstrap.
\newblock \emph{Journal of the American Statistical Association}, 89\penalty0
  (428), 1994.
\newblock \doi{10.1080/01621459.1994.10476870}.

\bibitem[Politis and White(2004)]{politis2004blocklength}
Dimitris~N. Politis and Halbert White.
\newblock Automatic block-length selection for the dependent bootstrap.
\newblock \emph{Econometric Reviews}, 23\penalty0 (1), 2004.
\newblock \doi{10.1081/ETC-120028836}.

\bibitem[Dudek et~al.(2014)Dudek, Le{\'s}kow, Paparoditis, and
  Politis]{dudek2014seasonal}
Anna~E. Dudek, Jacek Le{\'s}kow, Efstathios Paparoditis, and Dimitris~N.
  Politis.
\newblock A generalized block bootstrap for seasonal time series.
\newblock \emph{Journal of Time Series Analysis}, 35\penalty0 (2), 2014.

\bibitem[B{\"u}hlmann(1997)]{buhlmann1997sieve}
Peter B{\"u}hlmann.
\newblock Sieve bootstrap for time series.
\newblock \emph{Bernoulli}, 3\penalty0 (2), 1997.

\bibitem[Rajarshi(1990)]{rajarshi1990markov}
M.~B. Rajarshi.
\newblock Bootstrap in {Markov}-sequences based on estimates of transition
  density.
\newblock \emph{Annals of the Institute of Statistical Mathematics}, 42, 1990.
\newblock \doi{10.1007/BF00050835}.

\bibitem[Theiler et~al.(1992)Theiler, Eubank, Longtin, Galdrikian, and
  Farmer]{theiler1992surrogate}
James Theiler, Stephen Eubank, Andr{\'e} Longtin, Bryan Galdrikian, and
  J.~Doyne Farmer.
\newblock Testing for nonlinearity in time series: The method of surrogate
  data.
\newblock \emph{Physica D}, 58, 1992.
\newblock \doi{10.1016/0167-2789(92)90102-S}.

\bibitem[Schreiber and Schmitz(1996)]{schreiber1996iaaft}
Thomas Schreiber and Andreas Schmitz.
\newblock Improved surrogate data for nonlinearity tests.
\newblock \emph{Physical Review Letters}, 77, 1996.
\newblock \doi{10.1103/PhysRevLett.77.635}.

\bibitem[Carlstein et~al.(1998)Carlstein, Do, Hall, Hesterberg, and
  K{\"u}nsch]{carlstein1998matched}
Edward Carlstein, Kim-Anh Do, Peter Hall, Tim Hesterberg, and Hans~R.
  K{\"u}nsch.
\newblock Matched-block bootstrap for dependent data.
\newblock \emph{Bernoulli}, 4\penalty0 (3):\penalty0 305--328, 1998.

\bibitem[Lall and Sharma(1996)]{lall1996knn}
Upmanu Lall and Ashish Sharma.
\newblock A nearest neighbor bootstrap for resampling hydrologic time series.
\newblock \emph{Water Resources Research}, 32\penalty0 (3):\penalty0 679--693,
  1996.
\newblock \doi{10.1029/95WR02966}.

\bibitem[Vahdat and Kautz(2020)]{vahdat2020nvae}
Arash Vahdat and Jan Kautz.
\newblock {NVAE}: A deep hierarchical variational autoencoder.
\newblock In \emph{Advances in Neural Information Processing Systems},
  volume~33, pages 19667--19679, 2020.

\bibitem[Helsgaun(2000)]{helsgaun2000lkh}
Keld Helsgaun.
\newblock An effective implementation of the {Lin--Kernighan} traveling
  salesman heuristic.
\newblock \emph{European Journal of Operational Research}, 126\penalty0
  (1):\penalty0 106--130, 2000.
\newblock \doi{10.1016/S0377-2217(99)00284-2}.

\bibitem[Balas(1989)]{balas1989pctsp}
Egon Balas.
\newblock The prize collecting traveling salesman problem.
\newblock \emph{Networks}, 19\penalty0 (6):\penalty0 621--636, 1989.
\newblock \doi{10.1002/net.3230190602}.

\bibitem[Chen et~al.(2010)Chen, Welling, and Smola]{chen2010herding}
Yutian Chen, Max Welling, and Alex Smola.
\newblock Super-samples from kernel herding.
\newblock In \emph{Conference on Uncertainty in Artificial Intelligence (UAI)},
  2010.

\bibitem[Lakens(2017)]{lakens2017equivalence}
Dani{\"e}l Lakens.
\newblock Equivalence tests: A practical primer for $t$ tests, correlations,
  and meta-analyses.
\newblock \emph{Social Psychological and Personality Science}, 8\penalty0
  (4):\penalty0 355--362, 2017.
\newblock \doi{10.1177/1948550617697177}.

\bibitem[Wilcoxon(1945)]{wilcoxon1945}
Frank Wilcoxon.
\newblock Individual comparisons by ranking methods.
\newblock \emph{Biometrics Bulletin}, 1\penalty0 (6):\penalty0 80--83, 1945.
\newblock \doi{10.2307/3001968}.

\bibitem[Schuirmann(1987)]{schuirmann1987tost}
Donald~J. Schuirmann.
\newblock A comparison of the two one-sided tests procedure and the power
  approach for assessing the equivalence of average bioavailability.
\newblock \emph{Journal of Pharmacokinetics and Biopharmaceutics}, 15\penalty0
  (6):\penalty0 657--680, 1987.
\newblock \doi{10.1007/BF01068419}.

\bibitem[Holm(1979)]{holm1979}
Sture Holm.
\newblock A simple sequentially rejective multiple test procedure.
\newblock \emph{Scandinavian Journal of Statistics}, 6\penalty0 (2):\penalty0
  65--70, 1979.

\bibitem[Benjamini and Hochberg(1995)]{benjamini1995}
Yoav Benjamini and Yosef Hochberg.
\newblock Controlling the false discovery rate: A practical and powerful
  approach to multiple testing.
\newblock \emph{Journal of the Royal Statistical Society: Series B
  (Methodological)}, 57\penalty0 (1):\penalty0 289--300, 1995.
\newblock \doi{10.1111/j.2517-6161.1995.tb02031.x}.

\bibitem[Friedman(1937)]{friedman1937}
Milton Friedman.
\newblock The use of ranks to avoid the assumption of normality implicit in the
  analysis of variance.
\newblock \emph{Journal of the American Statistical Association}, 32\penalty0
  (200):\penalty0 675--701, 1937.
\newblock \doi{10.1080/01621459.1937.10503522}.

\bibitem[Benavoli et~al.(2016)Benavoli, Corani, and
  Mangili]{benavoli2016posthoc}
Alessio Benavoli, Giorgio Corani, and Francesca Mangili.
\newblock Should we really use post-hoc tests based on mean-ranks?
\newblock \emph{Journal of Machine Learning Research}, 17\penalty0
  (5):\penalty0 1--10, 2016.

\bibitem[Dem{\v{s}}ar(2006)]{demsar2006}
Janez Dem{\v{s}}ar.
\newblock Statistical comparisons of classifiers over multiple data sets.
\newblock \emph{Journal of Machine Learning Research}, 7:\penalty0 1--30, 2006.

\bibitem[Zhou et~al.(2021)Zhou, Zhang, Peng, Zhang, Li, Xiong, and
  Zhang]{zhou2021informer}
Haoyi Zhou, Shanghang Zhang, Jieqi Peng, Shuai Zhang, Jianxin Li, Hui Xiong,
  and Wancai Zhang.
\newblock Informer: Beyond efficient transformer for long sequence time-series
  forecasting.
\newblock In \emph{Proceedings of the AAAI Conference on Artificial
  Intelligence}, volume~35, pages 11106--11115, 2021.

\bibitem[Trindade(2015)]{trindade2015electricity}
Artur Trindade.
\newblock {ElectricityLoadDiagrams20112014}.
\newblock UCI Machine Learning Repository, 2015.
\newblock URL
  \url{https://archive.ics.uci.edu/dataset/321/electricityloaddiagrams20112014}.

\bibitem[H{\'e}brail and B{\'e}rard(2012)]{hebrail2012household}
Georges H{\'e}brail and Alice B{\'e}rard.
\newblock Individual household electric power consumption.
\newblock UCI Machine Learning Repository, 2012.

\bibitem[Mulla(2018)]{pjm2018kaggle}
Rob Mulla.
\newblock Hourly energy consumption.
\newblock Kaggle dataset \texttt{robikscube/hourly-energy-consumption}; data of
  PJM Interconnection LLC, 2018.
\newblock URL
  \url{https://www.kaggle.com/datasets/robikscube/hourly-energy-consumption}.

\bibitem[Liang et~al.(2015)Liang, Zou, Guo, Li, Zhang, Zhang, Huang, and
  Chen]{liang2015pm25}
Xuan Liang, Tao Zou, Bin Guo, Shuo Li, Haozhe Zhang, Shuyi Zhang, Hui Huang,
  and Song~Xi Chen.
\newblock Assessing {Beijing's} {PM2.5} pollution: Severity, weather impact,
  {APEC} and winter heating.
\newblock \emph{Proceedings of the Royal Society A}, 471\penalty0
  (2182):\penalty0 20150257, 2015.
\newblock \doi{10.1098/rspa.2015.0257}.

\bibitem[{Max Planck Institute for Biogeochemistry}(2016)]{jena2016climate}
{Max Planck Institute for Biogeochemistry}.
\newblock Weather station jena, 2009--2016.
\newblock Data download page of the MPI-BGC weather station; file as
  distributed with Keras, 2016.
\newblock URL \url{https://www.bgc-jena.mpg.de/wetter/}.

\bibitem[Hogue(2019)]{hogue2019metro}
John Hogue.
\newblock Metro interstate traffic volume.
\newblock UCI Machine Learning Repository, 2019.
\newblock URL
  \url{https://archive.ics.uci.edu/dataset/492/metro+interstate+traffic+volume}.

\bibitem[Fanaee-T and Gama(2014)]{fanaee2014bike}
Hadi Fanaee-T and Jo{\~a}o Gama.
\newblock Event labeling combining ensemble detectors and background knowledge.
\newblock \emph{Progress in Artificial Intelligence}, 2\penalty0
  (2--3):\penalty0 113--127, 2014.
\newblock \doi{10.1007/s13748-013-0040-3}.

\bibitem[Zielinski(2026)]{zielinski2026bitcoin}
Mark Zielinski.
\newblock Bitcoin historical data.
\newblock Kaggle dataset \texttt{mczielinski/bitcoin-historical-data}; Bitstamp
  exchange data, 2026.
\newblock URL
  \url{https://www.kaggle.com/datasets/mczielinski/bitcoin-historical-data}.
\newblock Kaggle display name ``Zielak''; the full name is that of the GitHub
  account \texttt{mczielinski} linked from the dataset description. Accessed
  October 2026.

\bibitem[{Mistral AI}(2026)]{mistral2026timestral}
{Mistral AI}.
\newblock Introducing {Timestral}.
\newblock \url{https://mistral.ai/news/timestral}, October 2026.
\newblock Blog post, 15 October 2026. Code and weights:
  \url{https://github.com/mistralai/timestral}. Accessed 26 October 2026.

\bibitem[Skillings and Mack(1981)]{skillings1981}
John~H. Skillings and Gregory~A. Mack.
\newblock On the use of a {F}riedman-type statistic in balanced and unbalanced
  block designs.
\newblock \emph{Technometrics}, 23\penalty0 (2):\penalty0 171--177, 1981.

\end{thebibliography}

\appendix
\section{Datasets: sources, licences and leakage audit}
\label{app:datasets}

\paragraph{Sources and licences.} \Cref{tab:app-datasets} lists the source, the licence and the
reference of every dataset. The data themselves are not redistributed: they are downloaded from the
sources at the first stage of a run. Results computed from the data (metrics, tables, rankings) are
published for all nine datasets. Artifacts that contain values derived from the data (synthetic series,
pools, generator weights, labels, predictions) are published under the licence of each dataset, with two
exceptions. ETTh1 is licensed under CC BY-ND~4.0, which does not permit sharing adapted material. Whether
synthetic series and generator weights are adapted material in this sense is not settled, so the
conservative reading is taken. The primary source of PJM restricts the redistribution of data derived
from it. For these two datasets the derived artifacts are withheld and are reproduced locally by running
the pipeline.

\begin{table}[htbp]
  \caption{Dataset sources and licences.}
  \label{tab:app-datasets}
  \centering
  \small
  \begin{tabularx}{\linewidth}{lXll}
    \toprule
    Dataset & Source & Licence & Reference \\
    \midrule
    ETTh1       & GitHub, ETDataset  & CC BY-ND 4.0 & \citet{zhou2021informer} \\
    Electricity & UCI \#321                                         & CC BY 4.0    & \citet{trindade2015electricity} \\
    Household   & UCI \#235                                         & CC BY 4.0    & \citet{hebrail2012household} \\
    PJME        & Kaggle, Hourly Energy Consumption & CC0 1.0      & \citet{pjm2018kaggle} \\
    PM2.5       & UCI \#381                                         & CC BY 4.0    & \citet{liang2015pm25} \\
    Jena        & MPI-BGC weather station (Keras mirror)            & CC BY 4.0    & \citet{jena2016climate} \\
    Metro       & UCI \#492                                         & CC BY 4.0    & \citet{hogue2019metro} \\
    Bike        & UCI \#275                                         & CC BY 4.0    & \citet{fanaee2014bike} \\
    Bitcoin     & Kaggle, Bitcoin Historical Data & CC BY-SA 4.0 & \citet{zielinski2026bitcoin} \\
    \bottomrule
  \end{tabularx}
\end{table}

\paragraph{Preprocessing per dataset.} For each source, one target column is taken and brought to an
hourly step. ETTh1 (oil temperature of a transformer, July 2016 -- June 2018), PJME (load of the PJM East region,
2002--2018), PM2.5 (concentration at the US Embassy in Beijing, 2010--2014) and Metro (westbound traffic
on I-94, Minneapolis -- St Paul) are hourly at the source. Electricity is the total load of 370 Portuguese
clients (2011--2014, 15-minute step), Household the active power of one household in France (2006--2010,
1-minute step) and Jena the air temperature of a weather station (2009--2016, 10-minute step); all three
are averaged per hour. Bitcoin is the logarithm of the last BTC/USD price of each hour (Bitstamp, 1-minute
step), taken from 1 January 2017, before which many minutes have no trades, to 1 October 2026. Duplicate
hours of Metro are averaged, and the series starts after its last gap longer than 30 days (June 2015). A
missing hour of Bike means that no bicycle was rented and is set to zero. PJME has isolated duplicates and
gaps at the changes of daylight saving time. In every dataset, gaps of up to 6 hours are interpolated and
longer gaps are left missing.

\paragraph{Leakage audit.} In October 2026, every pair of a pretrained model and a dataset was checked
against the primary description of the pretraining data: for Chronos, Table~3 of~\citet{ansari2024chronos},
whose Benchmark~II is used for zero-shot evaluation only; for TimesFM~3.0, the model card and the full list of
sub-datasets of its pretraining corpus, with the candidate sub-datasets downloaded and compared with the
benchmark's raw files; for Timestral, the technical report~\citep{petralia2026timestral}, which describes
pretraining on synthetic data only and post-training on a mix of real and synthetic data without a list of
the real data, the model card and the release note~\citep{mistral2026timestral}.
The results of the audit are shown in \cref{tab:app-leakage}. Exact and related pairs are excluded
(\cref{sec:datasets}); weak pairs, which share only the phenomenon, are kept. Earlier versions of TimesFM were also trained on Electricity and Jena; the benchmark uses version 3.0
only.

\begin{table}[htbp]
  \caption{Leakage audit of the pretrained models. \textbf{Exact}: the same data in the pretraining
  corpus; \textbf{related}: the same object at other stations, years or resolution. Both lead to
  exclusion. \textbf{Unknown}: the training data of the model are not listed, so the pair cannot be
  checked; it is kept.}
  \label{tab:app-leakage}
  \centering
  \small
  \begin{tabularx}{\linewidth}{lXXl}
    \toprule
    Dataset & Chronos & TimesFM 3.0 & Timestral \\
    \midrule
    ETTh1       & none (evaluation only)         & none (a test set, not in pretraining) & unknown \\
    Electricity & \textbf{exact}: Electricity at 15 min, hourly and weekly & none (a test set) & unknown \\
    Household   & none                            & \textbf{exact}: the same series under another name, 2007--2010 & unknown \\
    PJME        & none                            & none found & unknown \\
    PM2.5       & \textbf{related}: KDD Cup 2018, Beijing stations, 2017--2018 & \textbf{related}: Beijing Air Quality, 12 stations, 2013--2017 & unknown \\
    Jena        & weak: gridded reanalysis only   & weak: gridded reanalysis only & unknown \\
    Metro       & none                            & none & unknown \\
    Bike        & weak: bike rentals of another city & none & unknown \\
    Bitcoin     & none                            & \textbf{related}: daily Bitcoin price to 2021 & unknown \\
    \bottomrule
  \end{tabularx}
\end{table}

\section{Compared methods and their settings}
\label{app:methods}
\Cref{tab:app-methods} lists the methods of the main run (profile \texttt{min}) and their settings.
Every assembler and seam joiner is run on both pools, under the prefixes \texttt{vae\_} and
\texttt{zedge\_}. For the assemblers that minimise the jump at the seam, the jump is the Euclidean
distance between the last point of a window and the first point of the next over all three channels.

\begin{longtable}{@{}>{\raggedright\arraybackslash}p{0.25\linewidth}>{\raggedright\arraybackslash}p{0.39\linewidth}>{\raggedright\arraybackslash}p{0.28\linewidth}@{}}
  \caption{Methods of the main run and their settings. \texttt{<pool>} is \texttt{vae} or \texttt{zedge}.}
  \label{tab:app-methods}\\
  \toprule
  Method & Idea & Settings \\
  \midrule
  \endfirsthead
  \toprule
  Method & Idea & Settings \\
  \midrule
  \endhead
  \bottomrule
  \endfoot
  \multicolumn{3}{@{}l}{\emph{A. Window selection and order} ($K$ of $N$ windows, without repetition)} \\
  \texttt{<pool>\_zseries} & proprietary (zypl.ai), evaluated as a black box & not released \\
  \texttt{<pool>\_rand} & $K$ random windows in random order; control & --- \\
  \texttt{<pool>\_greedy} & nearest neighbour by the jump at the seam, from a random window & --- \\
  \texttt{<pool>\_atsp} & asymmetric TSP over the pool, Lin--Kernighan--Helsgaun~\citep{helsgaun2000lkh}; cheapest run of $K$ consecutive windows of the tour & LKH via \texttt{elkai} \\
  \texttt{<pool>\_pctsp} & prize-collecting TSP~\citep{balas1989pctsp}: which $K$ windows and their order at once & OR-Tools; at most 2\,000 solutions or 60\,s per unit \\
  \texttt{<pool>\_mbb} & matched-block bootstrap~\citep{carlstein1998matched}: $P(i \to j) \propto \exp(-\tfrac12 (d_{ij}/h)^2)$ & $h$ = median jump \\
  \texttt{<pool>\_knn} & $k$-NN resampling~\citep{lall1996knn}: next window among the $k$ nearest by first point, weight $\propto 1/\text{rank}$ & $k = \operatorname{round}(\sqrt{N})$ \\
  \texttt{<pool>\_quilting} & image quilting~\citep{efros2001quilting}: random among near-best overlaps, cut at minimum error & overlap 12; tolerance $1.1\times$ best \\
  \texttt{<pool>\_herding} & kernel herding~\citep{chen2010herding}: $K$ windows minimising the MMD to the training windows, random order & every 4th point of a window as features \\
  \midrule
  \multicolumn{3}{@{}l}{\emph{B. Seam processing} (on the path of \texttt{<pool>\_zseries})} \\
  \texttt{<pool>\_zseries\_xfade} & crossfade of the overlap with a Hann window~\citep{allen1977stft} & overlap 12 \\
  \texttt{<pool>\_zseries\_wsola} & WSOLA~\citep{verhelst1993wsola}: shift up to 12 points to the best correlation, then crossfade & overlap 12 \\
  \texttt{<pool>\_zseries\_quilt} & switch to the next window at the point of smallest difference & overlap 12 \\
  \midrule
  \multicolumn{3}{@{}l}{\emph{C. Long-series generators}} \\
  \texttt{diffusion\_ts} & Diffusion-TS~\citep{yuan2024diffusionts}, autoregressive infilling with context $T/2$; calendar given & official recipe, 18\,000 steps, early stopping on holdout; 50 DDIM steps; one model per dataset \\
  \texttt{chronos} & Chronos~\citep{ansari2024chronos}, one sampled trajectory in chunks of 64 & \texttt{chronos-t5-small}, context 512, default sampling \\
  \texttt{timesfm} & TimesFM~3.0~\citep{das2024timesfm}, median forecast in one pass & \texttt{google/\allowbreak timesfm-3.0-pytorch}, context up to 15\,360 \\
  \texttt{timestral} & Timestral-209M~\citep{petralia2026timestral}, median forecast in blocks of 1\,024 & \texttt{mistralai/\allowbreak Timestral-209M}, context up to 32\,768 \\
  \texttt{zedge\_ts\_long} & zEDGE\_TS (zypl.ai, proprietary): inpainting in blocks of 12 latent vectors, conditioned on the last 3 of the previous block & not released; one model per dataset, shared with \texttt{zedge\_ts\_ar} \\
  \texttt{zedge\_ts\_ar} & zEDGE\_TS (zypl.ai, proprietary): autoregressive, one latent vector at a time, conditioned on the preceding 11 & not released; the same model as \texttt{zedge\_ts\_long} \\
\end{longtable}

The pool generators are the $\beta$-VAE of \cref{sec:pool} (convolutional encoder and decoder of TSGM,
latent dimension 16, $\beta = 2$, Adam with learning rate $10^{-3}$, batch 64, up to 400 epochs, early
stopping on the holdout loss with patience 50 after a minimum of 150 epochs, sampling from
$\mathcal{N}(0, 0.5^2 I)$, one model per dataset) and zEDGE, which is proprietary and not described.
Both generators give $N = 300$ windows per dataset and seed. Diffusion-TS is trained with the official
recipe with the window length $T$: batch 128 with gradient accumulation 2, Adam with a warm-up from $10^{-5}$ to
$8 \cdot 10^{-4}$ over 500 steps followed by reduction on plateau, and an exponential moving average of the
weights. Two deviations from the official recipe are made: early stopping on the holdout windows is used, as for every
trained generator of the benchmark, and infilling is carried out with 50 instead of 200 sampling steps, to
bound the generation time. The pretrained forecasters are used zero-shot with the default settings of their
libraries and pinned model revisions.

\section{Full list of fidelity metrics}
\label{app:metrics}
\Cref{tab:app-metrics} lists the 66 fidelity metric IDs; 11 of them have several outputs (lags,
quantiles, window lengths), 83 outputs in all. Unless stated otherwise, every metric is a
deviation from the reference; that is, 0 is ideal and lower is better. $\Delta_k$ is the change over $k$ hours;
$z$ means that both series are standardised with the mean and standard deviation of the reference.
Windows are taken with length $T$ and step $T/4$, at most 3\,000 per series. The 16 window features are
the mean,
standard deviation, skewness, kurtosis, minimum, maximum, 10th and 90th percentiles, mean $|\Delta_1|$,
standard deviation of $\Delta_1$, lag-1 and lag-24 autocorrelation, rate of sign changes of the standardised
window, relative positions of the maximum and the minimum, and the last minus the first value.

\begin{longtable}{@{}p{0.07\linewidth}p{0.04\linewidth}>{\raggedright\arraybackslash}p{0.81\linewidth}@{}}
  \caption{Fidelity metrics. $n$: number of metrics of the ID.}
  \label{tab:app-metrics}\\
  \toprule
  ID & $n$ & How it is computed \\
  \midrule
  \endfirsthead
  \toprule
  ID & $n$ & How it is computed \\
  \midrule
  \endhead
  \bottomrule
  \endfoot
  \multicolumn{3}{@{}l}{\emph{F1. Marginal distribution of values}} \\
  F1.1 & 1 & Kolmogorov--Smirnov statistic \\
  F1.2 & 1 & Wasserstein-1 distance divided by the standard deviation of the reference \\
  F1.3 & 1 & Anderson--Darling $k$-sample statistic \\
  F1.4 & 1 & correlation of the quantiles 0.01--0.99 (higher is better) \\
  F1.5 & 3 & quantile errors in reference standard deviations: mean over $p \in \{0.01, 0.05, 0.25, 0.5, 0.75, 0.95, 0.99\}$, and at 0.01 and 0.99 \\
  F1.6 & 4 & errors of the mean, standard deviation ($|\log$ ratio$|$), skewness and excess kurtosis \\
  F1.7 & 1 & share of points outside the range of the reference \\
  F1.8 & 1 & share of the 50 quantile bins of the reference hit by the synthetic series (higher is better) \\
  F1.9 & 1 & Jensen--Shannon distance of the histograms on the same bins \\
  F1.10 & 1 & error of the Hill tail index on the upper 5\% \\
  \multicolumn{3}{@{}l}{\emph{F2. Increments}} \\
  F2.1 & 5 & KS of $\Delta_k$, $k \in \{1, 6, 24, 168\}$; Wasserstein-1 of $\Delta_{24}$ \\
  F2.2 & 2 & $|\log$ ratio$|$ of the standard deviations of $\Delta_1$ and $\Delta_{24}$ \\
  F2.3 & 1 & error of the share of positive $\Delta_1$ \\
  F2.4 & 1 & error of the share of $|\Delta_1|$ above the reference 99th percentile \\
  \multicolumn{3}{@{}l}{\emph{F3. Temporal dependence}} \\
  F3.1 & 1 & mean absolute error of the autocorrelation, lags $1 \dots T-1$ \\
  F3.2 & 1 & the same, lags $T \dots 3T$, beyond one window \\
  F3.3 & 1 & mean absolute error of the partial autocorrelation, lags 1--48 \\
  F3.4 & 2 & mean absolute error of the autocorrelation of $|\Delta_1|$ and of $\Delta_1^2$, lags 1--48 (volatility clustering) \\
  F3.5 & 1 & mean $|\log_{10}$ ratio$|$ of the normalised Welch power spectra \\
  F3.6 & 2 & error of the share of spectral power at the daily and the weekly frequency \\
  F3.7 & 1 & error of the exponent of detrended fluctuation analysis, scales 16--256 \\
  F3.8 & 1 & error of the correlation of the value with the sine and cosine of the hour of the week \\
  \multicolumn{3}{@{}l}{\emph{F4. Seasonality and calendar}} \\
  F4.1 & 1 & mean absolute error of the mean profile over the 168 hours of the week ($z$) \\
  F4.2 & 1 & the same for the standard deviation profile \\
  F4.3 & 1 & the same for the mean profile over the 24 hours of the day \\
  F4.4 & 1 & share of steps at which the hour of the week advances by exactly one (higher is better) \\
  F4.5 & 1 & share of seams at which it does not \\
  \multicolumn{3}{@{}l}{\emph{F5. Seams} (undefined for a series without seams, except F5.5)} \\
  F5.1 & 1 & KS between $\Delta_1$ at the seams and $\Delta_1$ of the reference \\
  F5.2 & 1 & KS between $\Delta_1$ at the seams and $\Delta_1$ elsewhere in the same series \\
  F5.3 & 2 & $|\Delta_1|$ at a seam in standard deviations of the reference $\Delta_1$: mean and 95th percentile \\
  F5.4 & 1 & difference of the lag-1 autocorrelation within $\pm 24$ points of the seams and of random positions \\
  F5.5 & 1 & error of the rate of change points (PELT, on 2-hour means) \\
  F5.6 & 1 & share of the detected change points within $\pm 3$ points of a seam \\
  F5.7 & 1 & $|\mathrm{AUC} - 0.5|$ of a logistic regression telling a snippet around a seam from one elsewhere \\
  \multicolumn{3}{@{}l}{\emph{F6. Distribution of windows}} \\
  F6.1 & 1 & discriminative score: $|\text{accuracy} - 0.5|$ of a GRU classifier, real against synthetic windows, split in time~\citep{yoon2019timegan} \\
  F6.2 & 1 & $|\mathrm{AUC} - 0.5|$ of a gradient-boosting classifier on window features (C2ST, \citealp{lopezpaz2017c2st}) \\
  F6.3 & 1 & predictive score: error of a one-step ridge forecast trained on the synthetic series against one trained on real data \\
  F6.4 & 2 & unbiased MMD$^2$ with an RBF kernel~\citep{gretton2012mmd} between windows of length $T$ and $2T$ \\
  F6.6 & 1 & Fréchet distance between the window features~\citep{heusel2017fid} \\
  F6.8 & 1 & $|\log$ ratio$|$ of the median nearest-neighbour distance synthetic $\to$ real to real $\to$ real \\
  \multicolumn{3}{@{}l}{\emph{F7. Long-range structure}} \\
  F7.1 & 1 & KS of the rolling mean ($z$), windows of 168 and 720 hours \\
  F7.2 & 1 & KS of the rolling standard deviation ($z$), same windows \\
  F7.3 & 1 & $|\log$ ratio$|$ of the rates of crossing the reference 0.1, 0.5 and 0.9 quantiles \\
  F7.4 & 1 & KS of the lengths of excursions above the 0.9 and below the 0.1 quantile \\
  F7.5 & 1 & KS of the drawdown from the rolling weekly maximum ($z$) \\
  F7.6 & 2 & error of the linear trend per week; $|\log$ ratio$|$ of the standard deviation of the weekly means \\
  F7.7 & 1 & $|\log$ ratio$|$ of the mean duration of the high-volatility regime of a two-state Gaussian HMM on $\Delta_1$ \\
  \multicolumn{3}{@{}l}{\emph{F8. Tail events of task U2.1} (24-hour increase above the 0.90 quantile of train)} \\
  F8.1 & 1 & error of the share of events \\
  F8.2 & 1 & KS of the durations of events \\
  F8.3 & 1 & KS of the times between the starts of events \\
  F8.4 & 1 & mean absolute error of the event rate by hour of the week \\
  F8.5 & 1 & error of the event rate given the quintile of the past changes $\Delta_1$, $\Delta_{24}$, $\Delta_{168}$ \\
  F8.6 & 1 & KS of the next 24-hour change given the quintile of the current value \\
  \multicolumn{3}{@{}l}{\emph{F9. Diversity and coverage} (on window features, $k = 5$; \citealp{naeem2020prdc})} \\
  F9.1 & 2 & precision and recall (higher is better) \\
  F9.2 & 2 & $|\log \text{density}|$; coverage (higher is better) \\
  \multicolumn{3}{@{}l}{\emph{F10. Memorisation} (against train only; ranked separately)} \\
  F10.1 & 1 & $|\log$ ratio$|$ of the median distance to the closest training window, synthetic against val windows \\
  F10.2 & 1 & share of synthetic windows closer to train than the 5th percentile of the val windows' distances \\
  \multicolumn{3}{@{}l}{\emph{F11. Class balance} (over every classification task)} \\
  F11.1 & 1 & mean $|\log$ ratio$|$ of the positive shares over binary tasks \\
  F11.2 & 1 & the same over rare-event tasks ($q \ge 0.95$) \\
  F11.3 & 1 & total variation distance of the classes of the three-class task \\
  F11.4 & 1 & share of binary tasks in which the synthetic series has no positive example \\
  \multicolumn{3}{@{}l}{\emph{F12. Outliers and extremes}} \\
  F12.1 & 1 & $|\log$ ratio$|$ of the shares of points beyond the reference 0.5\% and 99.5\% quantiles \\
  F12.2 & 1 & error of the mean exceedance beyond these quantiles \\
  F12.3 & 1 & $|\log$ ratio$|$ of the rates of local spikes (distance from the rolling median of 25 points) \\
  F12.4 & 1 & KS of the weekly maxima \\
  F12.5 & 1 & KS of the weekly minima \\
\end{longtable}

\section{Calendar of the series}
\label{app:calendar}
The calendar of a series is its hour of the week $w_t \in \{0, \dots, 167\}$. For an assembly method it is
decoded from the two calendar channels of each window,
$w_t = \operatorname{round}\bigl(168 \cdot \operatorname{atan2}(2s_t - 1,\, 2c_t - 1) / 2\pi\bigr) \bmod 168$,
where $s_t$ and $c_t$ are the sine and cosine channels; the other methods return it directly. Two measures
are computed for every series. The first is the share of steps at which the hour of the week advances by
exactly one, as in metric F4.4. The second is the error inside a \emph{piece}, the stretch between two
seams: for every piece, the valid calendar $(c + i) \bmod 168$, $i = 0, 1, \dots$, with the offset $c$
that is closest to the series is found, and the error of a step is the circular distance, in hours,
between its hour and that calendar. A jump at a seam is therefore not an error, since a new piece may
start at any hour, whereas an hour that jitters inside a window is. \Cref{tab:app-calendar} gives the
medians over datasets and seeds for both runs. Its code is \texttt{marathonts/hypotheses/calendar\_check.py}.

\begin{table}[htbp]
  \caption{Calendar of the series, medians over datasets and seeds. \emph{Step $+1$}: share of steps at
  which the hour of the week advances by exactly one (F4.4). \emph{Error}: distance of the hour of the
  week from the closest valid calendar inside a piece, in hours (median and 90th percentile).
  \emph{${\ge}\,3$\,h}: share of steps with an error of three hours or more. The replication run has
  neither the zEDGE pool nor zEDGE\_TS. Supplementary.}
  \label{tab:app-calendar}
  \centering
  \small
  \begin{tabular}{lccccccc}
    \toprule
    & \multicolumn{4}{c}{Main run} & & \multicolumn{2}{c}{Replication run} \\
    \cmidrule{2-5}\cmidrule{7-8}
    Method & Step $+1$ & Error & 90\,\% & ${\ge}\,3$\,h & & Step $+1$ & ${\ge}\,3$\,h \\
    \midrule
    \texttt{vae\_rand}      & 0.68 & 1 & 4   & 18\,\% & & 0.68 & 20\,\% \\
    \texttt{vae\_greedy}    & 0.72 & 1 & 3   & 15\,\% & & 0.68 & 16\,\% \\
    \texttt{vae\_atsp}      & 0.75 & 1 & 2   & 10\,\% & & 0.75 & 12\,\% \\
    \texttt{vae\_pctsp}     & 0.74 & 1 & 2.5 & 10\,\% & & 0.75 & 9\,\% \\
    \texttt{vae\_mbb}       & 0.68 & 1 & 5   & 20\,\% & & 0.69 & 19\,\% \\
    \texttt{vae\_knn}       & 0.70 & 1 & 4   & 17\,\% & & 0.70 & 17\,\% \\
    \texttt{vae\_quilting}  & 0.74 & 1 & 3   & 13\,\% & & 0.70 & 16\,\% \\
    \texttt{vae\_herding}   & 0.76 & 1 & 2   & 7\,\%  & & 0.69 & 16\,\% \\
    \texttt{vae\_zseries}   & 0.78 & 1 & 2   & 3\,\%  & & 0.78 & 7\,\% \\
    \texttt{vae\_zseries} + seam joiners & 0.78--0.79 & 1 & 2 & 3--7\,\% & & 0.77--0.78 & 6--10\,\% \\
    \midrule
    zEDGE pool, every assembler & 0.99 & 0 & 0 & 0\,\% & & -- & -- \\
    \texttt{zedge\_zseries} + crossfade, WSOLA & 0.99 & 0 & 0 & 4\,\% & & -- & -- \\
    \midrule
    long-series generators & 1.00 & 0 & 0 & 0\,\% & & 1.00 & 0\,\% \\
    \bottomrule
  \end{tabular}
\end{table}

\section{Utility tasks and models}
\label{app:utility}
\paragraph{Tasks.} \Cref{tab:app-tasks} lists the 14 tasks. The quantile thresholds are computed on the
changes of real train and are fixed in the repository before the run.

\begin{table}[htbp]
  \caption{Downstream tasks. $\Delta_h$ is the change over the next $h$ hours.}
  \label{tab:app-tasks}
  \centering
  \small
  \begin{tabularx}{\linewidth}{llXl}
    \toprule
    Group & Type & Target & Primary metric \\
    \midrule
    U2.1 & binary     & $\Delta_{24}$ above the 0.90 quantile & ROC AUC \\
    U2.2 & binary     & $\Delta_h$ above the 0.90 quantile, $h \in \{1, 6, 168\}$ & ROC AUC \\
    U2.3 & binary     & $\Delta_{24}$ above the 0.95 and the 0.99 quantile (rare events) & ROC AUC \\
    U2.4 & binary     & $\Delta_{24}$ below the 0.10 quantile & ROC AUC \\
    U2.5 & 3 classes  & $\Delta_{24}$ below the 0.10, between, or above the 0.90 quantile & log-loss \\
    U2.6 & regression & value $x_{t+h}$, $h \in \{1, 24, 168\}$ & MASE \\
    U2.7 & quantiles  & quantiles 0.1, 0.5 and 0.9 of $x_{t+24}$ & pinball loss \\
    U2.8 & multi-step & the vector $x_{t+1}, \dots, x_{t+24}$ & MASE \\
    U2.9 & regression & standard deviation of the changes over the next 24 hours & MASE \\
    \bottomrule
  \end{tabularx}
\end{table}

\paragraph{Models.} In the linear model and the network, the 11 features are standardised on their own
training data, and for regression also the target; the trees use the raw features. The hyperparameters of
profile \texttt{min} are fixed in advance and are as follows:
\begin{itemize}
  \item \textbf{linear}: logistic or softmax regression with balanced class weights and L2 penalty $C = 1$
  (L-BFGS, up to 300 iterations); ridge regression with $\alpha = 1$ for regression, volatility and the
  multi-step task (exact solution); linear quantile regression with a smoothed pinball loss for the three
  quantiles at once.
  \item \textbf{gbdt}: XGBoost with 100 trees, depth 4, learning rate 0.2 and 64 histogram bins; the
  positive class reweighted by $(1-p)/p$ for binary tasks and balanced weights for the three-class task;
  one model with the quantile loss for the three quantiles; a single ensemble with vector leaves for
  the multi-step task.
  \item \textbf{mlp}: a fully connected network $11 \to 64 \to 32 \to k$ with ReLU, Adam with learning rate
  $10^{-3}$, weight decay $10^{-4}$, batch 512, 30 epochs without early stopping; trained on task U2.1
  only in profiles \texttt{min} and \texttt{mid}.
\end{itemize}
The probabilities of every classifier are mapped back from the balanced shares to the class shares of
its own training data, $P(c \mid x) \propto P_{\mathrm{bal}}(c \mid x) \cdot \mathrm{share}(c)$, so that the
Brier score, the log-loss and the calibration error measure calibration and not the reweighting.

\paragraph{Metrics.} The following metrics are computed. Binary tasks: ROC AUC, PR AUC, Brier score, log-loss, expected calibration error
(10 equal-width bins), F1 at the threshold that predicts the positive share of real train, and the lift
in the top decile. Three-class task: macro one-vs-rest AUC, log-loss and macro F1. Regression and
volatility: MAE, RMSE, MASE, sMAPE and the MAE on the test rows with the 5\% lowest and highest targets.
Quantiles: pinball loss, coverage of the 10--90\% interval and the share of rows with crossing quantiles.
Multi-step: MAE and MASE over the horizon and at its last step. MASE, the pinball loss and the error on
extreme targets are divided by the MAE of the seasonal naive forecast (lag 168) on real train; MAE and
RMSE stay in the units of the series.

\section{Analysis protocol}
\label{app:protocol}
\Cref{tab:app-protocol} lists what the analysis protocol fixes and where it is stated in
\texttt{docs/HYPOTHESES.md}; the date of the freeze and the changes made after it are given in
\cref{sec:protocol} and \texttt{docs/PROTOCOL\_AMENDMENTS.md}. Two ablations of the zseries assembler listed in the
protocol are not part of the release, and the published analysis is run without them.

\begin{table}[htbp]
  \caption{Components fixed by the analysis protocol (frozen on 21 October 2026, before the main run).}
  \label{tab:app-protocol}
  \centering
  \small
  \begin{tabularx}{\linewidth}{@{}>{\raggedright\arraybackslash}p{0.2\linewidth}X>{\raggedright\arraybackslash}p{0.17\linewidth}@{}}
    \toprule
    Component & What is fixed & Protocol \\
    \midrule
    Core metrics & one or two canonical fidelity metrics per criterion for the core rankings; the primary
    metric of every utility task & RQ1, core rankings \\
    Confirmatory family & cells U2.1 and U2.6 at $h = 24$, linear model, OOT, primary metric; tests C1--C4;
    the long-series generators of C2 (Diffusion-TS, Chronos) & 3.9 \\
    Contrasts & RQ2: each long-series generator against the random path and the assembly median of one
    pool; RQ3: each assembler against the random path, each joiner against the same path without seam
    processing, and the main effects & RQ2, RQ3 \\
    Multiplicity & C1--C4: Holm over the ten tests; secondary questions (RQ0, RQ2--RQ6, RQ8, RQ9):
    Benjamini--Hochberg within a question; RQ7: exploratory, no conclusions & 2, multiple comparisons \\
    Noise and verdicts & noise: the median difference between seeds of the random path, or the larger
    between-seed spread for independently trained methods; ``better'' or ``worse'': significant after
    correction and a median effect of more than one noise unit; ``same'': TOST with the margin of one
    noise unit & 2, noise; 2, verdict \\
    Exclusion rules & pairs of a pretrained model and a dataset in its training data; seam-processing
    variants not ranked; calendar-continuity metrics F4.4--F4.5 left out; a missing utility cell enters at
    the chance level or the worst value of its block & 2; RQ1, ranked methods \\
    Ties and tiers & ties within the noise on the overall leaderboard; the top tier: the best method and
    every method not worse than it by a one-sided Wilcoxon test with Holm correction & RQ1 \\
    \bottomrule
  \end{tabularx}
\end{table}

\section{Chance levels of the utility metrics}
\label{app:chance}
\Cref{tab:app-chance} lists the value that a utility cell which a synthetic series could not train receives
under the pre-specified imputation (\cref{sec:rules}): the worse of the chance level of the metric and the worst
value of any method on the same dataset, seed, task, model and split.

\begin{table}[htbp]
  \caption{Chance levels of the utility metrics. $p$: the positive share of the test rows; $K$: the number of
  classes. A metric without a chance level takes the worst value of its block.}
  \label{tab:app-chance}
  \centering
  \small
  \begin{tabularx}{\linewidth}{@{}lX@{}}
    \toprule
    Metric & Chance level \\
    \midrule
    ROC AUC (binary; macro for three classes) & 0.5 \\
    PR AUC & $p$ \\
    Log-loss & the entropy $-p \ln p - (1-p)\ln(1-p)$ for a binary task; $\ln K$ for $K$ classes \\
    Brier score & $p(1-p)$ \\
    F1, ECE, lift at 10\,\%, macro F1 & none: the worst value of the block \\
    Losses of regression, quantile, volatility and multi-step tasks & none: the worst value of the block (not
    needed in the published runs, in which no such cell failed) \\
    \bottomrule
  \end{tabularx}
\end{table}

\section{Eligibility and calibration of C1}
\label{app:c1}
C1 is a Friedman test with the datasets as blocks (\cref{sec:confirmatory}). Its design is complete: Chronos
(left out on Beijing PM2.5 and Electricity) and TimesFM (Beijing PM2.5, Bitcoin and Household) are the only
ranked methods with a missing cell in either confirmatory cell, and both drop out of the test. C1 therefore
compares 22 methods in the main run (the nine assemblers of the VAE pool, the nine of the zEDGE pool, the two
variants of zEDGE\_TS, Diffusion-TS and Timestral) and 11 in the replication run (the nine assemblers of the
VAE pool, Diffusion-TS and Timestral), on all nine datasets. \Cref{tab:app-c1} gives the pre-specified test
with a permutation $p$ (the values permuted within each dataset, 100\,000 permutations) and three variants:
every ranked method with the incomplete datasets kept (Skillings--Mack test; \citealp{skillings1981}), every ranked method on the five
datasets where all of them have a value, and, in the main run, the 11 methods of the replication run. With the
permutation $p$, the Holm-corrected $p$ of C1 in the main run is 0.005 (classification) and $10^{-4}$
(regression), and no verdict of C1--C4 changes in either run.

\begin{table}[htbp]
  \caption{C1 and its sensitivity. $p$: chi-square approximation / permutation; $\leq$: the smallest
  attainable permutation $p$ (100\,000 permutations; 5\,000 for Skillings--Mack). Supplementary.}
  \label{tab:app-c1}
  \centering
  \small
  \begin{tabular}{@{}llccc@{}}
    \toprule
    Run & Variant & Methods $\times$ datasets & Classification $p$ & Regression $p$ \\
    \midrule
    main & pre-specified & $22 \times 9$ & 0.0014 / 0.0006 & $1.4 \cdot 10^{-8}$ / $\leq 10^{-5}$ \\
         & every method, incomplete (Skillings--Mack) & $24 \times 9$ & 0.0025 / 0.0012 & $3.6 \cdot 10^{-9}$ / $\leq 2 \cdot 10^{-4}$ \\
         & every method, complete datasets & $24 \times 5$ & 0.0015 / 0.0002 & 0.0018 / 0.0003 \\
         & methods of the replication run & $11 \times 9$ & 0.79 / 0.81 & 0.0002 / $3 \cdot 10^{-5}$ \\
    \midrule
    replication & pre-specified & $11 \times 9$ & 0.29 / 0.29 & $1.2 \cdot 10^{-4}$ / $4 \cdot 10^{-5}$ \\
         & every method, incomplete (Skillings--Mack) & $13 \times 9$ & 0.45 / 0.47 & $3.0 \cdot 10^{-5}$ / $\leq 2 \cdot 10^{-4}$ \\
         & every method, complete datasets & $13 \times 5$ & 0.88 / 0.90 & 0.0065 / 0.0015 \\
    \bottomrule
  \end{tabular}
\end{table}

\section{Absolute utility scores}
\label{app:absolute}
\Cref{tab:app-absolute} gives the absolute scores of the two confirmatory tasks in the main run: ROC AUC of
U2.1 (higher is better) and MASE of U2.6 at $h = 24$ (lower is better), for each downstream model and test
split. TRTR is the real series (mean over seeds, median over the nine datasets); TSTR is given as the median
over the 30 methods and for the best method, each the mean over seeds and the median over the datasets on
which the method is evaluated (TimesFM and Chronos on fewer). In single cells the best method reaches the real
series (a method of the zEDGE pool in classification, TimesFM and Timestral in regression); the transfer ratios
of RQ5 are medians over the tasks and models of a criterion. The scores of every task in both runs are in the
supplementary outputs (\texttt{rq5\_check}).

\begin{table}[tbp]
  \caption{Absolute scores of the confirmatory tasks, main run: TRTR, the median TSTR over the 30 methods and
  the best TSTR with its method. Supplementary. $^{*}$~proprietary (zypl.ai).}
  \label{tab:app-absolute}
  \centering
  \small
  \begin{tabular}{@{}lllcccl@{}}
    \toprule
    Task & Model & Split & TRTR & TSTR, median & TSTR, best & Best method \\
    \midrule
    U2.1 (AUC) & linear & val & 0.747 & 0.664 & 0.749 & \texttt{zedge\_pctsp}$^{*}$ \\
     & linear & OOT & 0.743 & 0.657 & 0.742 & \texttt{zedge\_ts\_ar}$^{*}$ \\
     & gbdt & val & 0.739 & 0.644 & 0.703 & \texttt{diffusion\_ts} \\
     & gbdt & OOT & 0.764 & 0.651 & 0.758 & \texttt{zedge\_ts\_ar}$^{*}$ \\
     & mlp & val & 0.757 & 0.652 & 0.727 & \texttt{zedge\_quilting}$^{*}$ \\
     & mlp & OOT & 0.789 & 0.672 & 0.757 & \texttt{zedge\_quilting}$^{*}$ \\
    \midrule
    U2.6 (MASE) & linear & val & 0.581 & 0.957 & 0.573 & \texttt{timestral} \\
     & linear & OOT & 0.832 & 0.888 & 0.841 & \texttt{zedge\_quilting}$^{*}$ \\
     & gbdt & val & 0.793 & 1.387 & 0.810 & \texttt{timesfm} \\
     & gbdt & OOT & 0.796 & 1.101 & 0.796 & \texttt{timesfm} \\
    \bottomrule
  \end{tabular}
\end{table}

\section{Seam metrics}
\label{app:seams}
F5.6 is the share of the change points of a series (PELT on two-hour means) that lie within three points of
a seam. If the change points had nothing to do with the seams, this share would be about the share of
positions within three points of a seam, which grows with the density of the seams. For every series, the
supplementary check (\texttt{seam\_check}) recomputes F5.6 (equal to the published value in every series)
and its chance level: the mean of the same share over 200 circular shifts of the seam positions by a random
offset, which keep the spacing of the seams but not their placement relative to the change points. The
ratio of the two is 1 when the change points fall on the seams as often as by chance and above 1 when more
often. \Cref{tab:app-seams} gives the medians per method in the main run; the replication run gives the same
picture. Across methods, the median F5.6 hardly follows the share of exposed positions (Spearman 0.08; 0.23
in the replication run), since most methods have a seam every 168 points; the methods with denser seams,
Chronos, Diffusion-TS and zEDGE\_TS, rank better when F5.6 is compared with its chance level, and Kendall's
$\tau$ between the two orders is 0.81.

The seam criterion of the core rankings is formed by F5.7 (\cref{tab:criteria}) and is undefined for
TimesFM, which has no seams, and for Timestral, which has fewer than the ten the classifier needs; both are then ranked on the remaining criteria of fidelity
(\cref{sec:rankings}). In a second supplementary variant, the analysis is repeated without the seam
metrics that are undefined for some method, so that only the rate of change points (F5.5) remains and the
seam criterion leaves the core rankings for every method. The overall order hardly changes (Kendall's
$\tau = 0.94$ with the published leaderboard in the main run, 0.92 in the replication run); in each run
one borderline method joins the top tier (\texttt{vae\_atsp} in the main run, \texttt{vae\_greedy} in the
replication run). The largest moves are \texttt{zedge\_ts\_ar}$^{*}$ from place 7 to 3 and Chronos from 20
to 22 in the main run and Chronos from 10 to 12 in the replication run; TimesFM and Timestral keep their
places, and no verdict of C1--C4 changes. The outputs of both variants are in the supplementary outputs
(\texttt{seam\_check}, \texttt{common\_metrics}).

\begin{table}[tbp]
  \caption{F5.6 against the density of the seams, main run: medians over the 90 series of each method of
  the number of seams, the share of positions within three points of a seam, F5.6, its chance level and
  their ratio (series with a seam and a change point). TimesFM has no seams. Sorted by F5.6. Supplementary. $^{*}$~proprietary (zypl.ai).}
  \label{tab:app-seams}
  \centering
  \small
  \begin{tabular}{@{}lrrrrr@{}}
    \toprule
    Method & Seams & Near a seam, \% & F5.6 & Chance level & Ratio \\
    \midrule
    \texttt{timestral} & 4 & 0.6 & 0.000 & 0.005 & 0.00 \\
    \texttt{vae\_zseries}$^{*}$ & 29 & 4.0 & 0.022 & 0.040 & 0.57 \\
    \texttt{zedge\_zseries}$^{*}$ & 29 & 4.0 & 0.024 & 0.040 & 0.60 \\
    \texttt{vae\_pctsp} & 29 & 4.0 & 0.027 & 0.040 & 0.67 \\
    \texttt{vae\_atsp} & 29 & 4.0 & 0.028 & 0.040 & 0.67 \\
    \texttt{zedge\_ts\_ar}$^{*}$ & 58 & 8.1 & 0.029 & 0.081 & 0.38 \\
    \texttt{vae\_greedy} & 29 & 4.0 & 0.029 & 0.040 & 0.72 \\
    \texttt{diffusion\_ts} & 58 & 8.1 & 0.036 & 0.080 & 0.45 \\
    \texttt{vae\_quilting} & 32 & 4.4 & 0.041 & 0.045 & 0.93 \\
    \texttt{zedge\_atsp}$^{*}$ & 29 & 4.0 & 0.045 & 0.040 & 1.15 \\
    \texttt{zedge\_pctsp}$^{*}$ & 29 & 4.0 & 0.045 & 0.040 & 1.15 \\
    \texttt{zedge\_quilting}$^{*}$ & 32 & 4.4 & 0.045 & 0.044 & 1.02 \\
    \texttt{vae\_knn} & 29 & 4.0 & 0.049 & 0.039 & 1.18 \\
    \texttt{vae\_zseries\_quilt}$^{*}$ & 32 & 4.4 & 0.051 & 0.044 & 1.16 \\
    \texttt{vae\_zseries\_xfade}$^{*}$ & 32 & 4.4 & 0.053 & 0.044 & 1.13 \\
    \texttt{zedge\_zseries\_quilt}$^{*}$ & 32 & 4.4 & 0.055 & 0.044 & 1.27 \\
    \texttt{zedge\_ts\_long}$^{*}$ & 58 & 8.1 & 0.055 & 0.079 & 0.69 \\
    \texttt{zedge\_zseries\_xfade}$^{*}$ & 32 & 4.4 & 0.059 & 0.044 & 1.34 \\
    \texttt{zedge\_greedy}$^{*}$ & 29 & 4.0 & 0.062 & 0.039 & 1.56 \\
    \texttt{vae\_zseries\_wsola}$^{*}$ & 33 & 4.6 & 0.081 & 0.046 & 1.75 \\
    \texttt{zedge\_knn}$^{*}$ & 29 & 4.0 & 0.081 & 0.039 & 2.00 \\
    \texttt{zedge\_zseries\_wsola}$^{*}$ & 33 & 4.6 & 0.086 & 0.046 & 1.87 \\
    \texttt{chronos} & 78 & 10.8 & 0.091 & 0.109 & 0.83 \\
    \texttt{vae\_mbb} & 29 & 4.0 & 0.095 & 0.040 & 2.49 \\
    \texttt{vae\_rand} & 29 & 4.0 & 0.100 & 0.039 & 2.51 \\
    \texttt{vae\_herding} & 29 & 4.0 & 0.105 & 0.039 & 2.65 \\
    \texttt{zedge\_mbb}$^{*}$ & 29 & 4.0 & 0.140 & 0.039 & 3.15 \\
    \texttt{zedge\_herding}$^{*}$ & 29 & 4.0 & 0.142 & 0.039 & 3.67 \\
    \texttt{zedge\_rand}$^{*}$ & 29 & 4.0 & 0.151 & 0.039 & 3.89 \\
    \texttt{timesfm} & 0 & 0.0 & -- & -- & -- \\
    \bottomrule
  \end{tabular}
\end{table}

\section{Memorisation diagnostics}
\label{app:memorisation}
F10 cuts windows of length $T$ from the synthetic series, the real val series and train, z-scored with val,
and takes for every synthetic and every val window the distance to the closest training window (DCR). F10.1
is the absolute logarithm of the ratio of the median DCR of the synthetic windows to that of the val windows;
F10.2 is the share of synthetic windows whose DCR is below the 5th percentile of the val windows' DCR (near
copies). The supplementary check (\texttt{memorisation\_check}) recomputes the same windows for every series
and for the real OOT series, keeps the sign of the logarithm (below 0: closer to train than the val windows),
counts the exact copies (DCR below 0.05 in z units; the training windows are cut with step $T/4$, so a copy
that is not aligned with this grid is not counted) and gives medians over datasets of the mean over seeds
with two-level bootstrap intervals (datasets, then seeds). The recomputed values equal the published ones
(largest difference $3 \times 10^{-4}$ for F10.1; F10.2 differs in one series, by 0.009). The same windows are scored for controls on every dataset and seed:
the first points of train (every window an exact copy), the same with Gaussian noise of 0.05 to 0.5 standard
deviations of train, and the real OOT series with its first tenth, quarter or half replaced by train.

The near-copy flag of the analysis compares the share of near copies of a method with that of the real OOT
series: it is raised when the median ratio over datasets is at least 2 and the excess is significant
(one-sided Wilcoxon over datasets, Holm). The share of the real series is 1 on three datasets (bike,
electricity, metro: its windows are closer to train than almost all val windows), 0 on three (Beijing PM2.5,
bitcoin, household) and 0.13 to 0.57 on the others. For an exact copy of train the ratio is then 1 or undefined on six of the
nine datasets, and its median is 1.38 (Holm $p = 0.125$); no control is
flagged. The signed ratio and the exact copies separate the controls (\cref{tab:app-memorisation}). Copies
with noise up to 0.25 standard deviations lie far below every method, whose windows are never exact copies;
noise of 0.5 standard deviations and partial copies are within the spread of the methods and of the real
series. The replication run gives the same picture.

\begin{table}[tbp]
  \caption{Memorisation diagnostics, main run: signed log ratio of the median DCR (synthetic against val
  windows; below 0: closer to train), share of near copies, each the median over datasets with a 95\,\%
  bootstrap interval, and exact copies among all windows. The real OOT series is scored once per dataset;
  the controls are described in the text (copy: DCR $= 0$). Supplementary.
  $^{*}$~proprietary (zypl.ai).}
  \label{tab:app-memorisation}
  \centering
  \small
  \begin{tabular}{@{}lrlrll@{}}
    \toprule
    Method & Log ratio & 95\,\% interval & Near copies & 95\,\% interval & Exact copies \\
    \midrule
    \texttt{timestral} & $-0.31$ & [$-0.74$, $-0.08$] & 0.61 & [0.00, 1.00] & 0 / 10\,530 \\
    \texttt{timesfm} & $-0.19$ & [$-0.50$, $-0.05$] & 0.50 & [0.09, 0.86] & 0 / 10\,530 \\
    \texttt{diffusion\_ts} & $-0.18$ & [$-0.47$, $0.17$] & 0.27 & [0.05, 0.69] & 0 / 10\,530 \\
    \texttt{vae\_pctsp} & $-0.17$ & [$-0.55$, $0.68$] & 0.22 & [0.00, 0.75] & 0 / 10\,530 \\
    \texttt{vae\_atsp} & $-0.15$ & [$-0.58$, $0.67$] & 0.24 & [0.00, 0.77] & 0 / 10\,530 \\
    \texttt{vae\_zseries\_xfade}$^{*}$ & $-0.12$ & [$-0.23$, $0.78$] & 0.12 & [0.00, 0.40] & 0 / 10\,530 \\
    \texttt{vae\_zseries\_wsola}$^{*}$ & $-0.12$ & [$-0.34$, $0.59$] & 0.35 & [0.00, 0.76] & 0 / 10\,530 \\
    \texttt{vae\_quilting} & $-0.12$ & [$-0.57$, $0.65$] & 0.21 & [0.00, 0.73] & 0 / 10\,530 \\
    \texttt{vae\_zseries\_quilt}$^{*}$ & $-0.11$ & [$-0.23$, $0.78$] & 0.11 & [0.00, 0.38] & 0 / 10\,530 \\
    \texttt{vae\_zseries}$^{*}$ & $-0.11$ & [$-0.59$, $0.58$] & 0.32 & [0.00, 0.79] & 0 / 10\,530 \\
    \texttt{vae\_knn} & $-0.08$ & [$-0.48$, $0.72$] & 0.19 & [0.00, 0.73] & 0 / 10\,530 \\
    \texttt{vae\_greedy} & $-0.07$ & [$-0.55$, $0.63$] & 0.21 & [0.00, 0.78] & 0 / 10\,530 \\
    \texttt{vae\_rand} & $-0.05$ & [$-0.35$, $0.76$] & 0.15 & [0.00, 0.55] & 0 / 10\,530 \\
    \texttt{vae\_mbb} & $-0.04$ & [$-0.37$, $0.77$] & 0.15 & [0.00, 0.50] & 0 / 10\,530 \\
    \texttt{vae\_herding} & $-0.02$ & [$-0.30$, $0.52$] & 0.09 & [0.01, 0.60] & 0 / 10\,530 \\
    \texttt{zedge\_quilting}$^{*}$ & $-0.02$ & [$-0.80$, $0.22$] & 0.25 & [0.02, 0.76] & 0 / 10\,530 \\
    \texttt{zedge\_atsp}$^{*}$ & $0.00$ & [$-0.94$, $0.20$] & 0.36 & [0.02, 0.86] & 0 / 10\,530 \\
    \texttt{zedge\_pctsp}$^{*}$ & $0.00$ & [$-0.95$, $0.21$] & 0.34 & [0.02, 0.84] & 0 / 10\,530 \\
    \texttt{zedge\_zseries}$^{*}$ & $0.01$ & [$-0.89$, $0.25$] & 0.26 & [0.03, 0.90] & 0 / 10\,530 \\
    \texttt{zedge\_zseries\_wsola}$^{*}$ & $0.01$ & [$-0.82$, $0.26$] & 0.32 & [0.01, 0.89] & 0 / 10\,530 \\
    \texttt{zedge\_zseries\_quilt}$^{*}$ & $0.02$ & [$-0.42$, $0.36$] & 0.05 & [0.02, 0.44] & 0 / 10\,530 \\
    \texttt{zedge\_zseries\_xfade}$^{*}$ & $0.02$ & [$-0.43$, $0.35$] & 0.05 & [0.02, 0.44] & 0 / 10\,530 \\
    \texttt{zedge\_knn}$^{*}$ & $0.02$ & [$-0.72$, $0.22$] & 0.33 & [0.01, 0.89] & 0 / 10\,530 \\
    \texttt{zedge\_greedy}$^{*}$ & $0.03$ & [$-0.89$, $0.22$] & 0.34 & [0.02, 0.88] & 0 / 10\,530 \\
    \texttt{zedge\_ts\_ar}$^{*}$ & $0.07$ & [$-0.43$, $0.25$] & 0.16 & [0.04, 0.77] & 0 / 10\,530 \\
    \texttt{zedge\_ts\_long}$^{*}$ & $0.08$ & [$-0.37$, $0.23$] & 0.17 & [0.05, 0.77] & 0 / 10\,530 \\
    \texttt{zedge\_herding}$^{*}$ & $0.11$ & [$-0.52$, $0.36$] & 0.21 & [0.01, 0.66] & 0 / 10\,530 \\
    \texttt{zedge\_mbb}$^{*}$ & $0.12$ & [$-0.51$, $0.37$] & 0.21 & [0.01, 0.61] & 0 / 10\,530 \\
    \texttt{zedge\_rand}$^{*}$ & $0.14$ & [$-0.47$, $0.38$] & 0.19 & [0.02, 0.60] & 0 / 10\,530 \\
    \texttt{chronos} & $0.28$ & [$-0.24$, $1.26$] & 0.23 & [0.01, 0.46] & 0 / 10\,530 \\
    \midrule
    real OOT series & $-0.13$ & -- & 0.13 & -- & 0 / 442 \\
    copy & $-\infty$ & -- & 1.00 & [1.00, 1.00] & 10\,300 / 10\,300 \\
    copy + noise 0.05 & $-2.67$ & [$-3.15$, $-2.12$] & 1.00 & [1.00, 1.00] & 0 / 10\,300 \\
    copy + noise 0.1 & $-1.98$ & [$-2.46$, $-1.43$] & 1.00 & [1.00, 1.00] & 0 / 10\,300 \\
    copy + noise 0.25 & $-1.06$ & [$-1.20$, $-0.51$] & 1.00 & [1.00, 1.00] & 0 / 10\,300 \\
    copy + noise 0.5 & $-0.37$ & [$-0.85$, $0.18$] & 0.50 & [0.00, 1.00] & 0 / 10\,300 \\
    partial copy 0.1 & $-0.13$ & [$-0.63$, $0.05$] & 0.12 & [0.00, 0.95] & 0 / 4\,419 \\
    partial copy 0.25 & $-0.15$ & [$-0.62$, $0.08$] & 0.17 & [0.02, 0.84] & 0 / 4\,403 \\
    partial copy 0.5 & $-0.30$ & [$-0.55$, $0.06$] & 0.38 & [0.14, 0.81] & 212 / 4\,509 \\
    \bottomrule
  \end{tabular}
\end{table}

\section{Noise-aware ranks}
\label{app:ties}
Within a block (dataset, seed and leaf) with $k$ methods and the noise $e$ of the leaf on that dataset, the
rank of method $i$ is $1 + \#\{j: s_j \text{ beats } s_i \text{ by more than } e\} + \tfrac12\,\#\{j \neq i:
|s_j - s_i| \le e\}$ (\cref{sec:rankings}). \Cref{tab:app-ties-example} works through three methods whose
ties are not transitive. The ranks do not depend on the order in which the methods are listed; this is
checked by a test of the released code, and on the published tables, where renaming the methods so that
they come in another order leaves every rank unchanged. In the main run, 28.5\,\% of the pairs of methods
compared within a block tie, and a further 11.7\,\% do not tie but are linked by a method that ties with
both (26.9\,\% and 10.3\,\% in the replication run). \Cref{tab:app-ties-plain} lists the methods whose
place or top-tier membership differs with ordinary average ranks. The contrasts of the research questions
and of C1--C4 use paired differences of the values, not ranks, and do not depend on this choice.

\begin{table}[htbp]
  \caption{Noise-aware and ordinary ranks of three methods in one block, noise $e = 1$, higher is better:
  A ties B and B ties C, but A beats C. The ranks sum to $k(k+1)/2 = 6$ in both cases.}
  \label{tab:app-ties-example}
  \centering
  \small
  \begin{tabular}{lccl}
    \toprule
    Method & Score & Noise-aware rank & Ordinary rank \\
    \midrule
    A & 1.2 & $1 + 0 + \tfrac12 \cdot 1 = 1.5$ & 1 \\
    B & 0.6 & $1 + 0 + \tfrac12 \cdot 2 = 2$   & 2 \\
    C & 0.0 & $1 + 1 + \tfrac12 \cdot 1 = 2.5$ & 3 \\
    \bottomrule
  \end{tabular}
\end{table}

\begin{table}[htbp]
  \caption{Methods whose place or top-tier membership differs between the noise-aware ranks of the
  leaderboard and ordinary average ranks; every other method keeps its place and tier. Kendall's $\tau$
  between the mean ranks: 0.97 (main run) and 0.95 (replication run).}
  \label{tab:app-ties-plain}
  \centering
  \small
  \begin{tabular}{llcccc}
    \toprule
    & & \multicolumn{2}{c}{Noise-aware} & \multicolumn{2}{c}{Ordinary} \\
    \cmidrule(lr){3-4}\cmidrule(lr){5-6}
    Run & Method & Place & Top tier & Place & Top tier \\
    \midrule
    Main        & \texttt{zedge\_knn}     & 5  & yes & 8  & yes \\
                & \texttt{zedge\_ts\_ar}  & 7  & yes & 5  & yes \\
                & \texttt{zedge\_ts\_long}& 8  & yes & 7  & yes \\
                & \texttt{zedge\_rand}    & 11 & yes & 11 & no  \\
    \midrule
    Replication & \texttt{vae\_mbb}       & 8  & yes & 9  & yes \\
                & \texttt{vae\_pctsp}     & 9  & yes & 8  & yes \\
                & \texttt{vae\_knn}       & 11 & yes & 12 & yes \\
                & \texttt{vae\_greedy}    & 12 & no  & 11 & no  \\
    \bottomrule
  \end{tabular}
\end{table}

\section{Top tier and the selection of the leader}
\label{app:tier}
\Cref{tab:app-tier} gives, for both runs, the number of datasets on which each method is paired with the
observed leader, the Holm-corrected $p$-value of the one-sided Wilcoxon test that it is worse, and its
membership of the top tier. Every method also belongs to the set of methods that no other method is shown
to beat when every ordered pair is tested with Holm correction over all pairs (552 pairs in the main run, 156
in the replication run): with nine datasets, the smallest one-sided $p$ is $1/512$, and no pair is rejected.
Both sets are non-rejection sets (\cref{sec:rankings}).

\begin{table}[tbp]
  \centering\small
  \caption{Comparisons with the observed leader of each run: paired datasets, Holm-corrected $p$ that the
  method is worse, and top-tier membership. Leader: \texttt{zedge\_quilting} (main run) and
  \texttt{vae\_herding} (replication run).}
  \label{tab:app-tier}
  \begin{tabular}{@{}lccclccc@{}}
  \toprule
  \multicolumn{4}{c}{Main run} & \multicolumn{4}{c}{Replication run} \\
  \cmidrule(r){1-4}\cmidrule(l){5-8}
  Method & Datasets & $p_{\mathrm{Holm}}$ & Tier & Method & Datasets & $p_{\mathrm{Holm}}$ & Tier \\
  \midrule
  \texttt{zedge\_quilting}$^{*}$ & 9 & -- & yes & \texttt{vae\_herding} & 9 & -- & yes \\
  \texttt{zedge\_zseries}$^{*}$ & 9 & 0.410 & yes & \texttt{diffusion\_ts} & 9 & 0.500 & yes \\
  \texttt{zedge\_greedy}$^{*}$ & 9 & 0.410 & yes & \texttt{vae\_quilting} & 9 & 0.500 & yes \\
  \texttt{zedge\_pctsp}$^{*}$ & 9 & 0.293 & yes & \texttt{vae\_zseries}$^{*}$ & 9 & 0.500 & yes \\
  \texttt{zedge\_knn}$^{*}$ & 9 & 0.410 & yes & \texttt{timesfm} & 6 & 0.500 & yes \\
  \texttt{zedge\_atsp}$^{*}$ & 9 & 0.410 & yes & \texttt{vae\_atsp} & 9 & 0.088 & yes \\
  \texttt{zedge\_ts\_ar}$^{*}$ & 9 & 0.410 & yes & \texttt{vae\_rand} & 9 & 0.088 & yes \\
  \texttt{zedge\_ts\_long}$^{*}$ & 9 & 0.191 & yes & \texttt{vae\_mbb} & 9 & 0.088 & yes \\
  \texttt{zedge\_herding}$^{*}$ & 9 & 0.107 & yes & \texttt{vae\_pctsp} & 9 & 0.088 & yes \\
  \texttt{zedge\_mbb}$^{*}$ & 9 & 0.123 & yes & \texttt{chronos} & 7 & 0.391 & yes \\
  \texttt{zedge\_rand}$^{*}$ & 9 & 0.107 & yes & \texttt{vae\_knn} & 9 & 0.059 & yes \\
  \texttt{vae\_herding} & 9 & 0.045 & no & \texttt{vae\_greedy} & 9 & 0.047 & no \\
  \texttt{vae\_zseries}$^{*}$ & 9 & 0.055 & yes & \texttt{timestral} & 9 & 0.047 & no \\
  \texttt{diffusion\_ts} & 9 & 0.045 & no &  & & &  \\
  \texttt{timesfm} & 6 & 0.125 & yes &  & & &  \\
  \texttt{vae\_quilting} & 9 & 0.055 & yes &  & & &  \\
  \texttt{vae\_pctsp} & 9 & 0.045 & no &  & & &  \\
  \texttt{vae\_atsp} & 9 & 0.045 & no &  & & &  \\
  \texttt{vae\_greedy} & 9 & 0.045 & no &  & & &  \\
  \texttt{chronos} & 7 & 0.094 & yes &  & & &  \\
  \texttt{vae\_knn} & 9 & 0.045 & no &  & & &  \\
  \texttt{vae\_mbb} & 9 & 0.045 & no &  & & &  \\
  \texttt{vae\_rand} & 9 & 0.045 & no &  & & &  \\
  \texttt{timestral} & 9 & 0.045 & no &  & & &  \\
  \bottomrule
  \end{tabular}
\end{table}

\section{Software}
\label{app:software}
The figures of this appendix give the details of \cref{sec:software}: the architecture of the code
(\cref{fig:architecture}), the software environments (\cref{fig:environments}), the fingerprints of
incremental recomputation (\cref{fig:fingerprints}), the layout of the outputs (\cref{fig:artifacts}), the run
profiles (\cref{fig:profiles}) and the extension points (\cref{fig:extensibility}).

\fig{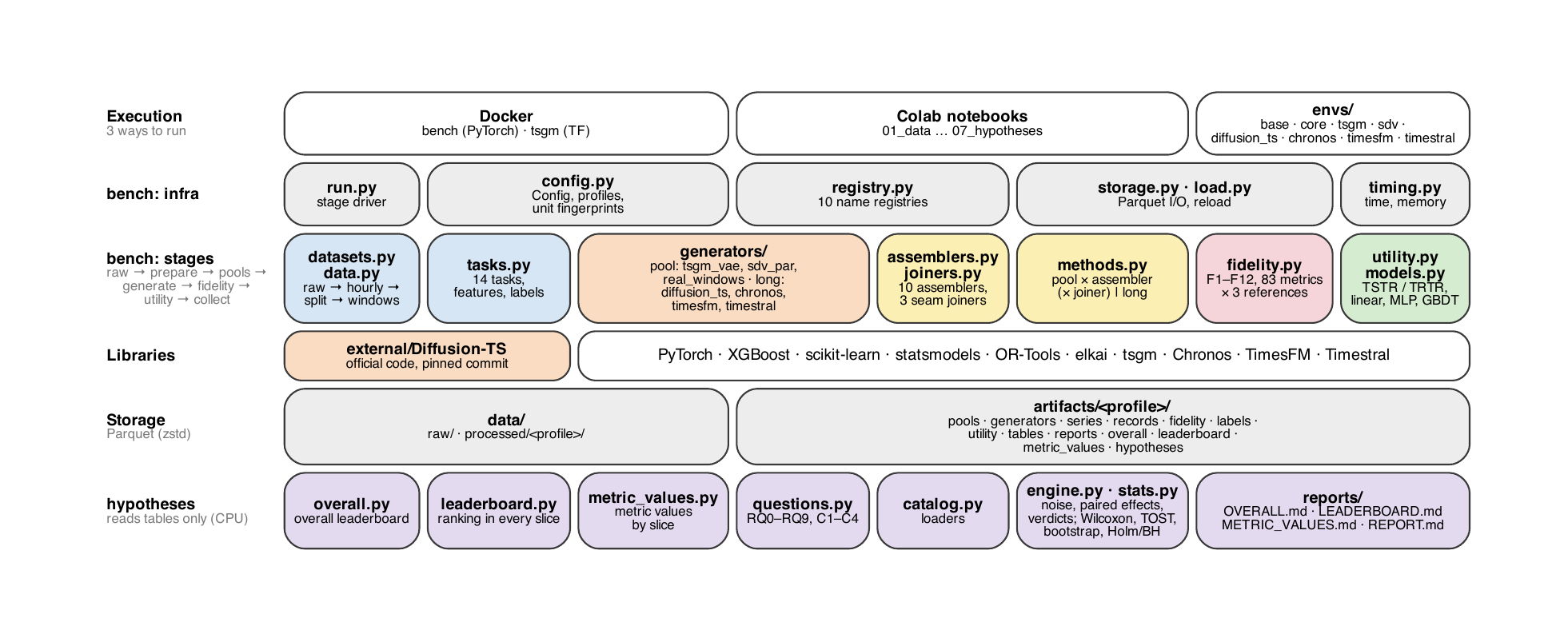}{architecture}{1}{Layered architecture of the benchmark. The
\texttt{bench} package runs the pipeline stages and finds every component by name in a registry. The
official Diffusion-TS code sits underneath it. All stages exchange data only through Parquet files
in \texttt{data/} and \texttt{artifacts/}. The \texttt{hypotheses} package reads the collected tables
and tests the hypotheses.}

\fig{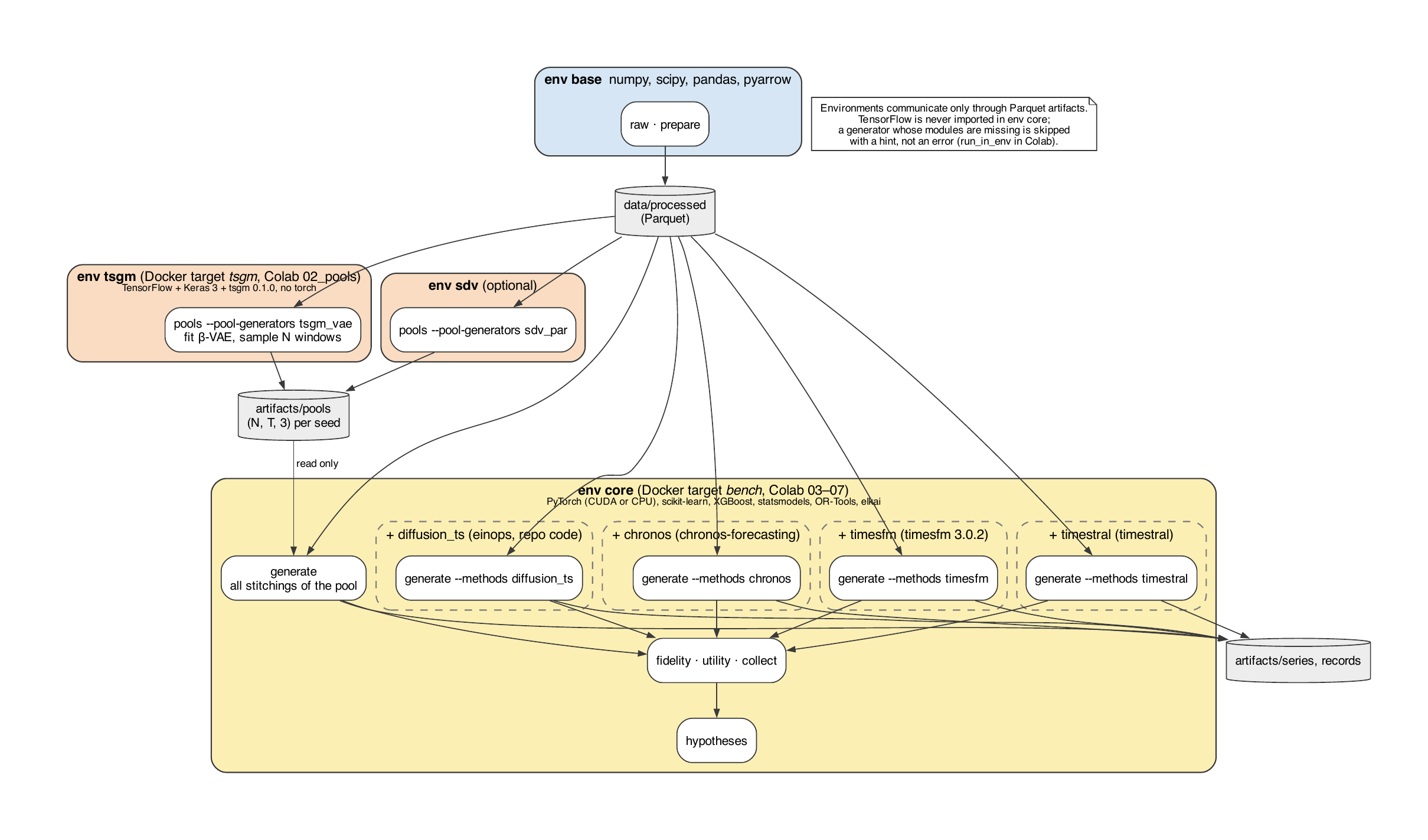}{environments}{0.9}{Isolation of conflicting dependencies. The VAE pool
generator runs in a TensorFlow environment without PyTorch. Diffusion-TS, Chronos, TimesFM and
Timestral each run in their own PyTorch environment, and assembly, evaluation and the statistics in a
common core environment that never imports TensorFlow. The only link between the environments is the
Parquet files they write.}

\fig{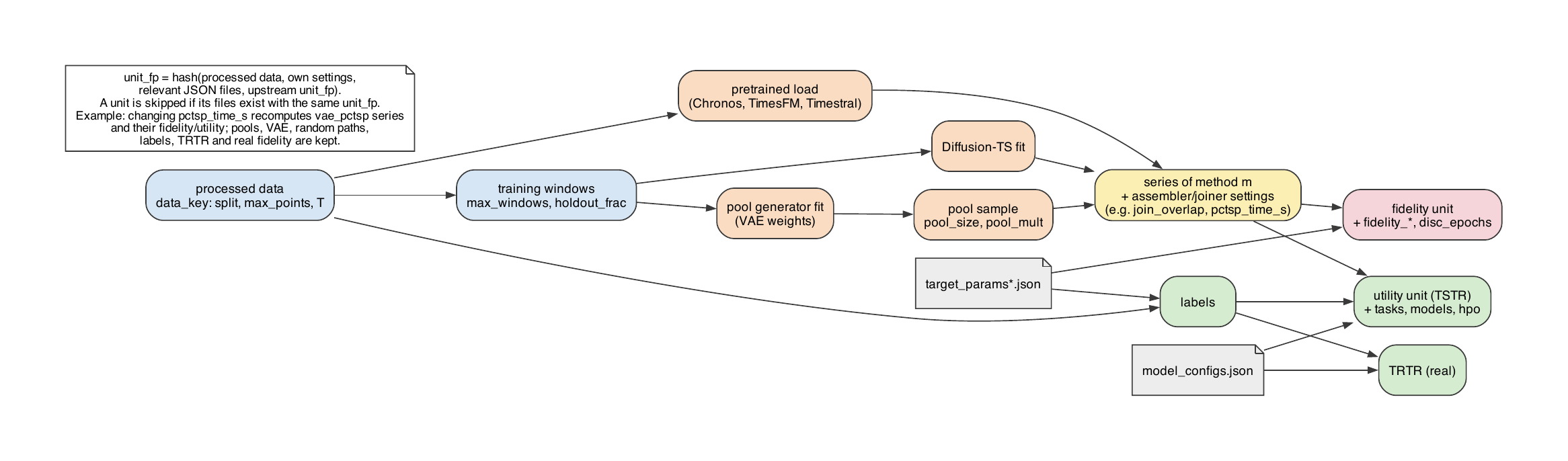}{fingerprints}{0.9}{Dependency graph of unit fingerprints. Each unit
of work stores a hash of its inputs, its own settings, the relevant configuration files, and the
fingerprints of the units it depends on. When a setting changes, only the units downstream of it are
recomputed.}

\fig{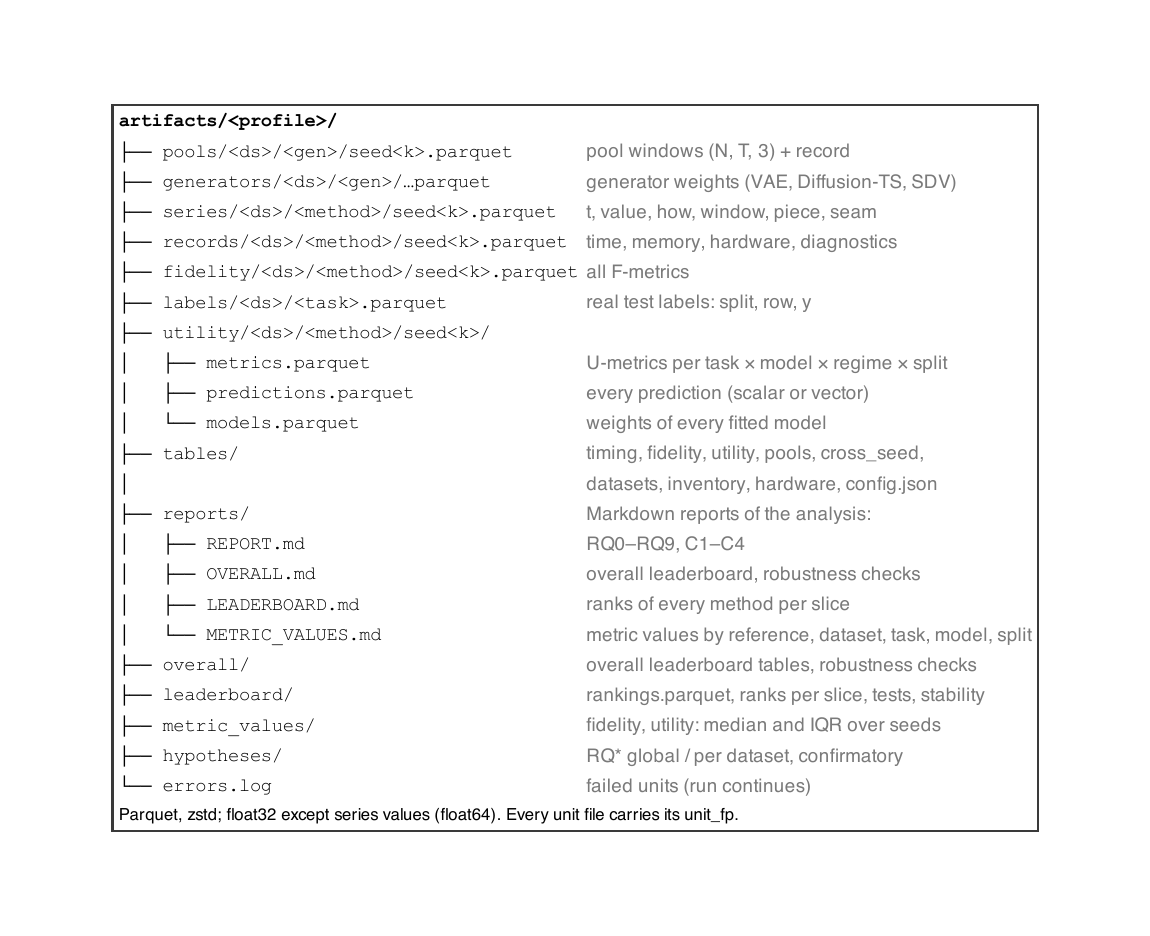}{artifacts}{0.9}{Layout of the benchmark outputs. Every unit of work
writes its own Parquet file. The \texttt{collect} stage merges them into the published tables in
\texttt{tables/}.}

\fig{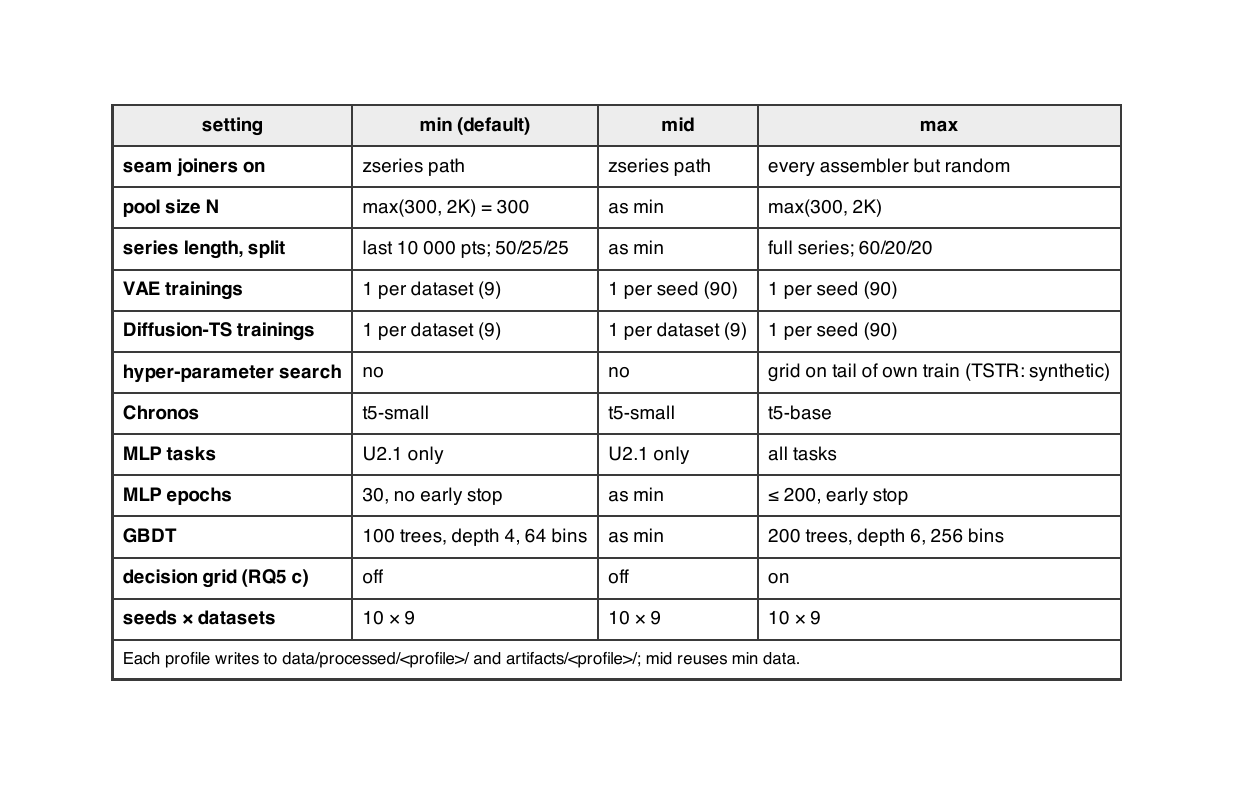}{profiles}{1}{The three run profiles. \texttt{min} (default, intended
for Colab) shares one trained generator per dataset across seeds. \texttt{mid} adds generator
replicas. \texttt{max} uses the full series, retrains generators on every seed, and turns on
hyperparameter search and the decision-transfer grid.}

\fig{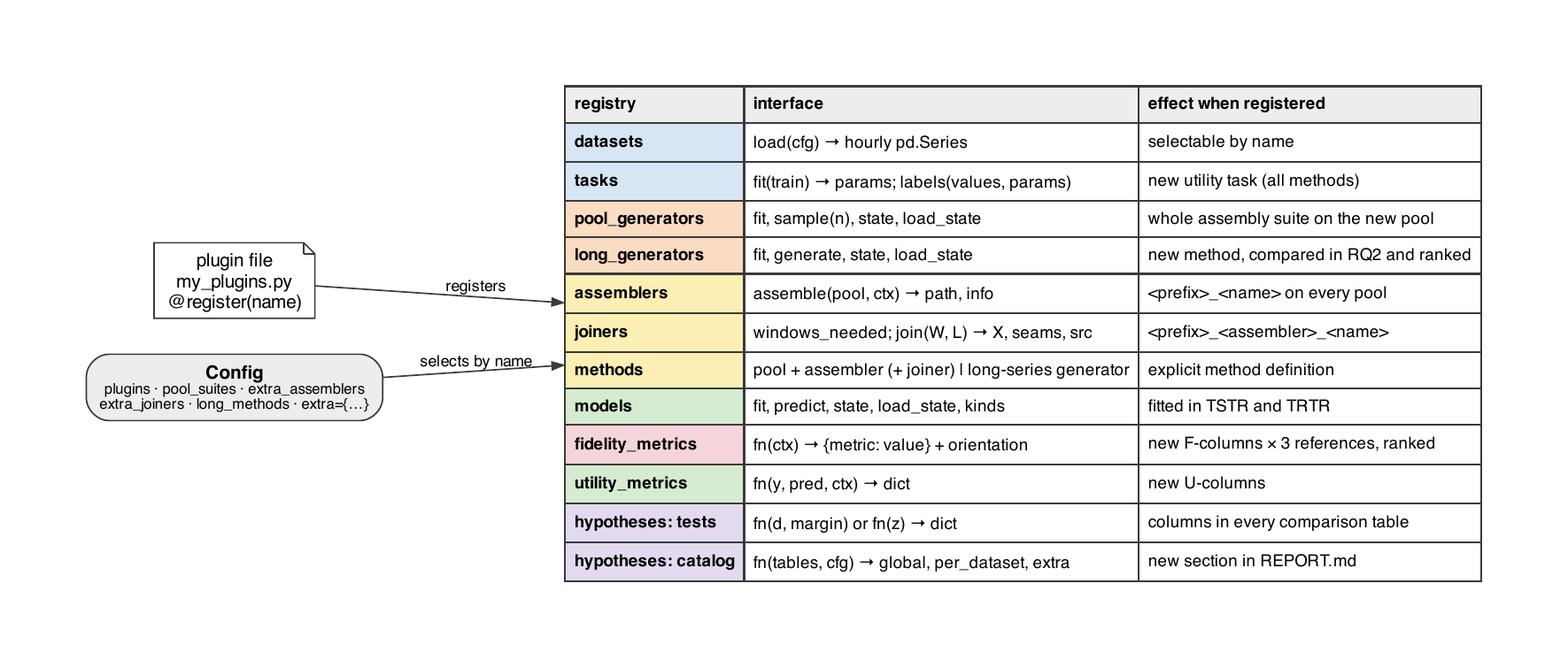}{extensibility}{1}{Extension points. Every part of the benchmark is
registered by name. A plugin file registers new components, and the configuration selects them. A
new pool generator, assembler, seam joiner or long-series generator becomes part of the comparison
and of the relevant hypotheses without changes to the benchmark code.}

\section{Runs and reproduction}
\label{app:runs}
\Cref{tab:app-runs} describes the two runs of \cref{sec:runs}. Both used profile \texttt{min}, the environments
pinned in the repository and the same task thresholds and seeds; every generation record stores the GPU on
which it ran, and each run used a single GPU type. The software versions of the replication run were not
recorded beyond the pinned requirements.

\begin{table}[tbp]
  \centering\small
  \caption{The two runs of profile \texttt{min}.}
  \label{tab:app-runs}
  \begin{tabularx}{\linewidth}{@{}l>{\raggedright\arraybackslash}X>{\raggedright\arraybackslash}X@{}}
    \toprule
     & Main run & Replication run \\
    \midrule
    GPU & 1$\times$ NVIDIA H200 (143\,GB) & 1$\times$ NVIDIA L4 (24\,GB) \\
    CPU cores & 192 & 12 \\
    System & Ubuntu 22.04.5, driver 580.173.02 & Google Colab (Linux 6.6) \\
    Software & Python 3.12.13, PyTorch 2.9.1, TensorFlow 2.20.0 (environment manifests published) & pinned requirements of the repository \\
    Code & internal commit 365232d6 (23 October 2026) & completed before the freeze; results in commit 7fea7530 (21 October 2026) \\
    Methods & 30 (24 ranked) & 16 of the VAE pool and the open generators (13 ranked) \\
    Role & every result of this paper unless stated & reproducibility check (\cref{sec:reproduction}) \\
    \bottomrule
  \end{tabularx}
\end{table}

{\small
\begin{longtable}{@{}lcccccccc@{}}
  \caption{Overall leaderboard of the 13 methods common to both runs: the replication run (NVIDIA L4) and the
  main run's own copies of these methods ranked among themselves (NVIDIA H200). Place, mean rank, its 95\,\%
  bootstrap interval and top-tier membership. $^{*}$~proprietary (zypl.ai).}
  \label{tab:app-replication}\\
  \toprule
   & \multicolumn{4}{c}{Replication run (L4)} & \multicolumn{4}{c}{Main run (H200)} \\
  \cmidrule(lr){2-5}\cmidrule(l){6-9}
  Method & Place & Mean rank & 95\,\% interval & Tier & Place & Mean rank & 95\,\% interval & Tier \\
  \midrule
  \endfirsthead
  \toprule
  Method & Place & Mean rank & 95\,\% interval & Tier & Place & Mean rank & 95\,\% interval & Tier \\
  \midrule
  \endhead
  \bottomrule
  \endfoot
  \texttt{vae\_herding} & 1 & 5.54 & [4.86, 6.18] & yes & 1 & 5.57 & [4.62, 6.49] & yes \\
  \texttt{diffusion\_ts} & 2 & 6.27 & [4.76, 7.59] & yes & 3 & 6.24 & [4.91, 7.43] & yes \\
  \texttt{vae\_quilting} & 3 & 6.36 & [5.82, 6.92] & yes & 4 & 6.51 & [5.98, 7.10] & yes \\
  \texttt{vae\_zseries}$^{*}$ & 4 & 6.41 & [5.67, 7.20] & yes & 2 & 5.76 & [5.16, 6.35] & yes \\
  \texttt{timesfm} & 5 & 6.72 & [5.29, 8.27] & yes & 5 & 6.77 & [5.43, 8.26] & yes \\
  \texttt{vae\_atsp} & 6 & 7.11 & [6.55, 7.74] & yes & 7 & 6.99 & [6.45, 7.51] & yes \\
  \texttt{vae\_rand} & 7 & 7.20 & [6.45, 7.86] & yes & 12 & 7.84 & [7.17, 8.52] & yes \\
  \texttt{vae\_mbb} & 8 & 7.22 & [6.57, 7.83] & yes & 10 & 7.71 & [7.02, 8.40] & yes \\
  \texttt{vae\_pctsp} & 9 & 7.23 & [6.79, 7.70] & yes & 6 & 6.97 & [6.63, 7.32] & yes \\
  \texttt{chronos} & 10 & 7.43 & [6.11, 8.59] & yes & 11 & 7.75 & [6.78, 8.58] & yes \\
  \texttt{vae\_knn} & 11 & 7.56 & [7.10, 8.08] & yes & 9 & 7.53 & [7.06, 8.03] & yes \\
  \texttt{vae\_greedy} & 12 & 7.60 & [7.16, 8.08] & no & 8 & 7.18 & [6.76, 7.64] & yes \\
  \texttt{timestral} & 13 & 8.56 & [7.46, 9.50] & no & 13 & 8.42 & [7.48, 9.27] & no \\
\end{longtable}
}

\end{document}